\pdfoutput=1

\documentclass[11pt]{article}

\usepackage[final]{acl}

\usepackage{times}
\usepackage{latexsym}

\usepackage[T1]{fontenc}

\usepackage[utf8]{inputenc}

\usepackage{microtype}

\usepackage{inconsolata}

\usepackage{graphicx}

\usepackage{booktabs}  
\usepackage{multirow}
\usepackage{graphicx}
\usepackage{subcaption}
\usepackage{url}
\usepackage{enumitem}
\usepackage{wrapfig}
\usepackage[utf8]{inputenc} 
\usepackage[T1]{fontenc}    
\usepackage{hyperref}       
\usepackage{url}            
\usepackage{amsfonts}       
\usepackage{nicefrac}       
\usepackage{microtype}      
\usepackage{xcolor}         
\usepackage{natbib}
\usepackage{caption}
\usepackage{multirow}
\usepackage{graphicx}
\usepackage{physics}
\usepackage{xr-hyper}
\usepackage{url}            
\usepackage{amsfonts}       
\usepackage{nicefrac} 
\usepackage{microtype}      
\usepackage{graphicx}
\usepackage{booktabs} 
\usepackage{amsfonts}
\usepackage{amssymb,bm}
\usepackage{amstext, enumitem}
\usepackage{amsmath}
\usepackage{xspace}
\usepackage{amsthm}
\usepackage{algorithm}
\usepackage{algpseudocode}
\usepackage{graphicx,color}
\usepackage{url}
\usepackage{graphics}
\usepackage{colordvi, wrapfig}
\usepackage{colordvi}
\usepackage{makecell}
\usepackage{amssymb}
\usepackage{todonotes}
\usepackage{multirow}

\usepackage{adjustbox}
\usepackage{array}

\newtheorem{Lem}{Lemma}
\newtheorem{myTheo}{Theorem}

\title{A Middle Path for On-Premises LLM Deployment: \\Preserving Privacy Without Sacrificing Model Confidentiality}

\author{Hanbo Huang\textsuperscript{1}, Yihan Li\textsuperscript{2}, Bowen Jiang\textsuperscript{1}, Bo Jiang\textsuperscript{1}, Lin Liu\textsuperscript{2}, \\
\textbf{Ruoyu Sun\textsuperscript{3}}, \textbf{Zhuotao Liu\textsuperscript{4}}, \textbf{Shiyu Liang\textsuperscript{1}} \\
\textsuperscript{1}Shanghai Jiao Tong University, \textsuperscript{2}National University of Defense Technology,\\
\textsuperscript{3}Chinese University of Hong Kong (Shenzhen), \textsuperscript{4}Tsinghua University\\
\texttt{\{hhuang417,lsy18602808513\}@sjtu.edu.cn}
}

\definecolor{myred}{HTML}{c3272b} 
\definecolor{mygreen}{HTML}{057748}
\newcommand{\red}[1]{\textcolor{myred}{#1}}
\newcommand{\green}[1]{\textcolor{mygreen}{#1}}

\begin{document}
\maketitle
\begin{abstract}
Privacy-sensitive users require deploying large language models (LLMs) within their own infrastructure (\textit{on-premises}) to safeguard private data and enable customization. However, vulnerabilities in local environments can lead to unauthorized access and potential model theft. To address this, prior research on small models has explored securing only the output layer within hardware-secured devices to balance model confidentiality and customization. Yet this approach fails to protect LLMs effectively. In this paper, we discover that (1) query-based distillation attacks targeting the secured top layer can produce a functionally equivalent replica of the victim model; (2) securing the same number of layers, bottom layers before a transition layer provide stronger protection against distillation attacks than top layers, with comparable effects on customization performance; and (3) the number of secured layers creates a trade-off between protection and customization flexibility. Based on these insights, we propose SOLID, a novel deployment framework that secures a few bottom layers in a secure environment and introduces an efficient metric to optimize the trade-off by determining the ideal number of hidden layers. Extensive experiments on five models (1.3B to 70B parameters) demonstrate that SOLID outperforms baselines, achieving a better balance between protection and downstream customization. Our code can be found at: \url{https://github.com/OTTO-OTO/SOLID-OnPremiseDeployment}.

\end{abstract}

\section{Introduction}
\begin{figure}
    \centering
    \includegraphics[width=0.95\linewidth]{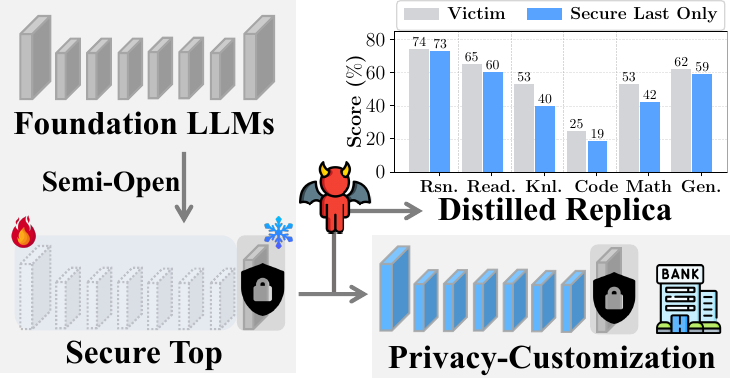}
    \caption{Semi-open Deployment.}
    \vspace{-0.5cm}
  \label{fig:Introduction}
\end{figure}

Vendors of Large Language Models (LLMs) have introduced advanced models with remarkable capabilities to address diverse user needs~\citep{minaee2024largelanguagemodelssurvey,zhao2023surveylargelanguagemodels}. To meet specific customization demands, vendors typically adopt two approaches. Closed-source vendors, such as \href{https://platform.openai.com/docs/guides/fine-tuning}{OpenAI}, provide fine-tuning APIs that allow users to upload data to customize proprietary models like \href{https://platform.openai.com/docs/guides/fine-tuning}{GPT-4}. In contrast, vendors like Meta offer open-weight models such as Llama3~\citep{dubey2024llama3}, which users can adapt within their own infrastructure, ensuring greater flexibility and control.

However, both approaches present notable limitations for privacy-sensitive users, such as healthcare organizations, who prioritize data security. Strict regulations prohibit these users from uploading sensitive data to third-party API services, necessitating \textit{on-premises deployment} of LLMs, where data processing and model customization are confined to local infrastructure~\citep{nevo2024securing}. Although fine-tuning open-weight models in such environments offers a viable path for customization, full disclosure of model architectures and weights increases the risk of exploitation by malicious actors, who may circumvent safety mechanisms~\citep{hendrycks2023overviewcatastrophicairisks}. Consequently, vendors may be hesitant to release SOTA models as open-weight, since uncontrolled access could lead to significant harm. Moreover, maintaining high-quality open-weight models imposes considerable computational and financial burdens~\citep{wolfe2024laboratory}. These growing concerns highlight the importance of \textit{secure on-premises deployment of closed-source models}, which preserves control and ensures regulatory compliance.

Despite the advantages, deploying closed-source LLMs locally introduces the risk of model theft. Unauthorized users can extract model parameters and architectures directly from CPUs and memory within local environments~\citep{hu2020deepsniffer}. To mitigate this, existing approaches use Trusted Execution Environments (TEEs) to protect proprietary models~\citep{nayan2024sok,narra2019privacy}. Yet, fully enclosing large models within TEEs results in prohibitive computational overhead, limiting their practicality~\citep{li2024teeslice}. 

Prior research has explored to mitigate this trade-off by securing only critical layers, such as the output layer, while leaving the remaining layers exposed for fine-tuning~\citep{zhang2024no,mo2020darknetz}. However, studies have shown that even with only black-box access, adversaries may still be able to replicate the weights of DNNs~\citep{tramer2016stealing,truong2021data}. More recent works~\citep{carlini2024stealing,finlayson2024logits} further suggests that the final-layer weights of large language models (LLMs) can be recovered from output logits alone, raising concerns about the robustness of such partial protection strategies. Consistent with previous findings, our results show that this partial protection remains vulnerable to distillation attacks~\citep{greybox}. 
When extended to Llama2-70B, we confirm that attackers can still extract nearly complete model functionality across six domains, as illustrated in Figure~\ref{fig:Introduction}. These vulnerabilities raise skepticism about \textit{whether model confidentiality and customization can truly coexist in on-premises deployment}, highlighting the need for security paradigms beyond output-layer protection.

In this paper, we show that this dilemma can be resolved. We begin by investigating the security-customization trade-off introduced by the placement of secured layers in LLMs. Specifically, we theoretically identify a transition layer in deep transformers, showing that securing bottom layers before this transition significantly reduces distillation success, while securing top layers has a more limited impact. Besides, we demonstrate that the number of secured layers creates a trade-off: securing more layers improves security but reduces customization flexibility. To optimize this trade-off, we introduce SOLID, a semi-open deployment framework that selectively secures a subset of bottom layers, using a distillation difficulty score to identify the optimal set for protection. Our experiments show that SOLID balances security and customization, achieving security comparable to fully secured models while maintaining strong customization flexibility, approaching full parameter fine-tuning. Our main contributions are as follows:

\begin{itemize}[leftmargin=*, itemsep=0pt, parsep=0pt, topsep=0pt]
    \item We extend query-based distillation attacks to LLMs, demonstrating that existing on-premises frameworks risk full functionality replication.
    \item We identify the security-customization trade-off introduced by the placement of secured layers, and theoretically prove that securing bottom layers before the transition layer offers stronger protection with similar customization effects.
    \item We discover that the number of secured layers affects both security and customization. We propose SOLID, which optimizes the security-customization trade-off by using a fine-tuning-free metric to secure minimal bottom decoder layers, protecting the model from distillation attacks while preserving customization flexibility.
    \item We evaluate SOLID against three baselines across five models (1.3B to 70B parameters), assessing security across three distillation strategies on sixteen benchmarks and customization flexibility across six tasks. Extensive experiments show SOLID effectively balances security and customization, despite some limitations
\end{itemize}


\section{Preliminaries}
\label{section:preliminaries}
\subsection{Security Threat: Model Distillation}
\label{subsection2.1}

\begin{figure}
    \centering
    \includegraphics[width=0.96\linewidth]{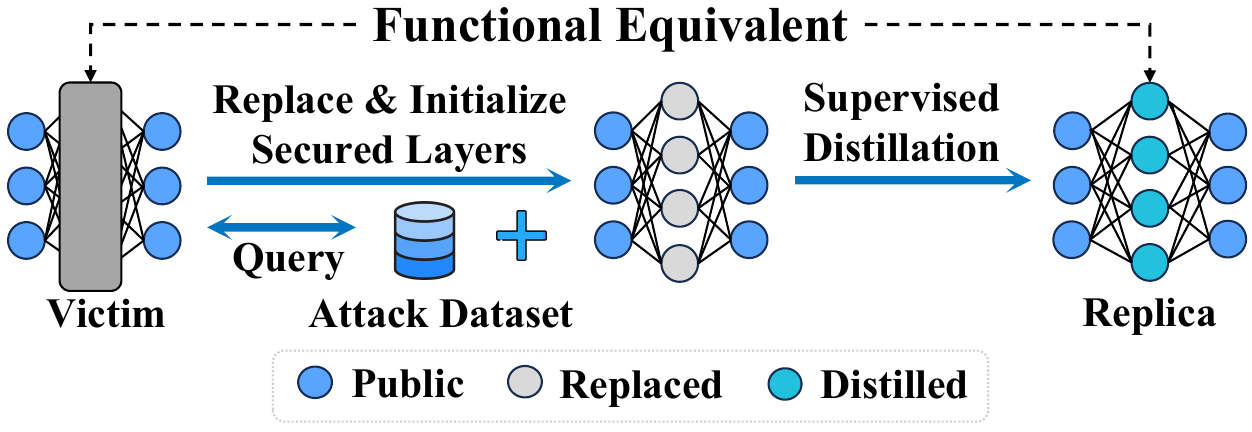}
    \caption{Workflow of model distillation attack}
    \vspace{-0.4cm}
\label{fig:flowchart}
\end{figure}

\textbf{Adversary's Objective.} The adversary aims to replicate the functionality of a semi-open victim LLM, partially secured in a protected environment, by training a substitute model. This replica facilitates white-box analysis to identify vulnerabilities, enhancing black-box attacks on related model families~\citep{sitawarin2024pal}. The agreement between the victim and replica is assessed via accuracy and fidelity on a designated test set.

\textbf{Adversary's Knowledge.}  It is assumed that the adversary knows the architectures of both secured and unsecured modules, as prior work~\cite{gou2021knowledge,boi2024theorymodeldistillation} has shown that using the same architecture as the secured module significantly improves the effectiveness of distillation attacks. However, the adversary knows only the parameters of the unsecured module, while those of the secured module remain unknown.

\textbf{Adversary's Capability.} The adversary is capable of querying the semi-open model, obtaining both the semantic output produced by the complete model and the representation vector generated by the secured module. Utilizing this information, the adversary constructs a distillation attack dataset denoted as $\mathcal{D}$. Since the adversary knows the architecture of the secured module, the adversary next replaces the secured module with a randomly initialized module of the same architecture. Using the constructed set \(\mathcal{D}\), the adversary employs three distinct supervised distillation strategies to replicate the functionality of the secured module:  (1) \textbf{FT-all:} Fine-tunes both the replacement and unsecured modules using output of the entire model as training labels.  (2) \textbf{FT-closed:} Fine-tunes only the replacement model using output of the entire model, keeping the unsecured module fixed.  (3) \textbf{SEM}~\citep{canHideApi}: Fine-tunes the replacement model using outputs from the secured module without involving the unsecured component.

\subsection{Problem Formulation}
\label{subsection2.2}
In this paper, we analyze the performance of a large language model  under a defined distribution $\mathbb{P}_{\mathbf{X}\times Y}$, describing the relationship between input matrix $\mathbf{X}$ and label $Y$. We assume the victim LLM $f(\mathbf{X};\bm{\theta})$ performs well on this distribution, and the attack set $\mathcal{D}$ comprises samples drawn from $\mathbb{P}_{\mathbf{X}\times Y}$. To evaluate agreement between the distilled LLM and ground-truth labels, we use a scoring function $s:\mathcal{Y}\times \mathcal{Y}\rightarrow \mathbb{R}^+$. Secured layers are indexed by $I \subseteq [L] = \{1, \dots, L\}$. Let $\bm{\theta}_{\text{dist}}(I, \mathcal{D})$ represent the parameter vector of the distilled replica of a victim model, where layers indexed by $I$ are secured, and adversaries utilize the attack set $\mathcal{D}$ to replicate its functionality.
For each secured set $I$, we define the "\textbf{Distillation Ratio}" $R(I)$, which quantifies how well the distilled model $\bm{\theta}_{\text{dist}}(I, \mathcal{D})$ replicates the behavior of $f(\mathbf{X};\bm{\theta})$,  expressed as
{\begin{equation}
    R(I)=\frac{\mathbb{E}[s(f(\mathbf{X};\bm{\theta}_{\text{dist}}(I,\mathcal{D})), Y)]}{\mathbb{E}[s(f(\mathbf{X};\bm{\theta}), Y)]}.
\end{equation}}
Here, $\mathbb{E}$ in the numerator reflects the expectation computed over random samples $(\mathbf{X}, Y)$ drawn from $\mathbb{P}_{\mathbf{X}\times Y}$, the random attack set $\mathcal{D}$, and the random initialization of parameters within the secured layers during fine-tuning. Conversely, the term $\mathbb{E}$ in the denominator solely considers the expectation over random samples. With this definition, $R([L])$ represents the distillation ratio when the entire model is secured, reflecting the highest level of security. This leads to the question:
\begin{align*}  
&\text{What is the smallest secured set } I \text{ such that } \\  
& \quad\quad  R(I) \text{ closely approximates } R([L])?  
\end{align*}  
This question aims to identify the minimal secured set \(I\) such that securing the layers indexed by \(I\) achieves a level of security comparable to securing the entire model.

\section{Methodology}
\label{section:Methodology}

\begin{figure}
    \centering
    \includegraphics[width=0.98\linewidth]{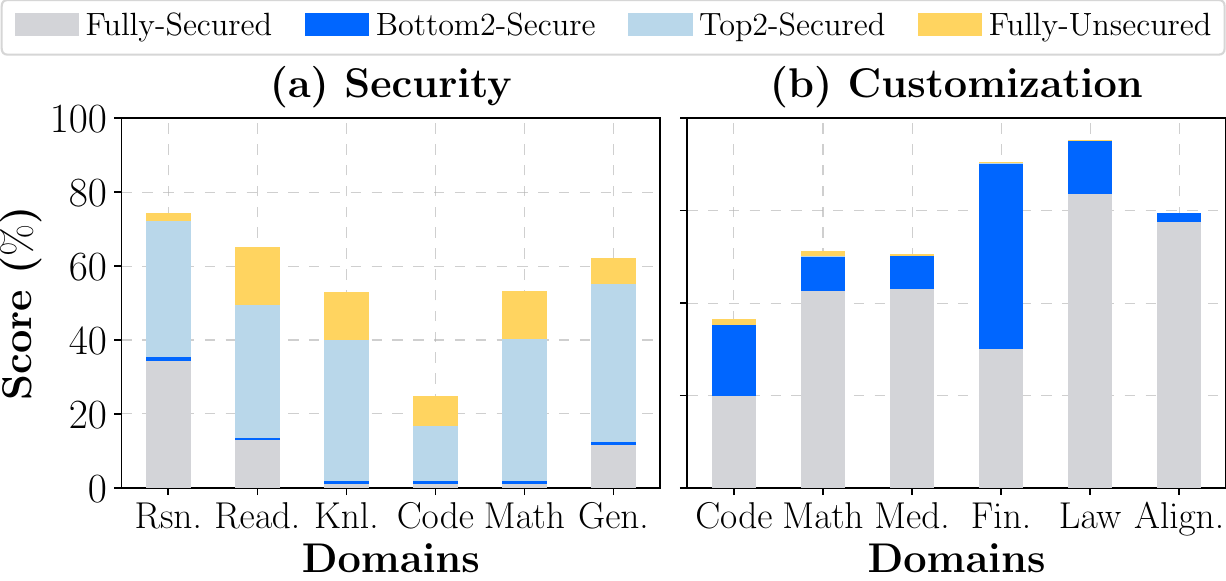}
      \caption{Security and adaptability comparison in Llama2-70B. Lower scores indicate better security in Fig. (a) and weaker adaptability in Fig. (b). Details can be found in Appendix~\ref{append:result:semi-open Llama2-70B}}
      \label{fig:MethodObservation}
      \vspace{-0.3cm}
\end{figure}

In this section, we investigate the impact of securing specific layers on security and customization against distillation attacks. We begin with an experiment with two semi-open deployments of Llama2-70B: one securing the bottom two decoder layers (Bottom2-Secured) and the other securing the top two decoder layers (Top2-Secured). As shown in Figure~\ref{fig:MethodObservation}, both deployments achieve similar customization performance in six downstream tasks. However, securing the bottom layers provides significantly stronger security. Additionally, comparing Bottom2-Secured to fully-secured deployment reveals comparable security with improved customizability. This suggests that securing a certain number of bottom layers can effectively balance strong security against distillation attacks and high customization performance.

\subsection{Security Transition in  Deep Transformers}
\label{subsection:Transition}

\textbf{Model Overview.} In this subsection, we consider a deep transformer $f$ with $L$ layers, expressed as \( f(\mathbf{X}; \bm{\theta}) = \varphi_L \circ \dots \circ \varphi_1(\mathbf{X}) \). The input feature matrix \(\mathbf{X} \in \mathbb{R}^{n \times d}\) consists of \(n\) rows, each representing a \(d\)-dimensional token vector. Each layer \(\varphi_i\) is a transformer that incorporates a normalized residual self-attention mechanism, defined as: 
{\begin{align*}
\resizebox{0.95\linewidth}{!}{$
\varphi_i(\mathbf{X}; K_i, Q_i) =
\mathbf{X} + \operatorname{softmax}\!\left(
\frac{\mathbf{X}Q_i (\mathbf{X}K_i)^\top}{\sqrt{d_Q}\,\|\mathbf{X}\|^2}
\right)\mathbf{X}
$}
\end{align*}}
Here, \(Q_i \in \mathbb{R}^{d \times d_Q}\) and \(K_i \in \mathbb{R}^{d \times d_Q}\) are projection matrices for the query and key components, respectively. The terms \(\sqrt{d_Q}\) and \(\|\mathbf{X}\|\) serve as normalization factors, ensuring stable computations within the attention mechanism.
We consider the semi-open deployment of securing the $\alpha L$-th layer with $\alpha\in[0,1]$ and $\alpha L\in\mathbb{N}$ while keeping other layers unsecured. After the distillation attack, we assume the parameters of the distilled model in the unsecured layers are identical to the victim model, while those in the secured layer deviate. Let $\hat{K}_{\alpha L}$ and $\hat{Q}_{\alpha L}$ denote the distilled weight matrix of the proprietary layer, i.e., ${\bm\theta}_{\text{dist}}(\{\alpha L\})=\{(K_1,Q_1),...,(\hat{K}_{\alpha L},\hat{Q}_{\alpha L}),...,({K}_{L},{Q}_{L})\}$. Let $\hat{\varphi}_{\alpha L}$ denote the function of the distilled proprietary layer, i.e., the $\alpha L-$th layer, in the distilled model. In this subsection, we consider the normalized output of an infinitely deep model whose $\alpha L$-th layer is hidden and subjected to the attack. The output of the distilled model is 
\begin{align*}
    \hat{f}_\infty(\mathbf{X})=\lim_{L\rightarrow \infty}\frac{f(\mathbf{X};{\bm\theta}_{\text{dist}}(\{\alpha L\}))}{\|f(\mathbf{X};{\bm\theta}_{\text{dist}}(\{\alpha L\}))\|_F},
\end{align*}
where $\|\cdot\|_F$ denotes the Frobenius norm. We consider an infinitely deep network as the ideal model, reflecting the sufficient depth of most large-scale models in practice. The following theorem establishes the existence of a critical value \(\alpha^*\) such that if \(\alpha < \alpha^*\), the output matrix of the distilled LLM has rank one. Conversely, if \(\alpha > \alpha^*\), the output matrix has rank strictly greater than one.
\begin{myTheo}
\label{theorem:transition layer}
Assume that $\mathbb{P}_{\mathbf{X}\times Y}$ is defined on a countable domain $\mathcal{X}\times \mathcal{Y}$ with $\mathbf{0}_{n\times d}\notin \mathcal{X}$. Assume that parameter matrices $\{K_i, Q_i\}_{i\ge1}$ in the victim model $f$ have uniform bounded norms, i.e., $\|K_i\|\le D$ and $\|Q_i\|\le D$ for some $D>0$. There exists an $\alpha^*\in(0,1)$ depending on $D$ such that the following two statements are true.

(1) If \(\alpha < \alpha^*\) and \(\{K_i, Q_i\}_{i \geq 1}\) are parameter matrices of the victim model, with \(\hat{K}_{\alpha L}\) and \(\hat{Q}_{\alpha L}\) as distilled parameters drawn from a continuous distribution on \(\mathbb{R}^{n \times d}\), the matrix \(\hat{f}_\infty(\mathbf{X})\) almost surely has rank one for all inputs \(\mathbf{X}\).

(2) If \(\alpha > \alpha^*\), there exists a victim model with parameter sequence \(\{K_i, Q_i\}_{i \geq 1}\) such that for any distilled parameters \(\hat{K}_{\alpha L}\) and \(\hat{Q}_{\alpha L}\), the matrix \(\hat{f}_\infty(\mathbf{X})\) has rank greater than one for some $\mathbf{X}$.
\end{myTheo}
\textbf{Remark 1:} The proof is provided in Appendix~\ref{appendix::thm-proof}. This theorem demonstrates that if the distilled parameters of the bottom layers (i.e., \(\alpha < \alpha^*\)) are obtained through a randomized algorithm, such as stochastic gradient descent, with a continuous distribution supported on $\mathbb{R}^{n \times d}$, the distillation will certainly fail, as the feature matrix degenerate. In contrast, keeping the later layers secured (i.e., \(\alpha > \alpha^*\)) does not maintain this property, indicating that it is more effective to secure the bottom layers before the transition layer, rather than the later ones. Further remarks are in Appendix~\ref{append:explor_remarks}.




\subsection{SOLID: \underline{S}emi-\underline{O}pen \underline{L}ocal \underline{I}nfrastructure \underline{D}eployment Framework}
\label{subsection:SOLID}
Theorem~\ref{theorem:transition layer} shows that securing bottom layers improves security. Inspired by this insight, we propose a method to approximately find the smallest bottom layer index set \( I \) that satisfies \( R(I) \leq (1+\varepsilon)R([L]) \) for any small $\varepsilon>0$. 
To achieve this, a straightforward implementation is to begin with
A simple approach is to start with \( I_l = \{1, \ldots, l\} \) for each \( l \) beginning from 1, then evaluate the distillation ratio \( R(I_l) \) after the attack, and identify the smallest \( l \) that meets the inequality.
This extensive fine-tuning process is time-consuming, prompting the critical question: \textit{Can we create a fine-tuning-free metric that predicts LLM performance under model distillation attacks?} Hence, our goal is to establish a metric directly correlated with the distillation ratio.

In the distillation ratio \( R(I) \), each \( I \) has the same denominator, so our focus is on a metric related to the numerator, specifically \( \mathbb{E}[s(f(\mathbf{X};\bm{\theta}_{\text{FT}}(I,\mathcal{D})), Y)] \), which measures the average performance score of the distilled model. This average performance score generally inversely correlates with the average testing loss with the expression \(L(\bm\theta)\triangleq \mathbb{E}_{\mathbf{X}\times Y}[\ell(f(\mathbf{X};\bm{\theta}), Y)] \), {where $\ell$ denotes the cross-entropy loss employed by LLM}.  Hence, we aim at finding the smallest $I$ such that 
{
\begin{align*}
&L({\bm\theta}_{\text{dist}}(I,\mathcal{D})) \geq (1-\varepsilon)L({\bm\theta}_{\text{dist}}([L],\mathcal{D})).
\end{align*}}
However, calculating both sides of this inequality requires knowing the distilled parameters from the fine-tuning process. To bypass this, we aim for an approximate solution. The distilled parameters are generated through gradient descent, starting from the initial parameters \( \bm{\theta}_0(I) \), with the hidden layers being randomly initialized. Using the Taylor Expansion, we find
{\begin{align*}
L({\bm\theta}_{\text{dist}}(I,\mathcal{D})) &= L({\bm\theta}_{\text{0}}(I,\mathcal{D})) \\
&\quad+ \mathcal{O}(\mathbb{E}\|\bm{\theta}_{\text{dist}}(I,\mathcal{D})-\bm{\theta}_0(I)\|_2).
\end{align*}}
Previous research~\citep{Parameterization,Effects-dataset-size} indicates that the difference \( \|\bm{\theta}_{\text{dist}}(I,\mathcal{D}) - \bm{\theta}_0(I)\|_2 \) is minimal in large networks compared to the dataset size \( |\mathcal{D}| \). In models such as single-layer ReLU networks~\citep{NNtheory,ReLU}, this difference scales as \(\mathcal{O}\left( \frac{|\mathcal{D}|}{\sqrt{N}} \right)\)~\citep{NTK-1,NTK-2}, where \( N \), the number of model parameters, far exceeds the dataset size in large language models (LLMs)~\citep{dubey2024llama3,APISecurity}. The first term, independent of fine-tuning, dominates and effectively predicts the distillation ratio. We refer to this term as the \textbf{Distillation Difficulty} (DD\( (I) \)), defined as
\begin{equation*}
\text{DD}(I) = \mathbb{E}[L(\bm{\theta}_0(I))].
\end{equation*}
This score, which can be estimated using a sample average, represents the distilled model performance of the model when specific layers \( I \) are secured. A higher \textbf{DD\( (I) \)} suggests better security performance, indicating a lower distillation ratio \( R(I) \). Therefore, our SOLID operates in the following way. SOLID begins by sampling evaluation data {targeting general capabilities} from the underlying distribution, and then computes DD\( (I_l) \) for each set of secured layers \( I_l = \{1,...,l\} \) for \( l = 1,...,L \). SOLID stops at the smallest \( l^* \)  that satisfies DD\( (I_{l^*}) \geq (1-\varepsilon) \text{DD}([L]) \). 

\section{Experiments}
\label{section:experiments}
In this section, we conduct experiments to answer the following research questions:
\begin{itemize}[leftmargin=*, itemsep=0pt, parsep=0pt, topsep=0pt]
    \item \textbf{RQ1.} Can query-based distillation attack distill the functionality of the entire model under the baseline deployment that secures the top layer?
    \item \textbf{RQ2.} How do secured layer location and amount affect the security-customization trade-off?
    \item \textbf{RQ3.} Does SOLID offer a better balance between model theft risk and customization performance compared to baseline deployments?
    \item \textbf{RQ4.} How does SOLID optimize this trade-off? Is it effective for both large and small models?
\end{itemize}

\subsection{Experimental Settings}
We begin by introducing our experimental setups. Details can be found in Appendix~\ref{append:Experiment}.

\textbf{Models.} We consider \textbf{five} open-source, decoder-only structured LLMs with various architectures. Specifically, we select Llama2-70B-chat, Llama2-7B-chat~\citep{llama2}, Mistral-7B-v0.1~\citep{mistral}, Phi-2~\citep{phi3}, and Phi-1.5~\citep{phi1-5}. We designate these pre-trained models as the base models for adaptation and victims in model distillation attacks. 

\textbf{Attack Methods.} We distill models produced by different protection approaches using three attack methods: FT-all, FT-closed and SEM. Following~\citep{bert-vulnerable}, a diverse attack set is required for full distillation. Therefore, we merge data evenly from two general datasets, MMLU benchmark~\citep{mmlu} and Alpaca 52k~\citep{alpacaDataset}, resulting in a 51k combined set.  Additionally, we build four larger general datasets (100k–500k) to strengthen the attack.

\textbf{Baselines.} We compare SOLID with three baselines: SAP-DP, the fully-secured approach~\citep{Semiopen1}, and DarkneTZ~\citep{mo2020darknetz}. The SAP~\citep{SAP} framework exposes the first six decoder layers and secures the rest. SAP-DP extends SAP by adding Laplace noise to model outputs to enhance protection~\citep{SAP-DP-supp}. The fully-secured approach represents the extreme, securing all layers for maximal security, while DarkneTZ protects only the final decoder layer.

{\textbf{Implementation Details of SOLID.}}
We apply the SOLID algorithm to identify the smallest secure set \( I \) such that \( R(I) \leq (1+\varepsilon)R([L]) \). To calculate distillation difficulty (DD), we use cross-entropy loss and approximate the expectation over samples distributed on the general domain and randomly initialized secured parameters. This is done using a 1,500-sample evaluation set randomly sampled from the MMLU benchmark and Alpaca 52k, with secured parameters initialized via Xavier initialization and averaged over three random seeds (20, 42, 1234). In our experiments, we find that \(\varepsilon = 0.05\) yields optimal performance. 

\textbf{Evaluation Benchmarks} We assess adaptability on six downstream tasks: Code~\citep{CodeInstruct}, Math~\citep{mathInstruct},  Medical~\citep{medicalInstruct}, Finance~\citep{FinanceInstruct}, Law~\citep{guha2024legalbench}, and Alignment~\citep{simpo}. 
To fully evaluate recovered functionalities, we focus on six capabilities domains following Llama2 report~\citep{llama2}. Specifically, we assess the recovered model across \textbf{sixteen} benchmarks grouped into (1) \textit{Commonsense Reasoning} (Rsn.); (2) \textit{Reading Comprehension} (Read.); (3) \textit{World Knowledge} (Knl.); (4) \textit{Code}; (5) \textit{Math}; and (6) \textit{General Ability} (Gen.). 

\textbf{Metrics.} We measure customization through model's improvements on benchmarks. For security, we calculate the ``Average Distillation Ratio'' (ADR) by averaging the distillation ratios across benchmarks. A lower ADR indicates higher security offered by the secure set. 

\begin{table*}[t]
    \centering
    \setlength{\tabcolsep}{5pt}
    \renewcommand{\arraystretch}{0.85} 
    \begin{tabular}{@{}lccccc@{}}
    \toprule
    \textbf{\footnotesize } & \textbf{\footnotesize Benchmark}  & \textbf{\footnotesize Llama2-70B} & \textbf{\footnotesize Llama2-7B} & \textbf{\footnotesize Mistral-7B}  & \textbf{\footnotesize Phi-2}\\ 
    \midrule
    \multirow{5}{*}{ \textbf{\small Rsn.}}& \footnotesize PIQA & \footnotesize 62.6$|$59.8$|$63.0$|$\underline{99.3} & \footnotesize 64.7$|$64.7$|$64.6$|$\underline{99.1} & \footnotesize 63.0$|$61.2$|$60.2$|$\underline{92.2} & \footnotesize 68.3$|$65.6$|$65.7$|$\underline{99.1}  \\
    \footnotesize & \footnotesize Winogrande & \footnotesize 68.5$|$67.7$|$68.3$|$\underline{98.3} & \footnotesize 76.8$|$74.8$|$76.6$|$\underline{100.} & \footnotesize 67.2$|$69.0$|$68.3$|$\underline{89.5} & \footnotesize 68.3$|$64.9$|$64.8$|$\underline{99.1}  \\
    \footnotesize  & \footnotesize ARC-easy & \footnotesize 31.9$|$32.8$|$31.3$|$\underline{98.5} &\footnotesize 36.3$|$35.5$|$34.9$|$\underline{97.6} & \footnotesize 32.3$|$34.7$|$32.0$|$\underline{86.6} & \footnotesize 43.2$|$35.3$|$33.9$|$\underline{99.5}  \\
    \footnotesize & \footnotesize ARC-challenge & \footnotesize 38.5$|$38.1$|$44.2$|$\underline{99.2} & \footnotesize 47.8$|$46.6$|$50.9$|$\underline{100.} & \footnotesize 39.7$|$42.6$|$44.5$|$\underline{81.4} & \footnotesize 36.8$|$36.6$|$35.3$|$\underline{99.5}  \\
    \footnotesize & \footnotesize Hellaswag & \footnotesize 31.4$|$31.4$|$32.4$|$\underline{98.1} & \footnotesize 33.9$|$34.0$|$35.0$|$\underline{96.6} & \footnotesize 32.2$|$32.0$|$31.3$|$\underline{84.6} & \footnotesize 37.4$|$37.3$|$34.3$|$\underline{96.5}  \\
    \midrule
    \multirow{4}{*}{\small \textbf{Read.}} & \footnotesize LAMBADA & \footnotesize 0.01$|$0.00$|$0.00$|$\underline{88.6} & \footnotesize 0.02$|$0.00$|$0.01$|$\underline{92.2} & \footnotesize 0.16$|$0.00$|$0.01$|$\underline{67.9} & \footnotesize 1.34$|$0.04$|$0.00$|$\underline{94.6}  \\
    \footnotesize & \footnotesize BoolQ & \footnotesize 47.2$|$47.1$|$53.9$|$\underline{100.} & \footnotesize 59.5$|$56.0$|$65.0$|$\underline{99.6} & \footnotesize 48.3$|$46.8$|$56.7$|$\underline{97.3} & \footnotesize 56.7$|$50.3$|$55.8$|$\underline{100.}  \\
    \footnotesize & \footnotesize SQuADv2 & \footnotesize 1.50$|$1.68$|$0.34$|$\underline{55.3} & \footnotesize 0.68$|$0.88$|$0.82$|$\underline{59.5} & \footnotesize 1.69$|$0.36$|$0.93$|$\underline{50.7} & \footnotesize 3.65$|$0.39$|$0.90$|$\underline{62.9}  \\
    \footnotesize & \footnotesize OBQA & \footnotesize 54.5$|$54.5$|$57.1$|$\underline{99.6}  & \footnotesize 57.4$|$52.5$|$59.2$|$\underline{94.8} & \footnotesize 57.7$|$56.8$|$56.3$|$\underline{84.0} & \footnotesize 0.00$|$0.00$|$0.02$|$\underline{94.3}  \\
    \midrule
    \multirow{2}{*}{\small \textbf{Knl.}}  & \footnotesize NaturalQuestions & \footnotesize  0.00$|$0.02$|$0.00$|$\underline{40.1} & \footnotesize 0.01$|$0.01$|$0.08$|$\underline{53.6}  & \footnotesize 0.00$|$0.00$|$0.02$|$\underline{31.8} & \footnotesize 0.01$|$0.00$|$0.06$|$\underline{87.4} \\
    
    \footnotesize & \footnotesize TriviaQA & \footnotesize 0.00$|$0.02$|$0.00$|$\underline{72.3} & \footnotesize 0.00$|$0.00$|$0.03$|$\underline{73.8} & \footnotesize 0.00$|$0.00$|$0.01$|$\underline{38.7} & \footnotesize 0.01$|$0.00$|$0.01$|$\underline{68.9} \\
    \midrule
    \multirow{1}{*}{\small \textbf{Code}} & \footnotesize MBPP\&H.E. & \footnotesize 0.00$|$0.00$|$0.00$|$\underline{58.6} & \footnotesize 0.00$|$0.00$|$0.00$|$\underline{90.9} & \footnotesize 0.00$|$0.00$|$0.00$|$\underline{40.2} & \footnotesize 0.00$|$0.00$|$0.00$|$\underline{91.1}  \\
    \midrule
    \footnotesize \textbf{Math} & \footnotesize GSM8K & \footnotesize 0.02$|$0.00$|$0.06$|$\underline{79.6} & \footnotesize 0.00$|$0.00$|$0.00$|$\underline{78.6} & \footnotesize 0.00$|$0.00$|$0.00$|$\underline{31.1} & \footnotesize 0.00$|$0.00$|$0.00$|$\underline{86.2} \\
    \midrule
    \multirow{2}{*}{\small \textbf{Gen.}} &\footnotesize MMLU & \footnotesize \footnotesize 36.8$|$38.3$|$36.5$|$\underline{96.7} & \footnotesize 52.9$|$50.0$|$53.3$|$\underline{110.} & \footnotesize 40.4$|$36.9$|$37.2$|$\underline{81.7} & \footnotesize 42.6$|$40.3$|$40.5$|$\underline{99.5}  \\
    \footnotesize & \footnotesize BBH & \footnotesize 0.00$|$0.00$|$0.00$|$\underline{93.3} & \footnotesize 0.00$|$0.00$|$0.00$|$\underline{101.} & \footnotesize 0.00$|$0.00$|$0.00$|$\underline{63.3} & \footnotesize 0.01$|$0.00$|$0.00$|$\underline{94.8}  \\
    \midrule
    \multicolumn{2}{c}{\footnotesize \textbf{Average Distillation Ratio($\downarrow$)}} & \footnotesize \green{\textbf{21.9}}$|$21.8$|$22.8$|$\red{\underline{\textbf{77.9}}} & \footnotesize \green{\textbf{25.3}}$|$24.4$|$25.9$|$\red{\underline{\textbf{86.5}}}   & \footnotesize \green{\textbf{22.5}}$|$22.4$|$22.8$|$\red{\underline{\textbf{73.7}}} & \footnotesize \green{\textbf{23.9}}$|$22.3$|$22.4$|$\red{\underline{\textbf{88.9}}} \\
    \multicolumn{2}{c}{\footnotesize \textbf{Secured Ratio($\downarrow$)}} & \footnotesize \green{\textbf{2.50}}$|$92.5$|$100.$|$\red{\underline{\textbf{1.25}}} & \footnotesize \green{\textbf{3.16}}$|$81.3$|$100.$|$\red{\textbf{\underline{3.16}}}   & \footnotesize \green{\textbf{3.16}}$|$81.3$|$100.$|$\red{\textbf{\underline{3.16}}} & \footnotesize \green{\textbf{6.25}}$|$81.3$|$100.$|$\red{\textbf{\underline{3.16}}}\\
    \bottomrule
    \end{tabular}
    \caption{Distillation ratios across six functionalities under FT-all (SOLID$|$SAP-DP$|$Fully-secured$|$\underline{DarkneTZ}). ``H.E.'' in the Code domain denotes the benchmark HumanEval. Green and red indicate the overall best- and worst-performing methods, respectively. Additional results are provided in Appendix~\ref{append:result:Table1}.} 
    \label{tab:recover-strategy1}
    \vspace{-0.3cm}
\end{table*}

\subsection{Failure in Defense (RQ1)}
We evaluate security of DarkneTZ using three distillation strategies. Based on the results shown in Tables~\ref{tab:recover-strategy1} and \ref{tab:recover-strategy2}, we have following observations.

\textbf{Obs1: DarkneTZ, which secures only the last decoder layer, fails to protect the model against all three attacks.} As shown in Table~\ref{tab:recover-strategy1}, DarkneTZ achieves ADRs generally exceeding 73\%. Notably, on Llama2-7B, it surpasses 100\% distillation ratio on the MMLU and BBH datasets, indicating that the distilled model outperforms the original on these tasks. Similarly, Table~\ref{tab:recover-strategy2} highlights consistent failure patterns against FT-closed and SEM attacks, with DarkneTZ maintaining ADRs above 75\%, demonstrating the ability of these strategies to recover significant model functionality.

\subsection{Security-Customization Trade-off (RQ2)} 
\label{subsection:sensitivity}

\begin{figure}[t]
    \centering
    \includegraphics[width=0.95\linewidth]{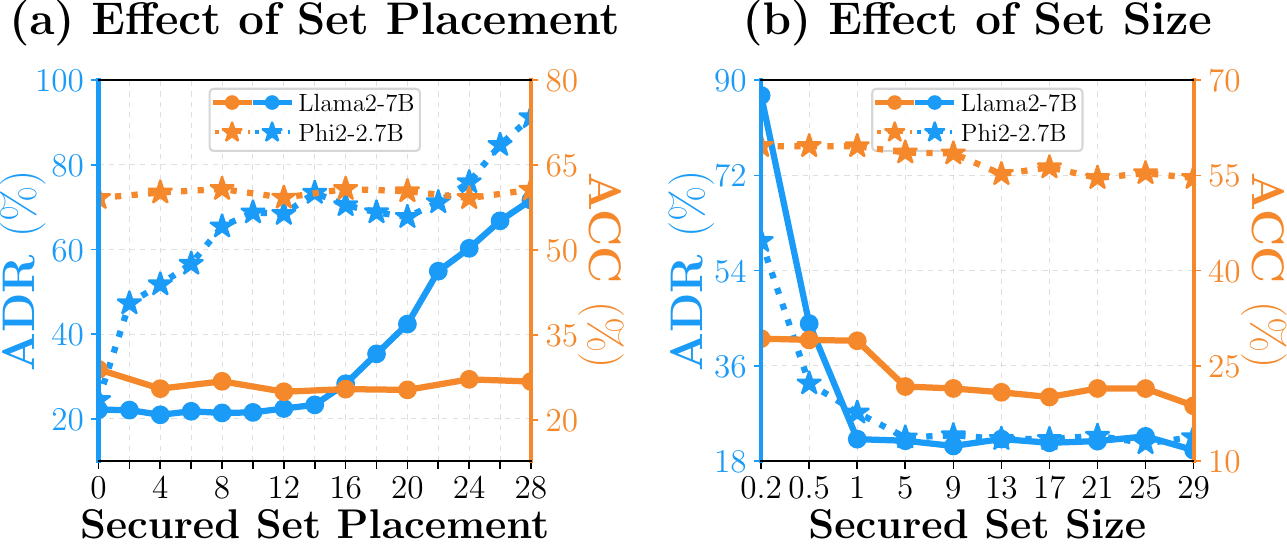}
    \caption{(a) shows the trade-off between security and customization for Llama2-7B and Phi-2 with different placements of same-sized secured sets. (b) shows the trade-off as the secured set size increases from the first decoder layer. Smaller ADR indicates higher security and higher ACC reflects better customizability.}
    \vspace{-0.5cm}
    \label{fig:Combination-resilienceClosedset}
\end{figure}

We conduct two experiments to analyze the impact of secured layer placement and quantity on the trade-off between security and customization. First, we secure one layer in Llama2-7B and two in Phi-2, varying their placement. Second, we incrementally secure both models by adding protected layers, starting from the smallest module ($\texttt{k\_project}$) of the first decoder layer. These models are evaluated under the FT-all distillation attack and customized for the math domain. The results, as shown in Figure~\ref{fig:Combination-resilienceClosedset}, lead to the following observations.

\textbf{\textbf{Obs2: Secured layer placement significantly impacts security, consistent with Theorem~\ref{theorem:transition layer}, but has small effect on customization performance.}} As shown in Figure~\ref{fig:Combination-resilienceClosedset}(a), for Llama2-7B, security transitions at the fourteenth layer, with ADR consistently near 20\% for earlier sets, indicating stronger security than protecting later layers. Meanwhile, customization accuracy remains stable across placements, highlighting the advantage of securing pre-transition layers. In contrast, Phi-2 transitions earlier at the first layer set, where only the first set balances security and customization, with later sets reducing security. These results suggest that securing layers before the transition layer optimizes the security-customization trade-off. Results for Mistral-7B and Phi-1.5 are in Appendix~\ref{append::Experiment transition layer}.

\textbf{Obs3: Increasing the number of secured layers enhances security but reduces customization.} As shown in Figure~\ref{fig:Combination-resilienceClosedset}(b), the ADR of Llama2-7B decreases from 85\% to 22\% after securing an entire decoder layer, indicating improved security. However, customization accuracy drops from 29\% to 21\% as the number of secured layers increases from one to five, reflecting reduced customization flexibility. A similar trend is observed in Phi-2, suggesting that while increasing the number of secured layers enhances security (lower ADR), it negatively impacts customization flexibility (lower ACC) in both models. Further details are in Appendix~\ref{append:Security Across Secure Sizes.}.

\begin{table}[t]
    \centering
    \setlength\tabcolsep{2.8pt} 
    \centering
    \renewcommand{\arraystretch}{0.75} 
    \begin{tabular}{@{}llcccccc@{}}
    \toprule
    \textbf{\small Strat.} & \textbf{\small Method} & \textbf{\footnotesize Rsn.} & \textbf{\footnotesize Read.} & \textbf{\footnotesize Knl.} & \textbf{\footnotesize C.\&M.} &  \textbf{\footnotesize Gen.} & \textbf{\footnotesize ADR}\\ 
    \midrule
    \multirow{4}{*}{ \textbf{\footnotesize FT-c.}} & \footnotesize SOLID  & \footnotesize 47.1 & \footnotesize 21.6 & \footnotesize 0.00 & \footnotesize 0.03  & \footnotesize 18.7 & \footnotesize 22.6\\
    & \footnotesize SAP-DP & \footnotesize 46.2 & \footnotesize 19.5 & \footnotesize 0.00 & \footnotesize 0.00  & \footnotesize 19.0 & \footnotesize 21.8\\
    & \footnotesize F-Secured  & \footnotesize 47.8 & \footnotesize 21.2 & \footnotesize 0.00 & \footnotesize 0.08  & \footnotesize 18.5 & \footnotesize 22.8\\
    & \footnotesize DarkneTZ  & \footnotesize 98.7 & \footnotesize 69.3 & \footnotesize 58.3 & \footnotesize 65.9  & \footnotesize 95.0 & \footnotesize 78.1\\
    \midrule
    \multirow{4}{*}{ \textbf{\footnotesize SEM}} & \footnotesize SOLID & \footnotesize 48.2 & \footnotesize 21.9 & \footnotesize 0.00 & \footnotesize 0.00  & \footnotesize 18.5 & \footnotesize 22.4\\
    & \footnotesize SAP-DP & \footnotesize 47.1 & \footnotesize 21.1 & \footnotesize 0.00& \footnotesize 0.00  & \footnotesize 18.3 & \footnotesize 22.3\\
    & \footnotesize F-Secured  & \footnotesize 47.8 & \footnotesize 21.2 & \footnotesize 0.00 & \footnotesize 0.08  & \footnotesize 18.5 & \footnotesize 22.8\\
    & \footnotesize DarkneTZ  & \footnotesize 98.8 & \footnotesize 71.2 & \footnotesize 54.2 & \footnotesize 66.3  & \footnotesize 94.1 & \footnotesize 77.4\\
    \bottomrule
    \end{tabular}
    \caption{Distillation ratios of Llama2-70B under FT-closed and SEM attacks.} 
    \label{tab:recover-strategy2}
    \vspace{-0.3cm}
\end{table}

\subsection{Effectiveness of SOLID (RQ3)}\label{subsection:mainresults}

We compare the security of SOLID with baseline deployments across three distillation strategies. The results lead to the following observations.

\textbf{Obs4: SOLID offers comparable security against model distillation to the highest level of protection (fully-secured), while securing significantly fewer parameters.} As shown in Table~\ref{tab:recover-strategy1}, SOLID achieves a similar security level (ADR) to SAP-DP and the fully-secured approach across four architectures and various domains, while securing at most 6.25\% of parameters, compared to at least 80\% for the others. For example, on Llama2-70B, SOLID secures only 1.25\% of parameters yet achieves an ADR of 21.9\%, comparable to SAP-DP (21.8\%) and the fully-secured approach (22.8\%), which protect 92.5\% and 100\% of parameters, respectively.
Furthermore, under FT-closed and SEM attacks, SOLID also matches the security level provided by SAP-DP and the fully-secured approach. Table~\ref{tab:recover-strategy2} shows that under FT-closed attack, the ADR differences between SOLID, SAP-DP, and the fully-secured approach remain below 2.1\% across six domains. Similarly, under SEM attack, the distillation ratios closely aligned with the other two approaches. These results confirm that SOLID effectively protects against distillation attacks while securing significantly fewer parameters. More details are in Appendix~\ref{appen::SOLID-Baselines-DATA} and \ref{appen::reovery-strategies-DATA}.

\begin{table}[t]
    \centering
    \small
    \centering
    \setlength{\tabcolsep}{3pt}
    \begin{tabular}{@{}lccccccc@{}}
    \toprule
    \textbf{\small Scale} & \textbf{\footnotesize Rsn.} & \textbf{\footnotesize Read.} & \textbf{\footnotesize Knl.} & \textbf{\footnotesize C.\&M.} &  \textbf{\footnotesize Gen.} & \textbf{\footnotesize ADR} & \textbf{\footnotesize ADR-Da.}\\ 
    \midrule
    \footnotesize 51k  & \footnotesize 51.7 & \footnotesize 21.6 & \footnotesize 0.01 & \footnotesize 0.00  & \footnotesize 28.3 & \footnotesize 25.3 & \footnotesize 86.5\\
    \footnotesize 100k & \footnotesize 51.3 & \footnotesize 21.5 & \footnotesize 0.13 & \footnotesize 0.00  & \footnotesize 29.6 & \footnotesize 25.3 & \footnotesize 89.1\\
    \footnotesize 200k & \footnotesize 51.4 & \footnotesize 21.7 & \footnotesize 0.11 & \footnotesize 0.00  & \footnotesize 29.7 & \footnotesize 25.2 & \footnotesize 91.3\\
    \footnotesize 300k & \footnotesize 51.6 & \footnotesize 21.7 & \footnotesize 0.11 & \footnotesize 0.00  & \footnotesize 30.5 & \footnotesize 25.5 & \footnotesize 94.5\\
    \footnotesize 500k & \footnotesize 51.8 & \footnotesize 22.0 & \footnotesize 0.09 & \footnotesize 0.00  & \footnotesize 30.8 & \footnotesize 25.8 & \footnotesize 96.9\\
    \bottomrule
    \end{tabular}
    \caption{SOLID vs. Dataset scales. ADR-Da. represents the ADR by DarkneTZ. Details are in Appendix~\ref{appendix:datasets scales}.}
    \label{tab:data-length}
\end{table}

\textbf{Obs5: The security of SOLID cannot be easily compromised by simply increasing the dataset scale.}  As shown in Table~\ref{tab:data-length}, the distillation ratios for SOLID increase marginally with larger datasets, showing only a 0.5\% ADR rise when scaling from 51k to 500k samples. In contrast, DarkneTZ exhibits a significant increase in the ADR, from \(86.5\%\) to \(96.9\%\), over the same dataset size range. This highlights the robustness of SOLID's security against increasing attack dataset sizes. Details of the attack datasets are provided in Appendix~\ref{append:Extra Distillation Datasets}.


\begin{figure}[t]
  \centering
  \includegraphics[width=0.95\linewidth]{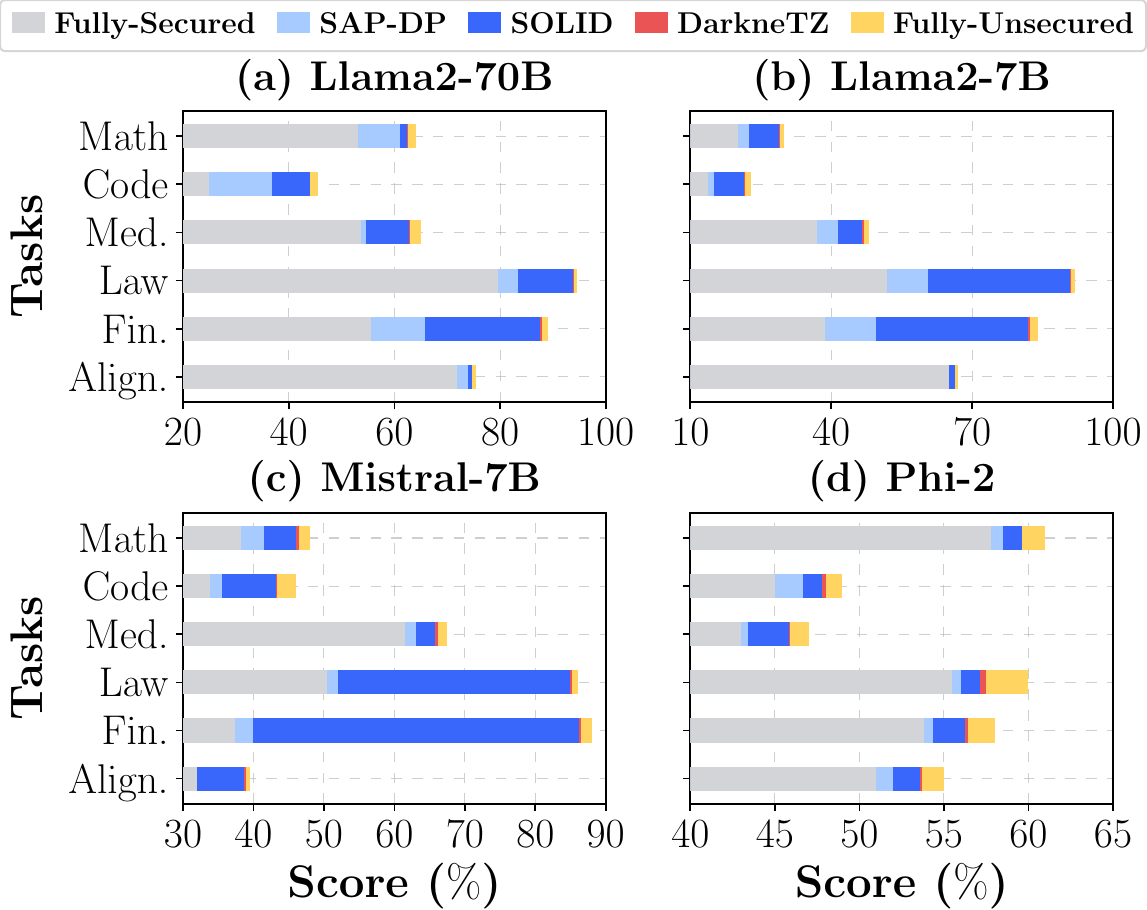}
  \caption{Customization performance comparison of secured models on six downstream tasks. }
  \label{fig:EXP-customizability}
  \vspace{-0.2cm}
\end{figure}

\begin{figure}[t]
    \centering
    \includegraphics[width=0.95\linewidth]{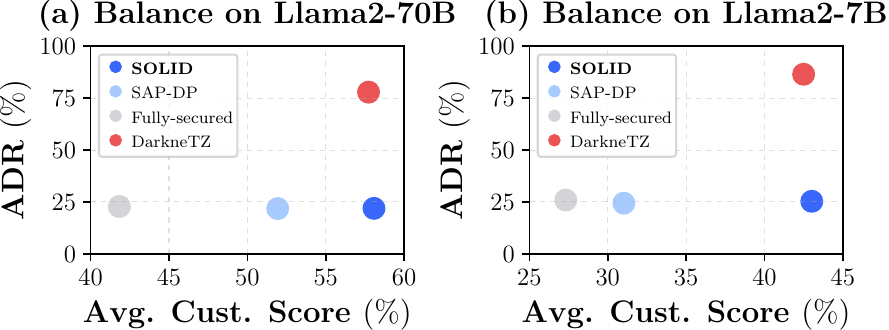}
    \caption{ADRs vs. average customization score. Points closer to the bottom-right indicate better balance.}
    \label{fig:balances}
    \vspace{-0.1cm}
\end{figure}

\textbf{Obs6: SOLID consistently outperforms baseline deployments in customization while achieving security levels comparable to fully-secured approaches. Its customization performance closely matches full parameter fine-tuning.} As shown in Figure~\ref{fig:EXP-customizability}, SOLID improves scores in the Law domain by 10\% over SAP-DP and fully-secured methods on Llama2-70B, and by 35\% on 7B models. Similar trends are observed on Phi-2, though the gain in Law narrows to 1\%. Additionally, the performance of SOLID consistently matches the performance of full parameter fine-tuning across four architectures, with differences within 4\%. This indicates that securing a small subset of parameters preserves customization while ensuring strong protection against distillation attacks. Further results are in Appendix~\ref{append:Customization Training Set} and~\ref{append:result:custom performance}.

We summarize the security and customization performance of each deployment in Figure~\ref{fig:balances}. SOLID achieves an optimal balance between distillation prevention and customization, outperforming other baselines. In the next subsection, we discuss how the distillation difficulty metric optimizes the security-customization trade-off.


\begin{figure}[t]
    \centering
    \includegraphics[width=0.98\linewidth]{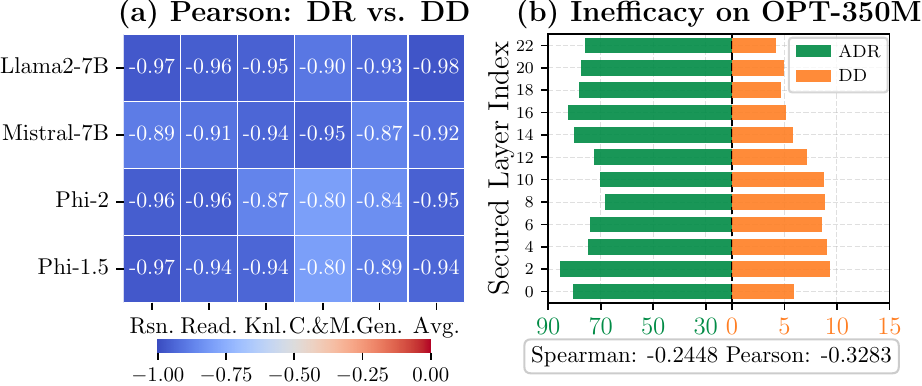}
    \vspace{-0.2cm}
    \caption{(a) presents the Pearson coefficient between distillation ratio (DR) and distillation difficulty (DD) across four models and six domains. (b) depicts the link between ADR and DD for Llama2-7B and OPT-350M.}
    \label{fig:Combined_correlation}
    \vspace{-0.2cm}
\end{figure}
\subsection{Discussion on DD (RQ4)}
\label{subsection:discussion}
We assess the efficacy of distillation difficulty (DD) in estimating distilled model performance by calculating the Pearson and Spearman correlation coefficients between DD and ADR across different domains. To address uncertainties regarding the effectiveness of DD in shallow transformers, we secure and attack 2-layer secured sets in OPT-350M~\citep{opt-350}, which has 350M parameters. Based on the results shown in Figure~\ref{fig:Combined_correlation}, we have following observations.

\textbf{Obs7: DD is effective in larger models, with a clear negative correlation between DD and average distillation ratios.} As shown in Figure~\ref{fig:Combined_correlation} (a), the Pearson coefficient for Llama2-7B consistently remains below -0.80, reaching as low as -0.98. We also observe similar phenomena in other models with varying architectures and sizes, confirming DD as a reliable predictor of distilles model performance and the effectiveness of SOLID. Results of Spearman coefficients are in Appendix~\ref{append::RS-ADR-coefficient}.


\textbf{Obs8: DD is ineffective in smaller OPT model, with notable inconsistencies with ADRs.} As shown in Figure~\ref{fig:Combined_correlation} (b), DD exhibits weak negative correlation with ADR in OPT-350M (coefficients > -0.33), showing its unsuitability for predicting distillation performance. Additionally, optimal security is achieved by protecting the middle layers rather than the initial or output layers, making SOLID unable to identify the smallest secured set. Further details are provided in Appendix~\ref{appendix:result:opt}.

\section{Related Works}
\label{section:relatedworks}

\textbf{On-premises deployment.} 
Using LLM services for customization poses significant privacy risks, as user data may be exposed during transmission, storage, and processing~\cite{li2024llm}. To mitigate this, privacy-sensitive sectors require \textit{on-premises deployment} of LLMs, which retains both data and models within their local infrastructure~\cite{schillaci2024site, nevo2024securing}. However, this shifts security risks to vendors, who lose control over model use and face increased threats of theft, especially from hardware- and communication-based attacks on GPUs~\cite{nayan2024sok, rakin2022deepsteal}. To secure models locally, hardware-based protections such as TrustZone have been proposed~\cite{pinto2019demystifying, zhanggroupcover, li2024teeslice}, but they are resource-intensive with limited flexibility~\cite{mo2020darknetz}. A more adaptable approach is \textit{layer-wise security}, which protects only selected layers~\cite{lin2024splitlearning6gedge, chen2024adaptivelayersplittingwireless, zhang2024no}. While prior work suggests securing shallow~\cite{elgamal2020serdab}, intermediate~\cite{shen2022soter}, or output layers~\cite{10446717}, most studies focus on smaller models. Our results show that securing a few, well-chosen bottom layers of LLMs can enhance security while preserving fine-tuning flexibility for on-premises deployment.

\textbf{Model Distillation Attacks.}
Model distillation attacks allow adversaries to replicate model functionality using only black-box access, a process also known as functional extraction~\citep{nevo2024securing, xu2024survey, ezzeddine2024knowledge}. While distillation attacks have been extensively studied in smaller models, such as CNNs~\cite{orekondy2018knockoffnetsstealingfunctionality}, BERT~\cite{DistilBERT, greybox}, and ReLU-based models~\cite{canales2024,jagielski2020highaccuracyhighfidelity}, their effectiveness against LLMs remains an open question. Our work extends these attacks to Llama2-70B and demonstrates that securing only the output layer remains insufficient to prevent near-complete functionality replication.


\section{Conclusion}
\label{section:conclusion}
In this paper, we explore minimal secured sets to protect LLMs from query-based distillation attacks while preserving customization flexibility in on-premises deployments. We find that (1) distillation attacks targeting the secured top layer can successfully replicate the victim model, and (2) both the placement and number of secured layers introduce a security-customization trade-off. Based on these insights, we propose SOLID, a theoretically inspired deployment that optimizes this trade-off. Through extensive experiments, we show that SOLID balances security and customization effectively, outperforming baseline deployments, though it also has certain limitations.


\section*{Limitations}
While our method effectively defends against distillation attacks and preserves model customization, it does not address other adversarial attacks in the black-box setting, such as membership inference attacks (MIA), as demonstrated in the Appendix~\ref{append::Adversarial Attack}. To the best of our knowledge, this is the first work to explore a semi-open deployment framework for LLMs. However, the current algorithm still performs identification at the layer level and does not delve into the impact of different submodules within decoder layers on model security. Furthermore, our proposed metric exhibits reduced effectiveness when applied to smaller models, as discussed in Section~\ref{subsection:discussion}. In future work, we aim to address these limitations by enhancing both the algorithm and the evaluation metric to improve the overall effectiveness of SOLID.

\section*{Acknowledgements}
We are deeply grateful to the anonymous reviewers for their insightful comments and constructive suggestions, which significantly improved this work. We also acknowledge the support from the National Natural Science Foundation of China (No. 62306179, No.62472247 and No. 12326608); the National Key Research and Development Program of China(2024YFB3310000); Hetao Shenzhen-Hong Kong Science and Technology Innovation Cooperation Zone Project (No.HZQSWS-KCCYB-2024016); Guangdong Provincial Key Laboratory of Mathematical Foundations for Artificial Intelligence (2023B1212010001) and the Scientific Research Program of Shanghai Municipal Science and Technology Commission (24BC3200100).

\bibliography{custom}

\begin{thebibliography}{96}
\providecommand{\natexlab}[1]{#1}

\bibitem[{Abdin et~al.(2024)Abdin, Jacobs, Awan, Aneja, Awadallah, Awadalla, Bach, Bahree, Bakhtiari, Behl et~al.}]{phi3}
Marah Abdin, Sam~Ade Jacobs, Ammar~Ahmad Awan, Jyoti Aneja, Ahmed Awadallah, Hany Awadalla, Nguyen Bach, Amit Bahree, Arash Bakhtiari, Harkirat Behl, and 1 others. 2024.
\newblock Phi-3 technical report: A highly capable language model locally on your phone.
\newblock \emph{arXiv preprint arXiv:2404.14219}.

\bibitem[{Anthony et~al.(1999)Anthony, Bartlett, Bartlett et~al.}]{NNtheory}
Martin Anthony, Peter~L Bartlett, Peter~L Bartlett, and 1 others. 1999.
\newblock \emph{Neural network learning: Theoretical foundations}, volume~9.
\newblock cambridge university press Cambridge.

\bibitem[{Austin et~al.(2021)Austin, Odena, Nye, Bosma, Michalewski, Dohan, Jiang, Cai, Terry, Le et~al.}]{mbpp}
Jacob Austin, Augustus Odena, Maxwell Nye, Maarten Bosma, Henryk Michalewski, David Dohan, Ellen Jiang, Carrie Cai, Michael Terry, Quoc Le, and 1 others. 2021.
\newblock Program synthesis with large language models.
\newblock \emph{arXiv preprint arXiv:2108.07732}.

\bibitem[{Bailly et~al.(2022)Bailly, Blanc, Francis, Guillotin, Jamal, Wakim, and Roy}]{Effects-dataset-size}
Alexandre Bailly, Corentin Blanc, {\'E}lie Francis, Thierry Guillotin, Fadi Jamal, B{\'e}chara Wakim, and Pascal Roy. 2022.
\newblock Effects of dataset size and interactions on the prediction performance of logistic regression and deep learning models.
\newblock \emph{Computer Methods and Programs in Biomedicine}, 213:106504.

\bibitem[{Ben~Allal et~al.(2022)Ben~Allal, Muennighoff, Kumar~Umapathi, Lipkin, and von Werra}]{bigcode-evaluation-harness}
Loubna Ben~Allal, Niklas Muennighoff, Logesh Kumar~Umapathi, Ben Lipkin, and Leandro von Werra. 2022.
\newblock A framework for the evaluation of code generation models.
\newblock \url{https://github.com/bigcode-project/bigcode-evaluation-harness}.

\bibitem[{Bisk et~al.(2020)Bisk, Zellers, Bras, Gao, and Choi}]{PIQA}
Yonatan Bisk, Rowan Zellers, Ronan~Le Bras, Jianfeng Gao, and Yejin Choi. 2020.
\newblock Piqa: Reasoning about physical commonsense in natural language.
\newblock In \emph{Thirty-Fourth AAAI Conference on Artificial Intelligence}.

\bibitem[{Boix-Adsera(2024)}]{boi2024theorymodeldistillation}
Enric Boix-Adsera. 2024.
\newblock \href {https://arxiv.org/abs/2403.09053} {Towards a theory of model distillation}.
\newblock \emph{Preprint}, arXiv:2403.09053.

\bibitem[{Canales-Mart{\'\i}nez et~al.(2024)Canales-Mart{\'\i}nez, Ch{\'a}vez-Saab, Hambitzer, Rodr{\'\i}guez-Henr{\'\i}quez, Satpute, and Shamir}]{canales2024}
Isaac~A Canales-Mart{\'\i}nez, Jorge Ch{\'a}vez-Saab, Anna Hambitzer, Francisco Rodr{\'\i}guez-Henr{\'\i}quez, Nitin Satpute, and Adi Shamir. 2024.
\newblock Polynomial time cryptanalytic extraction of neural network models.
\newblock In \emph{Annual International Conference on the Theory and Applications of Cryptographic Techniques}, pages 3--33. Springer.

\bibitem[{Carlini et~al.(2024)Carlini, Paleka, Dvijotham, Steinke, Hayase, Cooper, Lee, Jagielski, Nasr, Conmy et~al.}]{carlini2024stealing}
Nicholas Carlini, Daniel Paleka, Krishnamurthy~Dj Dvijotham, Thomas Steinke, Jonathan Hayase, A~Feder Cooper, Katherine Lee, Matthew Jagielski, Milad Nasr, Arthur Conmy, and 1 others. 2024.
\newblock Stealing part of a production language model.
\newblock \emph{arXiv preprint arXiv:2403.06634}.

\bibitem[{Chaudhary(2023)}]{codealpaca}
Sahil Chaudhary. 2023.
\newblock \href {https://github.com/sahil280114/codealpaca} {Code alpaca: An instruction-following llama model for code generation}.
\newblock Accessed: 2024-09-23.

\bibitem[{Chen et~al.(2021)Chen, Tworek, Jun, Yuan, Pinto, Kaplan, Edwards, Burda, Joseph, Brockman et~al.}]{humanEval}
Mark Chen, Jerry Tworek, Heewoo Jun, Qiming Yuan, Henrique Ponde de~Oliveira Pinto, Jared Kaplan, Harri Edwards, Yuri Burda, Nicholas Joseph, Greg Brockman, and 1 others. 2021.
\newblock Evaluating large language models trained on code.
\newblock \emph{arXiv preprint arXiv:2107.03374}.

\bibitem[{Chen et~al.(2024)Chen, Li, Yu, Zhao, and Zhang}]{chen2024adaptivelayersplittingwireless}
Yuxuan Chen, Rongpeng Li, Xiaoxue Yu, Zhifeng Zhao, and Honggang Zhang. 2024.
\newblock \href {https://arxiv.org/abs/2406.02616} {Adaptive layer splitting for wireless llm inference in edge computing: A model-based reinforcement learning approach}.
\newblock \emph{Preprint}, arXiv:2406.02616.

\bibitem[{Chen and Pattabiraman(2024)}]{MIA-appendix}
Zitao Chen and Karthik Pattabiraman. 2024.
\newblock A method to facilitate membership inference attacks in deep learning models.
\newblock \emph{arXiv preprint arXiv:2407.01919}.

\bibitem[{Choi et~al.(2024)Choi, Thiagarajan, Glatt, and Liu}]{Parameterization}
Hongjun Choi, Jayaraman~J. Thiagarajan, Ruben Glatt, and Shusen Liu. 2024.
\newblock \href {https://arxiv.org/abs/2407.00356} {Enhancing accuracy and parameter-efficiency of neural representations for network parameterization}.
\newblock \emph{Preprint}, arXiv:2407.00356.

\bibitem[{Clark et~al.(2019)Clark, Lee, Chang, Kwiatkowski, Collins, and Toutanova}]{clark2019boolq}
Christopher Clark, Kenton Lee, Ming-Wei Chang, Tom Kwiatkowski, Michael Collins, and Kristina Toutanova. 2019.
\newblock Boolq: Exploring the surprising difficulty of natural yes/no questions.
\newblock \emph{arXiv preprint arXiv:1905.10044}.

\bibitem[{Clark et~al.(2018)Clark, Cowhey, Etzioni, Khot, Sabharwal, Schoenick, and Tafjord}]{ARC}
Peter Clark, Isaac Cowhey, Oren Etzioni, Tushar Khot, Ashish Sabharwal, Carissa Schoenick, and Oyvind Tafjord. 2018.
\newblock Think you have solved question answering? try arc, the ai2 reasoning challenge.
\newblock \emph{ArXiv}, abs/1803.05457.

\bibitem[{Cobbe et~al.(2021)Cobbe, Kosaraju, Bavarian, Hilton, Nakano, Hesse, and Schulman}]{gsm8k}
Karl Cobbe, Vineet Kosaraju, Mohammad Bavarian, Jacob Hilton, Reiichiro Nakano, Christopher Hesse, and John Schulman. 2021.
\newblock \href {https://arxiv.org/abs/2110.14168} {Training verifiers to solve math word problems}.
\newblock \emph{Preprint}, arXiv:2110.14168.

\bibitem[{Cui et~al.(2024)Cui, Yuan, Ding, Yao, He, Zhu, Ni, Xie, Xie, Lin, Liu, and Sun}]{ultrafeedback}
Ganqu Cui, Lifan Yuan, Ning Ding, Guanming Yao, Bingxiang He, Wei Zhu, Yuan Ni, Guotong Xie, Ruobing Xie, Yankai Lin, Zhiyuan Liu, and Maosong Sun. 2024.
\newblock \href {https://arxiv.org/abs/2310.01377} {Ultrafeedback: Boosting language models with scaled ai feedback}.
\newblock \emph{Preprint}, arXiv:2310.01377.

\bibitem[{Dubey et~al.(2024)Dubey, Jauhri, Pandey, Kadian, Al-Dahle, Letman, Mathur, Schelten, Yang, Fan et~al.}]{dubey2024llama3}
Abhimanyu Dubey, Abhinav Jauhri, Abhinav Pandey, Abhishek Kadian, Ahmad Al-Dahle, Aiesha Letman, Akhil Mathur, Alan Schelten, Amy Yang, Angela Fan, and 1 others. 2024.
\newblock The llama 3 herd of models.
\newblock \emph{arXiv preprint arXiv:2407.21783}.

\bibitem[{Eiras et~al.(2024)Eiras, Petrov, Vidgen, Schroeder, Pizzati, Elkins, Mukhopadhyay, Bibi, Purewal, Botos, Steibel, Keshtkar, Barez, Smith, Guadagni, Chun, Cabot, Imperial, Nolazco, Landay, Jackson, Torr, Darrell, Lee, and Foerster}]{Semiopen1}
Francisco Eiras, Aleksandar Petrov, Bertie Vidgen, Christian Schroeder, Fabio Pizzati, Katherine Elkins, Supratik Mukhopadhyay, Adel Bibi, Aaron Purewal, Csaba Botos, Fabro Steibel, Fazel Keshtkar, Fazl Barez, Genevieve Smith, Gianluca Guadagni, Jon Chun, Jordi Cabot, Joseph Imperial, Juan~Arturo Nolazco, and 6 others. 2024.
\newblock \href {https://arxiv.org/abs/2405.08597} {Risks and opportunities of open-source generative ai}.
\newblock \emph{Preprint}, arXiv:2405.08597.

\bibitem[{Elgamal and Nahrstedt(2020)}]{elgamal2020serdab}
Tarek Elgamal and Klara Nahrstedt. 2020.
\newblock Serdab: An iot framework for partitioning neural networks computation across multiple enclaves.
\newblock In \emph{2020 20th IEEE/ACM International Symposium on Cluster, Cloud and Internet Computing (CCGRID)}, pages 519--528. IEEE.

\bibitem[{Ezzeddine et~al.(2024)Ezzeddine, Ayoub, and Giordano}]{ezzeddine2024knowledge}
Fatima Ezzeddine, Omran Ayoub, and Silvia Giordano. 2024.
\newblock Knowledge distillation-based model extraction attack using private counterfactual explanations.
\newblock \emph{arXiv preprint arXiv:2404.03348}.

\bibitem[{Finlayson et~al.(2024)Finlayson, Ren, and Swayamdipta}]{finlayson2024logits}
Matthew Finlayson, Xiang Ren, and Swabha Swayamdipta. 2024.
\newblock Logits of api-protected llms leak proprietary information.
\newblock \emph{arXiv preprint arXiv:2403.09539}.

\bibitem[{Fu et~al.(2023)Fu, Wang, Gao, Liu, Li, and Jiang}]{MIA}
Wenjie Fu, Huandong Wang, Chen Gao, Guanghua Liu, Yong Li, and Tao Jiang. 2023.
\newblock Practical membership inference attacks against fine-tuned large language models via self-prompt calibration.
\newblock \emph{arXiv preprint arXiv:2311.06062}.

\bibitem[{Gao et~al.(2023)Gao, Tow, Abbasi, Biderman, Black, DiPofi, Foster, Golding, Hsu, Le~Noac'h, Li, McDonell, Muennighoff, Ociepa, Phang, Reynolds, Schoelkopf, Skowron, Sutawika, Tang, Thite, Wang, Wang, and Zou}]{eval-harness}
Leo Gao, Jonathan Tow, Baber Abbasi, Stella Biderman, Sid Black, Anthony DiPofi, Charles Foster, Laurence Golding, Jeffrey Hsu, Alain Le~Noac'h, Haonan Li, Kyle McDonell, Niklas Muennighoff, Chris Ociepa, Jason Phang, Laria Reynolds, Hailey Schoelkopf, Aviya Skowron, Lintang Sutawika, and 5 others. 2023.
\newblock \href {https://doi.org/10.5281/zenodo.10256836} {A framework for few-shot language model evaluation}.

\bibitem[{Gou et~al.(2021)Gou, Yu, Maybank, and Tao}]{gou2021knowledge}
Jianping Gou, Baosheng Yu, Stephen~J Maybank, and Dacheng Tao. 2021.
\newblock Knowledge distillation: A survey.
\newblock \emph{International Journal of Computer Vision}, 129(6):1789--1819.

\bibitem[{Guha et~al.(2024)Guha, Nyarko, Ho, R{\'e}, Chilton, Chohlas-Wood, Peters, Waldon, Rockmore, Zambrano et~al.}]{guha2024legalbench}
Neel Guha, Julian Nyarko, Daniel Ho, Christopher R{\'e}, Adam Chilton, Alex Chohlas-Wood, Austin Peters, Brandon Waldon, Daniel Rockmore, Diego Zambrano, and 1 others. 2024.
\newblock Legalbench: A collaboratively built benchmark for measuring legal reasoning in large language models.
\newblock \emph{Advances in Neural Information Processing Systems}, 36.

\bibitem[{Guha et~al.(2023)Guha, Nyarko, Ho, Ré, Chilton, Narayana, Chohlas-Wood, Peters, Waldon, Rockmore, Zambrano, Talisman, Hoque, Surani, Fagan, Sarfaty, Dickinson, Porat, Hegland, Wu, Nudell, Niklaus, Nay, Choi, Tobia, Hagan, Ma, Livermore, Rasumov-Rahe, Holzenberger, Kolt, Henderson, Rehaag, Goel, Gao, Williams, Gandhi, Zur, Iyer, and Li}]{legalbench}
Neel Guha, Julian Nyarko, Daniel~E. Ho, Christopher Ré, Adam Chilton, Aditya Narayana, Alex Chohlas-Wood, Austin Peters, Brandon Waldon, Daniel~N. Rockmore, Diego Zambrano, Dmitry Talisman, Enam Hoque, Faiz Surani, Frank Fagan, Galit Sarfaty, Gregory~M. Dickinson, Haggai Porat, Jason Hegland, and 21 others. 2023.
\newblock \href {https://arxiv.org/abs/2308.11462} {Legalbench: A collaboratively built benchmark for measuring legal reasoning in large language models}.
\newblock \emph{Preprint}, arXiv:2308.11462.

\bibitem[{He et~al.(2021)He, Lyu, Xu, and Sun}]{bert-vulnerable}
Xuanli He, Lingjuan Lyu, Qiongkai Xu, and Lichao Sun. 2021.
\newblock Model extraction and adversarial transferability, your bert is vulnerable!
\newblock \emph{arXiv preprint arXiv:2103.10013}.

\bibitem[{Hendrycks et~al.(2021)Hendrycks, Burns, Basart, Zou, Mazeika, Song, and Steinhardt}]{mmlu}
Dan Hendrycks, Collin Burns, Steven Basart, Andy Zou, Mantas Mazeika, Dawn Song, and Jacob Steinhardt. 2021.
\newblock Measuring massive multitask language understanding.
\newblock \emph{Proceedings of the International Conference on Learning Representations (ICLR)}.

\bibitem[{Hendrycks et~al.(2023)Hendrycks, Mazeika, and Woodside}]{hendrycks2023overviewcatastrophicairisks}
Dan Hendrycks, Mantas Mazeika, and Thomas Woodside. 2023.
\newblock \href {https://arxiv.org/abs/2306.12001} {An overview of catastrophic ai risks}.
\newblock \emph{Preprint}, arXiv:2306.12001.

\bibitem[{Hu et~al.(2021)Hu, Shen, Wallis, Allen-Zhu, Li, Wang, Wang, and Chen}]{LoRA}
Edward~J. Hu, Yelong Shen, Phillip Wallis, Zeyuan Allen-Zhu, Yuanzhi Li, Shean Wang, Lu~Wang, and Weizhu Chen. 2021.
\newblock \href {https://arxiv.org/abs/2106.09685} {Lora: Low-rank adaptation of large language models}.
\newblock \emph{Preprint}, arXiv:2106.09685.

\bibitem[{Hu et~al.(2020)Hu, Liang, Li, Deng, Zuo, Ji, Xie, Ding, Liu, Sherwood et~al.}]{hu2020deepsniffer}
Xing Hu, Ling Liang, Shuangchen Li, Lei Deng, Pengfei Zuo, Yu~Ji, Xinfeng Xie, Yufei Ding, Chang Liu, Timothy Sherwood, and 1 others. 2020.
\newblock Deepsniffer: A dnn model extraction framework based on learning architectural hints.
\newblock In \emph{Proceedings of the Twenty-Fifth International Conference on Architectural Support for Programming Languages and Operating Systems}, pages 385--399.

\bibitem[{Huang et~al.(2024)Huang, Wang, Cheng, Zhou, Yu, and Wang}]{10446717}
Wei Huang, Yinggui Wang, Anda Cheng, Aihui Zhou, Chaofan Yu, and Lei Wang. 2024.
\newblock \href {https://doi.org/10.1109/ICASSP48485.2024.10446717} {A fast, performant, secure distributed training framework for llm}.
\newblock In \emph{ICASSP 2024 - 2024 IEEE International Conference on Acoustics, Speech and Signal Processing (ICASSP)}, pages 4800--4804.

\bibitem[{Jacot et~al.(2018)Jacot, Gabriel, and Hongler}]{NTK-1}
Arthur Jacot, Franck Gabriel, and Cl{\'e}ment Hongler. 2018.
\newblock Neural tangent kernel: Convergence and generalization in neural networks.
\newblock \emph{Advances in neural information processing systems}, 31.

\bibitem[{Jagielski et~al.(2020)Jagielski, Carlini, Berthelot, Kurakin, and Papernot}]{jagielski2020highaccuracyhighfidelity}
Matthew Jagielski, Nicholas Carlini, David Berthelot, Alex Kurakin, and Nicolas Papernot. 2020.
\newblock \href {https://arxiv.org/abs/1909.01838} {High accuracy and high fidelity extraction of neural networks}.
\newblock \emph{Preprint}, arXiv:1909.01838.

\bibitem[{Jiang et~al.(2023)Jiang, Sablayrolles, Mensch, Bamford, Chaplot, Casas, Bressand, Lengyel, Lample, Saulnier et~al.}]{mistral}
Albert~Q Jiang, Alexandre Sablayrolles, Arthur Mensch, Chris Bamford, Devendra~Singh Chaplot, Diego de~las Casas, Florian Bressand, Gianna Lengyel, Guillaume Lample, Lucile Saulnier, and 1 others. 2023.
\newblock Mistral 7b.
\newblock \emph{arXiv preprint arXiv:2310.06825}.

\bibitem[{Joshi et~al.(2017)Joshi, Choi, Weld, and Zettlemoyer}]{TriviaQA2017}
Mandar Joshi, Eunsol Choi, Daniel~S. Weld, and Luke Zettlemoyer. 2017.
\newblock Triviaqa: A large scale distantly supervised challenge dataset for reading comprehension.
\newblock In \emph{Proceedings of the 55th Annual Meeting of the Association for Computational Linguistics}, Vancouver, Canada. Association for Computational Linguistics.

\bibitem[{Kwiatkowski et~al.(2019)Kwiatkowski, Palomaki, Redfield, Collins, Parikh, Alberti, Epstein, Polosukhin, Devlin, Lee et~al.}]{nq_open}
Tom Kwiatkowski, Jennimaria Palomaki, Olivia Redfield, Michael Collins, Ankur Parikh, Chris Alberti, Danielle Epstein, Illia Polosukhin, Jacob Devlin, Kenton Lee, and 1 others. 2019.
\newblock Natural questions: a benchmark for question answering research.
\newblock \emph{Transactions of the Association for Computational Linguistics}, 7:453--466.

\bibitem[{Lee et~al.(2018)Lee, Edwards, Molloy, and Su}]{SAP-DP-supp}
Taesung Lee, Benjamin Edwards, Ian Molloy, and Dong Su. 2018.
\newblock Defending against machine learning model stealing attacks using deceptive perturbations.
\newblock \emph{arXiv preprint arXiv:1806.00054}.

\bibitem[{Lemmens and Nussbaum(2012)}]{lemmens2012nonlinear}
Bas Lemmens and Roger Nussbaum. 2012.
\newblock \emph{Nonlinear Perron-Frobenius Theory}, volume 189.
\newblock Cambridge University Press.

\bibitem[{Li et~al.(2024{\natexlab{a}})Li, Zhang, Yao, Cai, Guo, and Chen}]{li2024teeslice}
Ding Li, Ziqi Zhang, Mengyu Yao, Yifeng Cai, Yao Guo, and Xiangqun Chen. 2024{\natexlab{a}}.
\newblock Teeslice: Protecting sensitive neural network models in trusted execution environments when attackers have pre-trained models.
\newblock \emph{arXiv preprint arXiv:2411.09945}.

\bibitem[{Li et~al.(2024{\natexlab{b}})Li, Hong, Xie, Tan, Xin, Hou, Yin, Wang, Hendrycks, Wang, Li, He, and Song}]{AIA-appendix}
Qinbin Li, Junyuan Hong, Chulin Xie, Jeffrey Tan, Rachel Xin, Junyi Hou, Xavier Yin, Zhun Wang, Dan Hendrycks, Zhangyang Wang, Bo~Li, Bingsheng He, and Dawn Song. 2024{\natexlab{b}}.
\newblock \href {https://arxiv.org/abs/2408.12787} {Llm-pbe: Assessing data privacy in large language models}.
\newblock \emph{Preprint}, arXiv:2408.12787.

\bibitem[{Li et~al.(2024{\natexlab{c}})Li, Hong, Xie, Tan, Xin, Hou, Yin, Wang, Hendrycks, Wang et~al.}]{li2024llm}
Qinbin Li, Junyuan Hong, Chulin Xie, Jeffrey Tan, Rachel Xin, Junyi Hou, Xavier Yin, Zhun Wang, Dan Hendrycks, Zhangyang Wang, and 1 others. 2024{\natexlab{c}}.
\newblock Llm-pbe: Assessing data privacy in large language models.
\newblock \emph{arXiv preprint arXiv:2408.12787}.

\bibitem[{Li et~al.(2023)Li, Bubeck, Eldan, Del~Giorno, Gunasekar, and Lee}]{phi1-5}
Yuanzhi Li, S{\'e}bastien Bubeck, Ronen Eldan, Allie Del~Giorno, Suriya Gunasekar, and Yin~Tat Lee. 2023.
\newblock Textbooks are all you need ii: phi-1.5 technical report.
\newblock \emph{arXiv preprint arXiv:2309.05463}.

\bibitem[{Lin et~al.(2024)Lin, Qu, Chen, and Huang}]{lin2024splitlearning6gedge}
Zheng Lin, Guanqiao Qu, Xianhao Chen, and Kaibin Huang. 2024.
\newblock \href {https://arxiv.org/abs/2306.12194} {Split learning in 6g edge networks}.
\newblock \emph{Preprint}, arXiv:2306.12194.

\bibitem[{Liu et~al.(2024)Liu, Yang, Chen, and Lin}]{APISecurity}
Jinghua Liu, Yi~Yang, Kai Chen, and Miaoqian Lin. 2024.
\newblock Generating api parameter security rules with llm for api misuse detection.
\newblock \emph{arXiv preprint arXiv:2409.09288}.

\bibitem[{Liu et~al.(2023)Liu, Xu, Chen, and Xiao}]{liu2023autodan}
Xiaogeng Liu, Nan Xu, Muhao Chen, and Chaowei Xiao. 2023.
\newblock Autodan: Generating stealthy jailbreak prompts on aligned large language models.
\newblock \emph{arXiv preprint arXiv:2310.04451}.

\bibitem[{Meng et~al.(2024)Meng, Xia, and Chen}]{simpo}
Yu~Meng, Mengzhou Xia, and Danqi Chen. 2024.
\newblock Simpo: Simple preference optimization with a reference-free reward.
\newblock \emph{arXiv preprint arXiv:2405.14734}.

\bibitem[{Mihaylov et~al.(2018)Mihaylov, Clark, Khot, and Sabharwal}]{OpenBookQA2018}
Todor Mihaylov, Peter Clark, Tushar Khot, and Ashish Sabharwal. 2018.
\newblock Can a suit of armor conduct electricity? a new dataset for open book question answering.
\newblock In \emph{EMNLP}.

\bibitem[{Minaee et~al.(2024)Minaee, Mikolov, Nikzad, Chenaghlu, Socher, Amatriain, and Gao}]{minaee2024largelanguagemodelssurvey}
Shervin Minaee, Tomas Mikolov, Narjes Nikzad, Meysam Chenaghlu, Richard Socher, Xavier Amatriain, and Jianfeng Gao. 2024.
\newblock \href {https://arxiv.org/abs/2402.06196} {Large language models: A survey}.
\newblock \emph{Preprint}, arXiv:2402.06196.

\bibitem[{Mo et~al.(2020)Mo, Shamsabadi, Katevas, Demetriou, Leontiadis, Cavallaro, and Haddadi}]{mo2020darknetz}
Fan Mo, Ali~Shahin Shamsabadi, Kleomenis Katevas, Soteris Demetriou, Ilias Leontiadis, Andrea Cavallaro, and Hamed Haddadi. 2020.
\newblock Darknetz: towards model privacy at the edge using trusted execution environments.
\newblock In \emph{Proceedings of the 18th International Conference on Mobile Systems, Applications, and Services}, pages 161--174.

\bibitem[{Mukherjee et~al.(2023)Mukherjee, Mitra, Jawahar, Agarwal, Palangi, and Awadallah}]{orca}
Subhabrata Mukherjee, Arindam Mitra, Ganesh Jawahar, Sahaj Agarwal, Hamid Palangi, and Ahmed Awadallah. 2023.
\newblock Orca: Progressive learning from complex explanation traces of gpt-4.
\newblock \emph{arXiv preprint arXiv:2306.02707}.

\bibitem[{Narra et~al.(2019)Narra, Lin, Wang, Balasubramaniam, and Annavaram}]{narra2019privacy}
Krishna~Giri Narra, Zhifeng Lin, Yongqin Wang, Keshav Balasubramaniam, and Murali Annavaram. 2019.
\newblock Privacy-preserving inference in machine learning services using trusted execution environments.
\newblock \emph{arXiv preprint arXiv:1912.03485}.

\bibitem[{Nayan et~al.(2024)Nayan, Guo, Al~Duniawi, Botacin, Uluagac, and Sun}]{nayan2024sok}
Tushar Nayan, Qiming Guo, Mohammed Al~Duniawi, Marcus Botacin, Selcuk Uluagac, and Ruimin Sun. 2024.
\newblock $\{$SoK$\}$: All you need to know about $\{$On-Device$\}$$\{$ML$\}$ model extraction-the gap between research and practice.
\newblock In \emph{33rd USENIX Security Symposium (USENIX Security 24)}, pages 5233--5250.

\bibitem[{Nevo et~al.(2024)Nevo, Lahav, Karpur, Bar-On, Bradley, and Alstott}]{nevo2024securing}
Sella Nevo, Dan Lahav, Ajay Karpur, Yogev Bar-On, Henry-Alexander Bradley, and Jeff Alstott. 2024.
\newblock \emph{Securing AI model weights: Preventing theft and misuse of frontier models}.
\newblock 1. Rand Corporation.

\bibitem[{Orekondy et~al.(2018)Orekondy, Schiele, and Fritz}]{orekondy2018knockoffnetsstealingfunctionality}
Tribhuvanesh Orekondy, Bernt Schiele, and Mario Fritz. 2018.
\newblock \href {https://arxiv.org/abs/1812.02766} {Knockoff nets: Stealing functionality of black-box models}.
\newblock \emph{Preprint}, arXiv:1812.02766.

\bibitem[{Paperno et~al.(2016)Paperno, Kruszewski, Lazaridou, Pham, Bernardi, Pezzelle, Baroni, Boleda, and Fernández}]{LAMBADA}
Denis Paperno, Germán Kruszewski, Angeliki Lazaridou, Quan~Ngoc Pham, Raffaella Bernardi, Sandro Pezzelle, Marco Baroni, Gemma Boleda, and Raquel Fernández. 2016.
\newblock \href {https://doi.org/10.5281/zenodo.2630551} {The lambada dataset}.

\bibitem[{Pinto and Santos(2019)}]{pinto2019demystifying}
Sandro Pinto and Nuno Santos. 2019.
\newblock Demystifying arm trustzone: A comprehensive survey.
\newblock \emph{ACM computing surveys (CSUR)}, 51(6):1--36.

\bibitem[{Rajpurkar et~al.(2018)Rajpurkar, Jia, and Liang}]{SQuAD}
Pranav Rajpurkar, Robin Jia, and Percy Liang. 2018.
\newblock \href {https://arxiv.org/abs/1806.03822} {Know what you don't know: Unanswerable questions for squad}.
\newblock \emph{Preprint}, arXiv:1806.03822.

\bibitem[{Rakin et~al.(2022)Rakin, Chowdhuryy, Yao, and Fan}]{rakin2022deepsteal}
Adnan~Siraj Rakin, Md~Hafizul~Islam Chowdhuryy, Fan Yao, and Deliang Fan. 2022.
\newblock Deepsteal: Advanced model extractions leveraging efficient weight stealing in memories.
\newblock In \emph{2022 IEEE symposium on security and privacy (SP)}, pages 1157--1174. IEEE.

\bibitem[{Sakaguchi et~al.(2019)Sakaguchi, Bras, Bhagavatula, and Choi}]{sakaguchi2019winogrande}
Keisuke Sakaguchi, Ronan~Le Bras, Chandra Bhagavatula, and Yejin Choi. 2019.
\newblock Winogrande: An adversarial winograd schema challenge at scale.
\newblock \emph{arXiv preprint arXiv:1907.10641}.

\bibitem[{Sanh et~al.(2020)Sanh, Debut, Chaumond, and Wolf}]{DistilBERT}
Victor Sanh, Lysandre Debut, Julien Chaumond, and Thomas Wolf. 2020.
\newblock \href {https://arxiv.org/abs/1910.01108} {Distilbert, a distilled version of bert: smaller, faster, cheaper and lighter}.
\newblock \emph{Preprint}, arXiv:1910.01108.

\bibitem[{Schillaci(2024)}]{schillaci2024site}
Zachary Schillaci. 2024.
\newblock On-site deployment of llms.
\newblock In \emph{Large Language Models in Cybersecurity: Threats, Exposure and Mitigation}, pages 205--211. Springer Nature Switzerland Cham.

\bibitem[{Shen et~al.(2022)Shen, Qi, Jiang, Wang, Wen, Chen, Zhao, Wang, Chen, Luo et~al.}]{shen2022soter}
Tianxiang Shen, Ji~Qi, Jianyu Jiang, Xian Wang, Siyuan Wen, Xusheng Chen, Shixiong Zhao, Sen Wang, Li~Chen, Xiapu Luo, and 1 others. 2022.
\newblock $\{$SOTER$\}$: Guarding black-box inference for general neural networks at the edge.
\newblock In \emph{2022 USENIX Annual Technical Conference (USENIX ATC 22)}, pages 723--738.

\bibitem[{Shen et~al.(2023)Shen, Liu, Liu, Hong, Duan, Huang, Mao, Wu, and Wu}]{SAP}
Xicong Shen, Yang Liu, Huiqi Liu, Jue Hong, Bing Duan, Zirui Huang, Yunlong Mao, Ye~Wu, and Di~Wu. 2023.
\newblock \href {https://arxiv.org/abs/2312.15603} {A split-and-privatize framework for large language model fine-tuning}.
\newblock \emph{Preprint}, arXiv:2312.15603.

\bibitem[{Sitawarin et~al.(2024)Sitawarin, Mu, Wagner, and Araujo}]{sitawarin2024pal}
Chawin Sitawarin, Norman Mu, David Wagner, and Alexandre Araujo. 2024.
\newblock Pal: Proxy-guided black-box attack on large language models.
\newblock \emph{arXiv preprint arXiv:2402.09674}.

\bibitem[{Staab et~al.(2023)Staab, Vero, Balunovi{\'c}, and Vechev}]{AIA}
Robin Staab, Mark Vero, Mislav Balunovi{\'c}, and Martin Vechev. 2023.
\newblock Beyond memorization: Violating privacy via inference with large language models.
\newblock \emph{arXiv preprint arXiv:2310.07298}.

\bibitem[{Suzgun et~al.(2022)Suzgun, Scales, Sch{\"a}rli, Gehrmann, Tay, Chung, Chowdhery, Le, Chi, Zhou, , and Wei}]{bbh}
Mirac Suzgun, Nathan Scales, Nathanael Sch{\"a}rli, Sebastian Gehrmann, Yi~Tay, Hyung~Won Chung, Aakanksha Chowdhery, Quoc~V Le, Ed~H Chi, Denny Zhou, , and Jason Wei. 2022.
\newblock Challenging big-bench tasks and whether chain-of-thought can solve them.
\newblock \emph{arXiv preprint arXiv:2210.09261}.

\bibitem[{Tamber et~al.(2024)Tamber, Xian, and Lin}]{canHideApi}
Manveer~Singh Tamber, Jasper Xian, and Jimmy Lin. 2024.
\newblock Can't hide behind the api: Stealing black-box commercial embedding models.
\newblock \emph{arXiv preprint arXiv:2406.09355}.

\bibitem[{Taori et~al.(2023)Taori, Gulrajani, Zhang, Dubois, Li, Guestrin, Liang, and Hashimoto}]{stanfordAlpaca}
Rohan Taori, Ishaan Gulrajani, Tianyi Zhang, Yann Dubois, Xuechen Li, Carlos Guestrin, Percy Liang, and Tatsunori~B. Hashimoto. 2023.
\newblock Stanford alpaca: An instruction-following llama model.
\newblock \url{https://github.com/tatsu-lab/stanford_alpaca}.

\bibitem[{Touvron et~al.(2023)Touvron, Martin, Stone, Albert, Almahairi, Babaei, Bashlykov, Batra, Bhargava, Bhosale et~al.}]{llama2}
Hugo Touvron, Louis Martin, Kevin Stone, Peter Albert, Amjad Almahairi, Yasmine Babaei, Nikolay Bashlykov, Soumya Batra, Prajjwal Bhargava, Shruti Bhosale, and 1 others. 2023.
\newblock Llama 2: Open foundation and fine-tuned chat models.
\newblock \emph{arXiv preprint arXiv:2307.09288}.

\bibitem[{Tram{\`e}r et~al.(2016)Tram{\`e}r, Zhang, Juels, Reiter, and Ristenpart}]{tramer2016stealing}
Florian Tram{\`e}r, Fan Zhang, Ari Juels, Michael~K Reiter, and Thomas Ristenpart. 2016.
\newblock Stealing machine learning models via prediction $\{$APIs$\}$.
\newblock In \emph{25th USENIX security symposium (USENIX Security 16)}, pages 601--618.

\bibitem[{Truong et~al.(2021)Truong, Maini, Walls, and Papernot}]{truong2021data}
Jean-Baptiste Truong, Pratyush Maini, Robert~J Walls, and Nicolas Papernot. 2021.
\newblock Data-free model extraction.
\newblock In \emph{Proceedings of the IEEE/CVF conference on computer vision and pattern recognition}, pages 4771--4780.

\bibitem[{Wang et~al.(2023{\natexlab{a}})Wang, Yang, and Wang}]{FinanceInstruct}
Neng Wang, Hongyang Yang, and Christina~Dan Wang. 2023{\natexlab{a}}.
\newblock Fingpt: Instruction tuning benchmark for open-source large language models in financial datasets.
\newblock \emph{arXiv preprint arXiv:2310.04793}.

\bibitem[{Wang et~al.(2023{\natexlab{b}})Wang, Yang, and Wang}]{FinGpt}
Neng Wang, Hongyang Yang, and Christina~Dan Wang. 2023{\natexlab{b}}.
\newblock \href {https://arxiv.org/abs/2310.04793} {Fingpt: Instruction tuning benchmark for open-source large language models in financial datasets}.
\newblock \emph{Preprint}, arXiv:2310.04793.

\bibitem[{Wang et~al.(2022)Wang, Kordi, Mishra, Liu, Smith, Khashabi, and Hajishirzi}]{alpacaDataset}
Yizhong Wang, Yeganeh Kordi, Swaroop Mishra, Alisa Liu, Noah~A Smith, Daniel Khashabi, and Hannaneh Hajishirzi. 2022.
\newblock Self-instruct: Aligning language models with self-generated instructions.
\newblock \emph{arXiv preprint arXiv:2212.10560}.

\bibitem[{Wei et~al.(2019)Wei, Lee, Liu, and Ma}]{NTK-2}
Colin Wei, Jason~D Lee, Qiang Liu, and Tengyu Ma. 2019.
\newblock Regularization matters: Generalization and optimization of neural nets vs their induced kernel.
\newblock \emph{Advances in Neural Information Processing Systems}, 32.

\bibitem[{Wolfe et~al.(2024)Wolfe, Slaughter, Han, Wen, Yang, Rosenblatt, Herman, Brown, Qu, Weber et~al.}]{wolfe2024laboratory}
Robert Wolfe, Isaac Slaughter, Bin Han, Bingbing Wen, Yiwei Yang, Lucas Rosenblatt, Bernease Herman, Eva Brown, Zening Qu, Nic Weber, and 1 others. 2024.
\newblock Laboratory-scale ai: Open-weight models are competitive with chatgpt even in low-resource settings.
\newblock In \emph{The 2024 ACM Conference on Fairness, Accountability, and Transparency}, pages 1199--1210.

\bibitem[{Xu et~al.(2023)Xu, Guo, Duan, and McAuley}]{baize}
Canwen Xu, Daya Guo, Nan Duan, and Julian McAuley. 2023.
\newblock Baize: An open-source chat model with parameter-efficient tuning on self-chat data.
\newblock \emph{arXiv preprint arXiv:2304.01196}.

\bibitem[{Xu et~al.(2024{\natexlab{a}})Xu, Li, Tao, Shen, Cheng, Li, Xu, Tao, and Zhou}]{xu2024survey}
Xiaohan Xu, Ming Li, Chongyang Tao, Tao Shen, Reynold Cheng, Jinyang Li, Can Xu, Dacheng Tao, and Tianyi Zhou. 2024{\natexlab{a}}.
\newblock A survey on knowledge distillation of large language models.
\newblock \emph{arXiv preprint arXiv:2402.13116}.

\bibitem[{Xu et~al.(2024{\natexlab{b}})Xu, Liu, Deng, Li, and Picek}]{PIA-appendix2}
Zihao Xu, Yi~Liu, Gelei Deng, Yuekang Li, and Stjepan Picek. 2024{\natexlab{b}}.
\newblock \href {https://arxiv.org/abs/2402.13457} {A comprehensive study of jailbreak attack versus defense for large language models}.
\newblock \emph{Preprint}, arXiv:2402.13457.

\bibitem[{Yue et~al.(2023)Yue, Qu, Zhang, Fu, Huang, Sun, Su, and Chen}]{mathInstruct}
Xiang Yue, Xingwei Qu, Ge~Zhang, Yao Fu, Wenhao Huang, Huan Sun, Yu~Su, and Wenhu Chen. 2023.
\newblock Mammoth: Building math generalist models through hybrid instruction tuning.
\newblock \emph{arXiv preprint arXiv:2309.05653}.

\bibitem[{Zanella-Beguelin et~al.(2021)Zanella-Beguelin, Tople, Paverd, and K{\"o}pf}]{greybox}
Santiago Zanella-Beguelin, Shruti Tople, Andrew Paverd, and Boris K{\"o}pf. 2021.
\newblock Grey-box extraction of natural language models.
\newblock In \emph{International Conference on Machine Learning}, pages 12278--12286. PMLR.

\bibitem[{Zellers et~al.(2019)Zellers, Holtzman, Bisk, Farhadi, and Choi}]{hellaswag}
Rowan Zellers, Ari Holtzman, Yonatan Bisk, Ali Farhadi, and Yejin Choi. 2019.
\newblock Hellaswag: Can a machine really finish your sentence?
\newblock In \emph{Proceedings of the 57th Annual Meeting of the Association for Computational Linguistics}.

\bibitem[{Zhang et~al.(2022)Zhang, Roller, Goyal, Artetxe, Chen, Chen, Dewan, Diab, Li, Lin et~al.}]{opt-350}
Susan Zhang, Stephen Roller, Naman Goyal, Mikel Artetxe, Moya Chen, Shuohui Chen, Christopher Dewan, Mona Diab, Xian Li, Xi~Victoria Lin, and 1 others. 2022.
\newblock Opt: Open pre-trained transformer language models.
\newblock \emph{arXiv preprint arXiv:2205.01068}.

\bibitem[{Zhang et~al.(2016)Zhang, Zhao, and LeCun}]{AGnews}
Xiang Zhang, Junbo Zhao, and Yann LeCun. 2016.
\newblock \href {https://arxiv.org/abs/1509.01626} {Character-level convolutional networks for text classification}.
\newblock \emph{Preprint}, arXiv:1509.01626.

\bibitem[{Zhang et~al.(2023)Zhang, Tian, Yang, Chen, Li, and Petzold}]{medicalInstruct}
Xinlu Zhang, Chenxin Tian, Xianjun Yang, Lichang Chen, Zekun Li, and Linda~Ruth Petzold. 2023.
\newblock Alpacare: Instruction-tuned large language models for medical application.
\newblock \emph{arXiv preprint arXiv:2310.14558}.

\bibitem[{Zhang et~al.(2024{\natexlab{a}})Zhang, Wang, Zhang, Zhang, Zhang, Liu, and Wu}]{zhanggroupcover}
Zheng Zhang, Na~Wang, Ziqi Zhang, Yao Zhang, Tianyi Zhang, Jianwei Liu, and Ye~Wu. 2024{\natexlab{a}}.
\newblock Groupcover: A secure, efficient and scalable inference framework for on-device model protection based on tees.
\newblock In \emph{Proceedings of the 41st International Conference on Machine Learning}.

\bibitem[{Zhang et~al.(2024{\natexlab{b}})Zhang, Gong, Cai, Yuan, Liu, Li, Guo, and Chen}]{zhang2024no}
Ziqi Zhang, Chen Gong, Yifeng Cai, Yuanyuan Yuan, Bingyan Liu, Ding Li, Yao Guo, and Xiangqun Chen. 2024{\natexlab{b}}.
\newblock No privacy left outside: On the (in-) security of tee-shielded dnn partition for on-device ml.
\newblock In \emph{2024 IEEE Symposium on Security and Privacy (SP)}, pages 3327--3345. IEEE.

\bibitem[{Zhao et~al.(2023)Zhao, Zhou, Li, Tang, Wang, Hou, Min, Zhang, Zhang, Dong, Du, Yang, Chen, Chen, Jiang, Ren, Li, Tang, Liu, Liu, Nie, and Wen}]{zhao2023surveylargelanguagemodels}
Wayne~Xin Zhao, Kun Zhou, Junyi Li, Tianyi Tang, Xiaolei Wang, Yupeng Hou, Yingqian Min, Beichen Zhang, Junjie Zhang, Zican Dong, Yifan Du, Chen Yang, Yushuo Chen, Zhipeng Chen, Jinhao Jiang, Ruiyang Ren, Yifan Li, Xinyu Tang, Zikang Liu, and 3 others. 2023.
\newblock \href {https://arxiv.org/abs/2303.18223} {A survey of large language models}.
\newblock \emph{Preprint}, arXiv:2303.18223.

\bibitem[{Zhao et~al.(2024)Zhao, Li, Li, Zhang, and Sun}]{PIA-appendix}
Wei Zhao, Zhe Li, Yige Li, Ye~Zhang, and Jun Sun. 2024.
\newblock Defending large language models against jailbreak attacks via layer-specific editing.
\newblock \emph{arXiv preprint arXiv:2405.18166}.

\bibitem[{Zheng et~al.(2023)Zheng, Chiang, Sheng, Zhuang, Wu, Zhuang, Lin, Li, Li, Xing, Zhang, Gonzalez, and Stoica}]{MT-Bench}
Lianmin Zheng, Wei-Lin Chiang, Ying Sheng, Siyuan Zhuang, Zhanghao Wu, Yonghao Zhuang, Zi~Lin, Zhuohan Li, Dacheng Li, Eric~P. Xing, Hao Zhang, Joseph~E. Gonzalez, and Ion Stoica. 2023.
\newblock \href {https://arxiv.org/abs/2306.05685} {Judging llm-as-a-judge with mt-bench and chatbot arena}.
\newblock \emph{Preprint}, arXiv:2306.05685.

\bibitem[{Zheng et~al.(2024{\natexlab{a}})Zheng, Zhang, Shen, Liu, Lin, Fu, Chen, and Yue}]{CodeInstruct}
Tianyu Zheng, Ge~Zhang, Tianhao Shen, Xueling Liu, Bill~Yuchen Lin, Jie Fu, Wenhu Chen, and Xiang Yue. 2024{\natexlab{a}}.
\newblock Opencodeinterpreter: Integrating code generation with execution and refinement.
\newblock \emph{arXiv preprint arXiv:2402.14658}.

\bibitem[{Zheng et~al.(2024{\natexlab{b}})Zheng, Zhang, Shen, Liu, Lin, Fu, Chen, and Yue}]{Code-Feedback}
Tianyu Zheng, Ge~Zhang, Tianhao Shen, Xueling Liu, Bill~Yuchen Lin, Jie Fu, Wenhu Chen, and Xiang Yue. 2024{\natexlab{b}}.
\newblock Opencodeinterpreter: Integrating code generation with execution and refinement.
\newblock \emph{arXiv preprint arXiv:2402.14658}.

\bibitem[{Zou et~al.(2020)Zou, Cao, Zhou, and Gu}]{ReLU}
Difan Zou, Yuan Cao, Dongruo Zhou, and Quanquan Gu. 2020.
\newblock Gradient descent optimizes over-parameterized deep relu networks.
\newblock \emph{Machine learning}, 109:467--492.

\end{thebibliography}

\appendix
\newpage
\section{Proof of Theorem~\ref{theorem:transition layer}}\label{appendix::thm-proof}
In this section, we prove Theorem~\ref{theorem:transition layer}. We first revisit the our model, present several important lemmas and finally present the proof. Additional explanatory remarks are included in Appendix~\ref{append:explor_remarks}.

\subsection{Model Overview}
The distilled model $f(\mathbf{X};\bm{\theta})$ is structured as a sequence of $L$ transformer layers, 
{\small
\begin{equation}
    f(\mathbf{X})=
    \varphi_L \circ \varphi_{L-1}\circ ...\circ \varphi_{\alpha L+1}\circ \hat{\varphi}_{\alpha L}\circ_{\alpha L -1} \circ ...\circ \varphi_{1}(\mathbf{X}),
\end{equation}}
where $\mathbf{X} \in \mathbb{R}^{n \times d}$ represents the input, interpreted as an assembly of $n$ tokens, each possessing $d$ hidden dimensions.
Each transformer layer, indexed by $1\le i\le L$, is represented by $\varphi_{i}$, which maps $ \mathbb{R}^{n\times d}$ to $\mathbb{R}^{n\times d}$ and can be defined as follows,
{\small
\begin{equation}
    \varphi_i\left(\mathbf{X} ; K_i, Q_i\right)=
\left [ \mathbf{I}_n +\operatorname{softmax}\left(\frac{\mathbf{X}Q_i (\mathbf{X}K_i)^\top}{\sqrt{d_Q}\|\mathbf{X}\|^2}\right)\right ]\mathbf{X}, 
\end{equation}}
where $Q_i \in \mathbb{R}^{d \times d_Q}$, $K_i \in \mathbb{R}^{d\times d_Q}$ represent projection parameter matrices.
Here, the $\alpha L$-th layer is the distilled layer and the others are the public layers. 
For simplicity, we use the function $\hat{\varphi}_{\alpha L}$ to denote mapping of the distilled layer, i.e., $\hat{\varphi}_{\alpha L}(\mathbf{X})=\varphi_{\alpha L}(\mathbf{X};\hat{K}_{\alpha L}, \hat{Q}_{\alpha L})$. 

\subsection{Bounds on Different Orthogonal Components }

\begin{Lem}\label{v}
For any $1\leq l \leq L$, $1\leq p \leq d$, any $  \mathbf{X} \in \mathbb{R}^{n \times d}$, we have
{\small\begin{equation}
\begin{split}
   \max_{\boldsymbol {v}:\left\|\boldsymbol {v}\right\|_2=1,  \boldsymbol {v}\perp \mathbb{I}_n  }& 
\left | \boldsymbol {v}^{\top}\varphi_l\left(\mathbf{X} ; K_l, Q_l\right)[p]  \right |\\ 
&\leq (1+\beta_D)\max_{\boldsymbol {v}:\left\|\boldsymbol {v}\right\|_2=1, 
\boldsymbol {v}\perp \mathbb{I}_n  } 
\left | \boldsymbol {v}^{\top}\mathbf{X}[p]  \right |
\end{split}
\end{equation}}
where $\mathbb{I}_n$  is a column vector with dimensions  $n \times 1$ and each element is 1,
$\mathbf{X}[p]$ is the $p$-th column of the input $\mathbf{X}$, $\varphi_l\left(\mathbf{X} ; K_l, Q_l\right)[p]$ is the $p$-th column of the $l$-th self-attention output,
the coefficient $\beta_D$ satisfies $0<\beta_D<1$ and it is related to the upper bound of the L2-norm of matrices $K_l, Q_l$.
\end{Lem}

\begin{proof}

Let $\boldsymbol {u}=
\left \{ \boldsymbol {u}_{l,1}=\frac{\mathbb{I}_n}{\sqrt{n} } ,
\boldsymbol {u}_{l,2}, \ldots, \boldsymbol {u}_{l,n} \right \} $ denote the eigenvectors of $\operatorname{softmax}\left(\frac{\mathbf{X}Q_l (\mathbf{X}K_l)^\top}{\sqrt{d_Q}\|\mathbf{X}\|^2 }\right)$. 
Assume $\sigma_{l,1},\sigma_{l,2}, \ldots, \sigma_{l,n}$ denote the eigenvalues of $\operatorname{softmax}\left(\frac{\mathbf{X}Q_i (\mathbf{X}K_i)^\top}{\sqrt{d_Q}\|\mathbf{X}\|^2 }\right)$ and $-1<\sigma_{l,n}<\beta_D$ for any $l,n$.
Thus we have
{\small\begin{subequations}
    \begin{align}
    &\boldsymbol {v}^{\top}\varphi_l\left(\mathbf{X} ; K_l, Q_l\right)[p] \\
    &= \boldsymbol {v}^{\top} \left [ \mathbf{I}_n +\operatorname{softmax}\left(\frac{\mathbf{X}Q_l (\mathbf{X}K_l)^\top}{\sqrt{d_Q}\|\mathbf{X}\|^2}\right)\right ]\mathbf{X}[p] \\
    &= \boldsymbol {v}^{\top} \left [ \mathbf{I}_n +\operatorname{softmax}\left(\frac{\mathbf{X}Q_l (\mathbf{X}K_l)^\top}{\sqrt{d_Q}\|\mathbf{X}\|^2}\right)\right ]
   \sum_{k=1}^n \alpha_{p k}\boldsymbol {u}_{l,k}\\
   &= \boldsymbol {v}^{\top} \sum_{k=1}^n \alpha_{p k}(1+\sigma_{l,k})\boldsymbol {u}_{l,k}\label{12c}
   \\
   &\leq \max_{\boldsymbol {v}:\left\|\boldsymbol {v}\right\|_2=1, \boldsymbol {v}\perp \mathbb{I}_n  } 
\left |  \sum_{k=2}^n \alpha_{p k} (1+\sigma_{l,k})\boldsymbol {v}^{\top} {u}_{l,k} \right | \\
&= \left\|   \sum_{k=2}^n \alpha_{p k} (1+\sigma_{l,k})\boldsymbol  {u}_{l,k}    \right\|_2 \label{12e}\\
&=  \left [ \sum_{k=2}^n \alpha^{2}_{p k} (1+\sigma_{l,k})^2 \right ] ^{1/2}
\\
&\leq (1+\beta_D)\max_{\boldsymbol {v}:\left\|\boldsymbol {v}\right\|_2=1, 
\boldsymbol {v}\perp \mathbb{I}_n  } 
\left | \boldsymbol {v}^{\top}\mathbf{X}[p]  \right |,
    \end{align}
\end{subequations}}
where 
{\small\begin{align*}
\beta_D &= \max_{\substack{\left\|K_l\right\|_2\leq D,\\\left\|Q_l\right\|_2\leq D}}
\max_{\substack{\boldsymbol{v}:\left\|\boldsymbol{v}\right\|_2=1,\\ \boldsymbol{v}\perp \mathbb{I}_n}} 
\left\|
\operatorname{softmax}\left(
\frac{\mathbf{X}Q_l (\mathbf{X}K_l)^\top}{\sqrt{d_Q}\|\mathbf{X}\|^2}
\right)
\boldsymbol{v}
\right\|_2 \\&< 1
\end{align*}
}

The equation \eqref{12c} is due to $\boldsymbol {u}_{l,k}$ are the eigenvectors of $\operatorname{softmax}\left(\frac{\mathbf{X}Q_l (\mathbf{X}K_l)^\top}{\sqrt{d_Q}\|\mathbf{X}\|^2 }\right)$. The inequality \eqref{12e} is because when $\boldsymbol {v}=\frac{\sum_{k=2}^n \alpha_{p k} (1+\sigma_{l,k})\boldsymbol {u}_{l,k}}{\left \| \sum_{k=2}^n \alpha_{p k} (1+\sigma_{l,k}) {u}_{l,k} \right \|_2 } 
$, we have the maximum value.

\end{proof}

\begin{Lem}\label{self-att}
For any $K_l, Q_l \in \mathbb{R}^{d \times s}$ and any $\mathbf{X} \in \mathbb{R}^{n \times d}$, the following equation always holds:
\begin{equation}
\left|\mathbb{I}_n^\top \varphi_i\left(\mathbf{X} ; K_i, Q_i\right)[p]\right| = 2\left|\mathbb{I}_n^\top \mathbf{X}[p]\right|,
\end{equation}
where $\mathbf{X}[p]$ is the $p$-th column of the input $\mathbf{X}$, $\varphi_i\left(\mathbf{X} ; K_i, Q_i\right)[p]$ is the $p$-th column of the $l$-th self-attention output.
\end{Lem}

\begin{proof}
Assume that a set of orthogonal basis for $\mathbb{R}^{n}$ is $\left \{  {\boldsymbol {u_1},\boldsymbol {u_2},\dots,\boldsymbol {u_n}}\right \} $, where $\boldsymbol {u_1}=\frac{\mathbb{I}_n}{\sqrt{n} } $.
Then we can rewrite $\mathbf{X}[p]$ as $\mathbf{X}[p]=\sum_{j=1}^n \alpha_{p j}\boldsymbol {u_j} $,
where $\alpha_{p j}$$(1\le p \le d)$ are the corresponding coefficients for the $p$-th column of $\mathbf{X}$ under the orthogonal basis.
Next, we calculate $\left|\mathbb{I}_n^\top f(\mathbf{X})[p]\right|$ and $\left|\mathbb{I}_n^\top \mathbf{X}[p]\right|$, respectively.
Note that $\mathbb{I}_n^\top \boldsymbol {u_j}=0 $ for all $j\ne 1$. Therefore, we can obtain that,
\begin{equation}
\mathbb{I}_n^\top\mathbf{X}[p]=\sqrt{n}\alpha_{p1}.
\end{equation}
Then we can get
\begin{equation}\label{right}
   \left|\mathbb{I}_n^\top \mathbf{X}[p]\right|=|\sqrt{n}\alpha_{p1}|.
\end{equation}

Let $\sigma_{i1},\sigma_{i2}, \ldots, \sigma_{in}$ denote the eigenvalues of $\operatorname{softmax}\left(\frac{\mathbf{X}Q_i (\mathbf{X}K_i)^\top}{\sqrt{d_Q}\|\mathbf{X}\|^2 }\right)$.
Applying the Perron–Frobenius theorem for Markov matrices~\cite{lemmens2012nonlinear}, we deduce that for the matrix $\operatorname{softmax}\left(\frac{\mathbf{X}Q_l (\mathbf{X}K_i)^\top}{\sqrt{d_Q} \|\mathbf{X}\|^2}\right)$, there exists only one eigenvalue equal to 1, while all other eigenvalues in absolute value are strictly less than 1. Without loss of generality, we assume $\sigma_{i1}=1$, implying $\left|\sigma_{ij}\right| < 1$ for $j \neq 1$.
Recalling the definition of $\varphi_i\left(\mathbf{X}; K_i, Q_i\right)$ and considering the linear operation, we can rewrite it as follows:
\begin{equation}\label{si2}
   \varphi_i\left(\mathbf{X} ; K_i, Q_i\right)[p]=
    \sum_{j=1}^n \alpha_{p j}\left(1+\sigma_{i j}\right) \boldsymbol {u_j}.
\end{equation}

Then we calculate the term {\small$\left|\mathbb{I}_n^\top \varphi_i\left(\mathbf{X} ; K_i, Q_i\right)[p]\right|$} as follows,
{\small\begin{subequations}
  \begin{align}
      \left|\mathbb{I}_n^\top \varphi_i\left(\mathbf{X} ; K_i, Q_i\right)[p]\right|
     &=  \left|\mathbb{I}_n^\top (\sum_{j=1}^n \alpha_{p j}\left(1+\sigma_{i j}\right) \boldsymbol {u_j}  \right|
     \label{8a}\\
     &= \left| \sqrt{n} \left( \alpha_{p 1}(1+\sigma_{i 1}) 
\right)\right|  \label{121}\\
     &= 2|\sqrt{n}\alpha_{p1}|\label{8b},
\end{align}
\end{subequations} } 
where \eqref{8a} is induced by substituting the equation \eqref{si2} into $\left|\mathbb{I}_n^\top \varphi_i\left(\mathbf{X} ; K_i, Q_i\right)[p]\right|$, \eqref{121} is due to $\mathbb{I}_n^\top \boldsymbol {u_j}=0 $ for all $j\ne 1$,
\eqref{8b} follows the fact that $\sigma_{i1}=1$ .
    
\end{proof}

\subsection{Proof of Theorem~\ref{theorem:transition layer}}

We first prove the following result. For simplicity of notations, we use $f(\mathbf{X})\left [ p \right ]$ to denote the $p$-th ($1\leq p \leq d$) column of the the distilled model $f(\mathbf{X})$, where the parameters in the $\alpha L$-th layer is replaced with the matrices $\hat{K}_{\alpha L}$ and $\hat{Q}_{\alpha L}$. We use the function $\hat{\varphi}_{\alpha L}(\mathbf{X})=\varphi_{\alpha L}(\mathbf{X};\hat{K}_{\alpha L}, \hat{Q}_{\alpha L})$ to denote the mapping of the $(\alpha L)$-th layer. Then we are going to show that there exists 
$\alpha^{\star}=\log_{2}{\frac{2}{1+\beta_D} }$ and $0<\beta_D<1$ makes the following equations hold.

(1) Assume $\alpha<\alpha^{\star}$. For any $\mathbf{X}$, 
  $\left\|K_i\right\|_2\leq D, \left\|Q_i\right\|_2\leq D$, there exists  a zero measure set $\mathcal{K}(\mathbf{X})$ and $\mathcal{Q}(\mathbf{X})$ such that 
  \begin{equation}\label{thm1}
      \lim_{L \to \infty} \left\|\frac{f(\mathbf{X})\left [ p \right ] }
{\left\|f(\mathbf{X})\left [ p \right ]\right\|_2} - \frac{\mathbb{I}_n}{\sqrt{n} } 
\right\|_2=0.
  \end{equation}
(2) For any $\alpha>\alpha^{\star}$, there exists a sequence of matrix $\{K_i, Q_i\}_{i\ge1}$ such that for any distilled matrix $K_{\alpha L}$ and $Q_{\alpha L}$, we have 
  $\left\|K_i\right\|_2\leq D, \left\|Q_i\right\|_2\leq D$, we have,
  \begin{equation}\label{thm2}
      \lim_{L \to \infty} \left\|\frac{f(\mathbf{X})\left [ p \right ] }
{\left\|f(\mathbf{X})\left [ p \right ]\right\|_2} - \frac{\mathbb{I}_n}{\sqrt{n} } 
\right\|_2 =\sqrt{2}.
  \end{equation}

\begin{proof}
    Based on Lemma \eqref{v}, we obtain that 
\begin{equation}\label{88}
\begin{split}
&\max_{\boldsymbol{v}:\left\|\boldsymbol{v}\right\|_2=1,\;\boldsymbol{v}\perp \mathbb{I}_n} 
\left| \boldsymbol{v}^{\top} f\left(\mathbf{X}\right)[p] \right|
\\
&\leq\;(1+\beta)^L\max_{\boldsymbol{v}:\left\|\boldsymbol{v}\right\|_2=1,\;\boldsymbol{v}\perp \mathbb{I}_n} 
\left| \boldsymbol{v}^{\top} \mathbf{X}[p] \right|.
\end{split}
\end{equation}
Based on Lemma \eqref{self-att}, we know that
{\small\begin{equation}\label{99}
\begin{split}
&\left| \mathbb{I}_n^{\top} f(\mathbf{X})[p] \right|
\\&=\,  2^{(1-\alpha)L - 1} \left| \mathbb{I}_n^{\top} \hat{\varphi}_{\alpha L} \circ \varphi_{\alpha L-1} \circ \cdots \circ \varphi_{1}(\mathbf{X})[p] \right|.
\end{split}
\end{equation}}
We firstly prove the equation \eqref{thm1}. When 
\begin{equation}\label{99-a}
    \left | \mathbb{I}_n^{\top}f(\mathbf{X})[p]  \right |\neq 0,
\end{equation}
then we have 
{\small\begin{subequations}
    \begin{align}
        &\left\|\frac{f(\mathbf{X})\left [ p \right ] }
{\left\|f(\mathbf{X})\left [ p \right ]\right\|_2} - \frac{\mathbb{I}_n}{\sqrt{n} } 
\right\|_2\\
&=\left[2-\frac{2\mathbb{I}_n^{\top} f(\mathbf{X})[p]}
{\sqrt{n}
\sqrt{\frac{(\mathbb{I}_n^{\top} f(\mathbf{X})[p])^2}{n} +(\boldsymbol {v}^{\top} f(\mathbf{X})[p])^2)} }
\right]^{1/2}\\
& = \sqrt{2}\left[1-\frac{1}
{\sqrt{1+\frac{n(\boldsymbol {v}^{\top} f(\mathbf{X})[p])^2}
{(\mathbb{I}_n^{\top} f(\mathbf{X})[p])^2} } }
\right]^{1/2}\\
&\leq \sqrt{2} \left[1-\frac{1}
{\sqrt{1+\frac{n(1+\beta)^{2L}|\boldsymbol {v}^{\top} \mathbf{X}[p]|^2}
{2^{2[(1-\alpha )L-1]}
\left | \mathbb{I}_n^{\top}\hat{\varphi}_{\alpha L}  
\circ \cdots \circ \varphi_{1}(\mathbf{X})[p]  \right |^2} } }
\right]^{1/2}\label{10cc}
\\
& \leq 2\sqrt{2n}
\left (\frac{1+\beta}{2^{1-\alpha}}   \right ) ^L 
\frac{\left|\boldsymbol {v}^{\top} \mathbf{X}[p]\right|}{\left | \mathbb{I}_n^{\top}\hat{\varphi}_{\alpha L} \circ \varphi_{\alpha L-1 } 
\circ \cdots \circ \varphi_{1}(\mathbf{X})[p]  \right |} \label{10dd},
    \end{align}
\end{subequations}}
where the inequality \eqref{10cc} is based on the inequality \eqref{88} and \eqref{99}. The inequality \eqref{10dd} is based on Lemma \eqref{t11}.
Therefore, if $\alpha<\log_{2}{\frac{2}{1+\beta_D} }$ and $\left | \mathbb{I}_n^{\top}f(\mathbf{X})[p]  \right |\neq 0$, then we have 
 $\lim_{L \to \infty}  \left (\frac{1+\beta_D}{2^{1-\alpha}}   \right ) ^L =0$. Now we can consider when $ \left | \mathbb{I}_n^{\top}f(\mathbf{X})[p]  \right |=0.$ In fact, it is easy to show that this can only happens when $\hat{K}_{\alpha L}$ and $\hat{Q}_{\alpha L}$ belong to  certain sets making  $ \left | \mathbb{I}_n^{\top}f(\mathbf{X})[p]  \right |=0,$ which corresponds to zero measure set $\mathcal{K}(\mathbf{X})$ and $\mathcal{Q}(\mathbf{X})$ depending on the input $\mathbf{X}$. Since the input space is countable, therefore, the union $\cup_{\mathbf{X}\in\mathcal{X}}\mathcal{K}(\mathbf{X})$  and $\cup_{\mathbf{X}\in\mathcal{X}}\mathcal{Q}(\mathbf{X})$ are also zero-measure sets.

To prove equation \eqref{thm2}, let  $K^{\star}$, $Q^{\star}$ with 
$\left\|K^{\star}\right\|_2\leq D, \left\|Q^{\star}\right\|_2\leq D$ satisfy the following condition,
{\small\begin{equation}\label{max}
    \max_{\boldsymbol {v}:\left\|\boldsymbol {v}\right\|_2=1, \boldsymbol {v}\perp \mathbb{I}_n  } 
{\left\|\operatorname{softmax}\left(\frac{\mathbf{X}Q_l (\mathbf{X}K_l)^\top}{\sqrt{d_Q}\|\mathbf{X}\|^2}\right)
\boldsymbol {v}\right\|_2}
=\beta_D.
\end{equation}}
Let $\boldsymbol {v}^{\star}$ be the solver of the above optimization problem \eqref{max} and consider the $K_l=K^{\star}$, $Q_l=Q^{\star}$ and $\mathbf{X}^\star=[\boldsymbol {v}^{\star}, \boldsymbol {v}^{\star}, \cdots,\boldsymbol {v}^{\star}]$. Clearly, $\bm{v}^\star\perp \mathbb{I}_n.$ 
Assume there exists $\boldsymbol{u}: \left \|\boldsymbol{u}^{\star}  \right \| _2=1$ satisfying 
$\boldsymbol {u}^{\star}\perp \mathbb{I}_n$, $\boldsymbol {u}^{\star}\perp \boldsymbol {v}^{\star}$, therefore we can rewrite $f(\mathbf{X}^\star)\left [ p \right ]$ as follows,
{\small \begin{equation}\label{factor1}
\begin{split}
f(\mathbf{X}^\star)\left [ p \right ]&=\frac{\mathbb{I}_n^{\top}}{\sqrt{n} } 
f(\mathbf{X}^\star)\frac{\mathbb{I}_n}{\sqrt{n} }
\\
&+\boldsymbol {v}^{\star\top}f(\mathbf{X}^\star)\boldsymbol {v}^{\star}
+\boldsymbol {u}^{\star\top}f(\mathbf{X}^\star)\boldsymbol {u}^{\star}.
\end{split}
\end{equation}}

For any $1\leq l \leq L $, based on Lemma \eqref{v}, we know that
{\small\begin{equation}\label{maxmax1}
    \left | \boldsymbol {v}^{*\top}f\left(\mathbf{X}^\star\right)[p]  \right | 
= (1+\beta_D)^L
\left | \boldsymbol {v}^{*\top}\mathbf{X}^\star[p]  \right |.
\end{equation}}
Since 
{\small\begin{equation}\label{maxmax2}
    \left | \mathbb{I}_n^\top f\left(\mathbf{X}^\star\right)[p]  \right | 
= 2^L
\left | \mathbb{I}_n^\top\mathbf{X}^\star[p]  \right |= |\mathbb{I}_n^\top \bm{v}^\star|=0
\end{equation}}
and 
{\small\begin{equation}\label{maxmax}
    \left | \boldsymbol {v}^{*\top}f\left(\mathbf{X}^\star\right)[p]  \right | 
= (1+\beta_D)^L
\left | \boldsymbol {v}^{*\top}\mathbf{X}^\star[p]  \right |\neq 0.
\end{equation}}
Then we have
{\small
\begin{subequations}
    \begin{align}
         &\left\|\frac{f(\mathbf{X}^\star)\left [ p \right ] }
{\left\|f(\mathbf{X}^\star)\left [ p \right ]\right\|_2} - \frac{\mathbb{I}_n}{\sqrt{n} } 
\right\|_2\\
&=\left[2-\frac{2\mathbb{I}_n^{\top} f(\mathbf{X}^\star)[p]}
{\sqrt{n}
\left\|f(\mathbf{X}^\star)\left [ p \right ]\right\|_2 }
\right]^{1/2} \\
&= \left[
2 - \frac{2\mathbb{I}_n^{\top}}{\sqrt{n}} \cdot 
\frac{f(\mathbf{X}^\star)[p]}{
\sqrt{
\begin{array}{l}
\frac{1}{n} \left(\mathbb{I}_n^{\top} f(\mathbf{X}^\star)[p] \right)^2 +
\left(\boldsymbol{v}^{\star\top} f(\mathbf{X}^\star)[p] \right)^2 \\[0.5ex]
+ \left(\boldsymbol{u}^{\star\top} f(\mathbf{X}^\star)[p] \right)^2
\end{array}
}
}
\right]^{1/2} \label{22b}\\
& \ge \left[2-\frac{2\mathbb{I}_n^{\top}}{\sqrt{n}} \frac{ f(\mathbf{X}^\star)[p]}
{\sqrt{\frac{1}{n} (\mathbb{I}_n^{\top}f(\mathbf{X}^\star)[p])^2+
(\boldsymbol {v}^{\star\top}f(\mathbf{X}^\star)[p])^2} }
\right]^{1/2}\\
&=\left[2-2 \frac{ \frac{\mathbb{I}_n^{\top}f(\mathbf{X}^\star)[p]}{\sqrt{n} 
\left |  \boldsymbol {v}^{\star\top}f(\mathbf{X}^\star)[p]\right | } }
{\sqrt{1+\frac{\left | \mathbb{I}_n^{\top}f(\mathbf{X}^\star)[p] \right |^2}
{n\left | \boldsymbol {v}^{\star\top}f(\mathbf{X}^\star)[p]) \right |^2} } }
\right]^{1/2}\\
&= \left[2-2 \frac{ \frac{2^{(1-\alpha )L-1}
\left | \mathbb{I}_n^{\top}\hat{\varphi}_{\alpha L} \circ \varphi_{\alpha L-1 } 
\circ \cdots \circ \varphi_{1}(\mathbf{X}^\star)[p]  \right |}
{\sqrt{n} (1+\beta_D)^L
\left |  \boldsymbol {v}^{\star\top}\mathbf{X}^\star[p]\right | } }
{\sqrt{1+\frac{2^{2[(1-\alpha )L-1]}}{n(1+\beta_D)^{2L}} \frac{\left | 
\mathbb{I}_n^{\top}\hat{\varphi}_{\alpha L} \circ \varphi_{\alpha L-1 } 
\circ \cdots \circ \varphi_{1}(\mathbf{X}^\star)[p]  \right |^2}
{\left | \boldsymbol {v}^{\star\top}\mathbf{X}^\star[p] \right |^2} } }
\right]^{1/2}\label{22e},
    \end{align}
\end{subequations}}
where equation \eqref{22b} is based on \eqref{factor1}, equation \eqref{22e} is based on \eqref{maxmax} and \eqref{99}.
When $\alpha>\log_{2}{\frac{2}{1+\beta_D} }$, we have 
 $\lim_{L \to \infty}  \left (\frac{2^{1-\alpha}}{1+\beta_D}   \right ) ^L =0$. Thus we have $\lim_{L \to \infty} \left\|\frac{f(\mathbf{X}^\star)\left [ p \right ] }
{\left\|f(\mathbf{X}^\star\left [ p \right ]\right\|_2} - \frac{\mathbb{I}_n}{\sqrt{n} } 
\right\|_2 =\sqrt{2}$.
This indicates that the $p$-th column of the output matrix $f(\mathbf{X}^\star)$ is not parallel to $\mathbf{I}_n$ for any $p$. This further indicates that the output matrix does not have the identical vector in each row. 
\end{proof}

\subsection{Technical Lemma}

\begin{Lem}\label{t11}
For any $x\in(0,1)$, it always holds $\left [ 1-\frac{1}{\sqrt{1+x^2} }  \right ]^{1/2}\leq x $.
\end{Lem}
\begin{proof}
To establish the inequality $\left [ 1-\frac{1}{\sqrt{1+x^2} } \right ]^{1/2}\leq x $, we begin by proving,
\begin{equation}\label{tlem1}
1-\frac{1}{\sqrt{1+x^2} } \leq x^2.
\end{equation}
To demonstrate \eqref{tlem1}, we equivalently show
\begin{equation}
1-x^2 \leq \frac{1}{\sqrt{1+x^2} }.
\end{equation}
Subsequently, it suffices to verify
\begin{equation}
(1-x^2) (\sqrt{1+x^2}) \leq 1.
\end{equation}
This is equivalent to proving
\begin{equation}
(1-x^2)^2 (1+x^2) \leq 1.
\end{equation}
Thus, our focus shifts to demonstrating
\begin{equation}\label{final}
(1-x^2) (1-x^4) \leq 1.
\end{equation}
Clearly, \eqref{final} holds true for any $x\in(0,1)$.
\end{proof}

\subsection{Remarks}
\label{append:explor_remarks}

\textbf{Remark 2:}  The existence of \(\hat{f}_\infty(\mathbf{X})\) is a non-trivial  result. While the mapping \(\varphi_i\) admits a fixed point at \(\mathbf{X} = \mathbf{0}_{n \times d}\), the convergence of the iterative process governed by \(\varphi_i\) cannot be guaranteed using the contraction mapping theorem, as \(\varphi_i\) does not satisfy the contraction property for any pair \((Q_i, K_i)\). This complexity becomes particularly evident in the special case where \(n = 1\) and \(\mathbf{X}\) is a column vector. Here, the output of \(\varphi_i\) satisfies the relation \(\langle \mathbf{1}_d, \varphi_i(\mathbf{X}; K_i, Q_i) \rangle = 2 \langle \mathbf{1}_d, \mathbf{X} \rangle\), implying that the iteration diverges unless \(\mathbf{X}\) is orthogonal to \(\mathbf{1}_d\). However, the divergence is not arbitrary; rather, the theorem reveals that it occurs in a fixed, well-defined direction. This insight ensures the existence of a normalized output, which remains stable and meaningful despite the lack of strict convergence.

\textbf{Remark 3:} The existence of \(\alpha^* \in (0,1)\) is also a non-trivial statement, as \(\alpha^*\) could potentially be zero, which would imply the absence of a critical layer such that securing layers prior to it guarantees the failure of the recovered model's functionality. The primary challenge lies in demonstrating that perturbations to the earlier layers result in rank-one outputs, a property that does not universally hold for arbitrary perturbations. To address this, we establish an alternative result: given an input matrix \(\mathbf{X}\), rank-one outputs can be guaranteed if the perturbation matrices \(K_i\) and \(Q_i\) are chosen to avoid specific zero-measure sets, denoted as \(\mathcal{K}(\mathbf{X})\) and \(\mathcal{Q}(\mathbf{X})\), respectively. Assuming a countable domain \(\mathcal{X} \times \mathcal{Y}\), which is typical for structured inputs such as sentences or images, it follows that the perturbation matrices to be avoided belong to the countable union of these sets, defined as \(\mathcal{K} = \bigcup_{\mathbf{X} \in \mathcal{X}} \mathcal{K}(\mathbf{X})\) and \(\mathcal{Q} = \bigcup_{\mathbf{X} \in \mathcal{X}} \mathcal{Q}(\mathbf{X})\). Since this union remains a zero-measure set, avoiding these specific sets ensures that the conditions of the theorem are satisfied for any input matrix \(\mathbf{X}\).

\newpage
\section{Experiment Details}
\label{append:Experiment}
To more intuitively compare the security differences between the SOLID method and a fully-secured approach, we define \(\Delta\textbf{ADR}(I) = \text{ADR}(I) - \text{ADR}([L])\) to assess the resilience of the secured set \(I\) relative to the fully-secured approach. A smaller value of \(\Delta\textbf{ADR}\) indicates resilience similar to that of the fully-secured model.

\subsection{Model Details.}
\label{append:Model Details}

The foundation models we use in our experiments are selected from open-source repositories, and Table~\ref{tab:model detials} shows the basic information of the models and their sources. Specifically, we employ Llama2-70B-chat\footnote{https://huggingface.co/meta-llama/Llama-2-70b-chat-hf}~\citep{llama2}, Llama2-7B-chat\footnote{https://huggingface.co/meta-llama/Llama-2-7b-chat-hf}~\citep{llama2}, and Mistral-7B-v0.1\footnote{https://huggingface.co/mistralai/Mistral-7B-v0.1}~\citep{mistral}. For smaller models, we select Phi-2\footnote{https://huggingface.co/microsoft/phi-2}~\citep{phi3} and Phi-1.5\footnote{https://huggingface.co/microsoft/phi-1\_5}~\citep{phi1-5}. We also consider OPT model\footnote{https://huggingface.co/facebook/opt-350m}~\citep{opt-350}, which has only 350 million parameters and 24 decoder layers.

\begin{table}[ht]
    \centering
    \begin{tabular}{llc}
    \toprule
        \textbf{Model} & \textbf{Size} & \textbf{Decoder Layers} \\
    \midrule
    Llama2-70B-chat & 70B & 80  \\
        Llama2-7B-chat  & 7B & 32  \\
        Mistral-7B-v0.1 & 7B & 32  \\
        Phi-2  & 2.7B & 32  \\
        Phi-1.5  & 1.3B & 24  \\
        OPT & 350M & 24  \\
        \bottomrule
    \end{tabular}
   \caption{Model Info}
    \label{tab:model detials}
\end{table}

\subsection{Distillation Attacks.} 
\label{append:Datasets Details}

\textbf{Attack implementation details.} In performing FT-all and FT-secure model distillation attacks, we adhere to the training hyper-parameters outlined in the Llama2 report~\citep{llama2}, employing the AdamW optimizer with a cosine learning rate scheduler. The initial learning rate is set to $2 \times 10^{-5}$, with a weight decay of 0.1, a batch size of 128, and bfloat16 precision for input sequences of 512 tokens. The LLaMA2-70B model is trained for 3 epochs with a random seed of 42, while other models are trained for 5 epochs across three seeds: 42, 1234, and 20. Despite limiting training to 3 epochs for the 70B model, the training loss stabilized effectively. Our implementation builds upon the \href{https://github.com/meta-llama/llama-recipes}{llama-recipes} repository provided by META.

For SEM attacks, distinct configurations were employed for SOLID and SAP-DP. In the case of SOLID, hidden representations from the secure-source components were collected and paired with the input data to train a substitute model. In contrast, for SAP-DP, representations from the sixth decoder layer and the model's final logits were utilized to construct the training dataset. In accordance with~\citep{canHideApi}, we applied a learning rate of 1.5e-4, a weight decay of 0.01, and a linear learning rate scheduler with 500 warmup steps. Both training and validation batch sizes were set to 32, with MSE as the loss function. SOLID was trained for 30 epochs due to its smaller model size, whereas SAP-DP was trained for 5 epochs.

All distillation experiments were conducted on Nvidia 4090 24G, 6000 Ada 48G, and A100 80G GPUs, utilizing PyTorch 2.2.0 and CUDA 11.8 on Ubuntu 20.04.6 LTS.

\textbf{Base 51k Distillation Dataset.}
We ensure dataset coverage and reliability by using a 1:1 ratio of the MMLU~\footnote{https://github.com/hendrycks/test}~\citep{mmlu} auxiliary training set and Alpaca dataset~\footnote{https://github.com/tatsu-lab/stanford\_alpaca/blob/main/alpaca\_data.json}~\citep{stanfordAlpaca}, extracting 25.5k samples from each. From the MMLU auxiliary training data, we sample 50\%, and from Alpaca, we use a step size of 2 to enhance diversity. The datasets are then formatted for model training, applying Alpaca and MMLU prompts from Table~\ref{tab:combined-prompts}.

\begin{table*}[t]
    \centering
    \small
    \begin{tabular}{c c p{0.6\textwidth}}
    \toprule
        \textbf{Dataset} & \textbf{Prompt Type} & \textbf{Description} \\
    \toprule
        \multirow{5}{*}{\textbf{Alpaca}} & \multirow{3}{*}{with input}  & Below is an instruction that describes a task, paired with an input that provides further context. Write a response that appropriately completes the request. \\
        \cmidrule(r){2-3}
        & \multirow{2}{*}{w/o input} & Below is an instruction that describes a task. Write a response that appropriately completes the request. \\
    \toprule
        \multirow{5}{*}{\textbf{MMLU}} & \multirow{2}{*}{Question Answering} & Below is a question with no choices. Write the correct answer that appropriately solves the question. \\
    \cmidrule(r){2-3}
        & \multirow{2}{*}{Multiple Choice} & The following is a multiple choice question, paired with choices. Answer the question in the format: ``Choice:content''. \\
    \bottomrule
    \end{tabular}
   \caption{Prompts for Alpaca and MMLU auxiliary training data}
   \label{tab:combined-prompts}
\end{table*}

\textbf{Extra Distillation Datasets.}
\label{append:Extra Distillation Datasets}
To enhance dataset diversity, the 100K, 200K, 300K, and 500K datasets integrate additional specialized sources. As detailed in Table~\ref{tab:datasets-composition}, these sources include Baize~\citep{baize} (158K English multi-turn conversations via ChatGPT’s self-chat), MathInstruct~\citep{mathInstruct} (260K curated math instruction instances focusing on hybrid reasoning), and OpenOrca~\citep{orca} (augmented FLAN collection with 1M GPT-4 completions and 3.2M GPT-3.5 completions). These enrichments are intended to support complex computational and theoretical tasks, offering broader topic coverage.

\begin{table*}[t]
    \centering
    \begin{tabular}{lccccc}
    \toprule
    \textbf{Raw Data Set} & \textbf{51k} & \textbf{100k} & \textbf{200k} & \textbf{300k} & \textbf{500k} \\
        \midrule
        \href{https://github.com/tatsu-lab/stanford\_alpaca/blob/main/alpaca\_data.json}{Alpaca}             & 25.5  & 50   & 40   & 50   & 50   \\
        \href{https://huggingface.co/datasets/cais/mmlu}{MMLU auxiliary training set}      & 25.5  & 50   & 40   & 100  & 100  \\
        \href{https://github.com/project-baize/baize-chatbot/blob/main/data/medical\_chat\_data.json}{Baize-MedQuAD}                       & 0   & 0    & 40   & 50   & 50   \\
        \href{https://github.com/project-baize/baize-chatbot/blob/main/data/quora\_chat\_data.json}{Baize-Quora}                       & 0   & 0    & 40   & 50   & 50   \\
        \href{https://github.com/project-baize/baize-chatbot/blob/main/data/stackoverflow\_chat\_data.json}{Baize-Stackoverflow}                    & 0   & 0    & 40   & 50   & 50   \\
        \href{https://huggingface.co/datasets/TIGER-Lab/MathInstruct}{MathInstruct}                      & 0   & 0    & 4    & 6    & 20   \\
        \href{https://huggingface.co/datasets/Open-Orca/OpenOrca}{OpenOrca}                       & 0   & 0    & 0    & 0    & 180  \\
        \bottomrule
    \end{tabular}
    
   \caption{Composition of variously sized datasets}
    \label{tab:datasets-composition}
\end{table*}

\textbf{Validation Datasets.} 
Table~\ref{tab:val_dataset} outlines the composition of the validation datasets. For \textit{Validation Dataset 1}, we extracted 50\% from each of the 57 MMLU validation sub-datasets, totaling 1.5K instances, paired with Alpaca data selected using a step size of 751. This dataset is used with the 51K and 100K training sets. For larger training sets (200K, 300K, and 500K), \textit{Validation Dataset 2} was created by adding 400 instances from three Baize subsets, expanding the validation set to 4.0K.

\begin{table*}[t]
    \centering
    \begin{tabular}{lcc}
    \toprule
        \textbf{Raw Data Set} & \textbf{Validation Set} & \textbf{Evaluation Set} \\ 
        \midrule
        \href{https://github.com/tatsu-lab/stanford\_alpaca/blob/main/alpaca\_data.json}{Alpaca} & 765 & 765  \\ 
        \href{https://huggingface.co/datasets/cais/mmlu}{MMLU auxiliary training set} & 751 & 751  \\ 
        \href{https://github.com/project-baize/baize-chatbot/blob/main/data/medical\_chat\_data.json}{Baize-MedQuAD} & 0 & 850 \\ 
        \href{https://github.com/project-baize/baize-chatbot/blob/main/data/quora\_chat\_data.json}{Baize-Quora} & 0 & 850 \\ 
        \href{https://github.com/project-baize/baize-chatbot/blob/main/data/stackoverflow\_chat\_data.json}{Baize-Stackoverflow} & 0 & 850 \\
        \midrule
        \textbf{Total Length} & 1516 & 4066 \\ 
    \bottomrule
    \end{tabular}
   \caption{Composition of validation datasets of different sizes}
    \label{tab:val_dataset}
\end{table*}

\subsection{Baselines.}
\label{appendix:baselines}
In this section, we provide further details on the baselines used in our comparisons: SAP-DP and fully-secured. These schemes represent different strategies, each with distinct trade-offs in terms of customizability and security against model distillation attacks.

\textbf{SAP.} The Split-and-Privatize (SAP) framework~\citep{SAP} offers an approach to balance between protecting model privacy and data privacy while maintaining competitive performance. Specifically, the SAP framework keeps the bottom six encoder layers open, allowing user access and fine-tuning while securing the deeper layers on the vendor.

\textbf{SAP-DP.} To further strengthen protection while maintaining competitive performance, we extend SAP by incorporating differential privacy techniques by adding Laplace noise to perturb the logits during the fine-tuning process~\citep{SAP-DP-supp}.
The Laplace Distribution with mean $\mu$ and scale $b$ is the distribution with probability density function:
\[
\text{Laplace}(x|\mu,b) = \frac{1}{2b} \exp\left(-\frac{|x-\mu|}{b}\right)
\]
Specifically, in SAP-DP, the noise $n$ is sampled: $n \sim \text{Laplace}(0, 0.5)$ and added to the output logits of the model to balance privacy protection and model performance.

\textbf{Fully-secured.} Following~\citep{Semiopen1}, we use the fully-secured approach as a baseline. This assumes the adversary has no access to internal model parameters, treating the model as a black-box, where only output data can be collected. We slightly broaden this setup by assuming the adversary knows the model's architecture but no other details. Thus, distilling the fully-secured model involves using the collected data to retrain a model with the same architecture to restore its general functionality.

\textbf{DarkneTZ.} Based on the work of \citep{mo2020darknetz}, we use DarkneTZ as a baseline to test whether protecting only the output layers is sufficient to defend against distillation attacks. In this setup, we assume the adversary has no access to the model parameters of the output layers, specifically the last decoder layer. Similar to the SAP framework, this approach allows the adversary to access and fine-tune all layers except the final decoder layer.

\subsection{Implementation Details of SOLID.}
\label{appendix:Implementation Details}

\textbf{Evaluation Datasets.} We created a 1.5K Evaluation Set to assess model security under various secure-sourcing strategies. This set includes 50\% of entries from each of the 57 MMLU validation sub-datasets~\citep{mmlu}, distinct from Validation Set outlined in Table~\ref{tab:val_dataset}. Additionally, we selected an equal number of Alpaca dataset~\citep{stanfordAlpaca}, using a step size of 751, ensuring no overlap with the Validation Set.

\begin{figure}[h]
    \centering
    \includegraphics[width=0.7\linewidth]{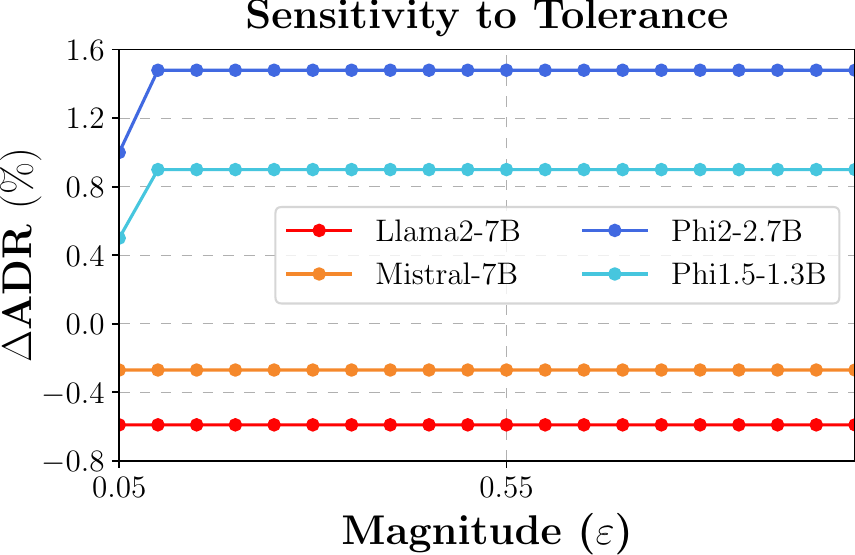}
    \caption{Sensitivity on $\varepsilon$.}
    \label{fig:appendix:Sensitivity-SOLID-appendix}
\end{figure}

\textbf{Hyper-parameter Sensitivity.} 
As shown in Figure~\ref{fig:appendix:Sensitivity-SOLID-appendix}, we evaluate SOLID's sensitivity to tolerance magnitude $\varepsilon$, adjusting it from 0.05 to 1 in 0.05 increments while calculating the $\Delta\text{ADR}$ for six distilled models. The results indicate that SOLID is minimally sensitive to changes in $\varepsilon$, with $\Delta\text{ADR}$ values stabilizing as $\varepsilon$ increases. This stability arises from the need for a smaller secured layer at higher $\varepsilon$, allowing the condition $R(I) \leq (1+\varepsilon)R([L])$ to be met with fewer layers. Additionally, the increase in $\Delta\text{ADR}$ is smaller for larger models, suggesting that privatizing more parameters beyond a certain point offers diminishing returns in security.

\subsection{Evaluation Benchmarks}
\label{appendix:evaluation datasets}
Most of our evaluations are conducted using the \href{https://github.com/EleutherAI/lm-evaluation-harness}{lm-evaluation suite}~\citep{eval-harness}, the \href{https://github.com/bigcode-project/bigcode-evaluation-harness}{bigcode-evaluation-harness platform}~\citep{bigcode-evaluation-harness}, and \href{https://github.com/lm-sys/FastChat/tree/main}{MT-Bench}~\citep{MT-Bench}. For specific domains, such as finance and law, we utilize the official benchmark testing codes provided by their respective communities, as detailed below.

\textbf{Evaluation on Customizabilities.}
We assess the customizability of models across six domains, as detailed in Table~\ref{fig:custom Benchmarks}. Each domain includes specific benchmarks and metrics designed to evaluate different aspects of the model's performance in relation to customizability. In particular, for evaluating medical capabilities, we select two subcategories from the MMLU benchmark that are related to the medical domain: \textit{mmlu\_anatomy} and \textit{mmlu\_professional\_medicine}. For assessing legal reasoning, we select 10 multiple-choice and judgment-based subcategories from Legalbench. The performance of the model in these legal tasks is measured using perplexity, following the prompt structure provided by Legalbench. Specifically, the selected subcategories include:
\begin{itemize}[leftmargin=*, itemsep=0pt, parsep=0pt, topsep=0pt]
    \item \textit{cuad\_audit\_rights}
    \item \textit{canada\_tax\_court\_outcomes}
    \item \textit{definition\_classification}
    \item \textit{cuad\_affiliate\_license-licensee}
    \item \textit{learned\_hands\_business}
    \item \textit{contract\_nli\_survival\_of\_obligations}
    \item \textit{contract\_nli\_explicit\_identification}
    \item \textit{contract\_nli\_confidentiality\_of\_agreement}
    \item \textit{hearsay}
    \item \textit{contract\_qa}
\end{itemize}

\begin{table*}[t]
\centering
\label{tab:customizability-domains}
\begin{tabular}{lllcc}
\toprule
\textbf{Domain}  & \textbf{Benchmark} & \textbf{Metric} & \textbf{n-shot} & \textbf{Reference}\\ \midrule
       \multirow{2}{*}{\textbf{Code}} & HumanEval & Pass@1 & 0  & \cite{humanEval} \\ 
        & MBPP & Pass@1 & 1  & \cite{mbpp} \\ 
       \midrule
       \textbf{Math} & GSM8K & Exact Match & 8  & \cite{gsm8k} \\ 
       \midrule
\textbf{Medical} & MMLU\_Medical& Accuracy& 5 &\cite{mmlu} \\ \midrule
\textbf{Finance}  &FPB &F1&0&\cite{FinGpt} \\ \midrule
\textbf{Law}      &LegalBench &Accuracy&0&\cite{legalbench} \\ \midrule
\textbf{Alignment}  &MT-Bench &Score& (GPT-4) &\cite{MT-Bench} \\ \bottomrule
\end{tabular}
\caption{Details of the Six Customizability Benchmarks}
\label{fig:custom Benchmarks}
\end{table*}

\textbf{Evaluation on Security.}
We follow the Llama-2 report~\cite{llama2} to evaluate the distilled model, including 16 benchmarks, which are categorized into 6 groups. Table~\ref{tab:benchmark datasets2} summarizes the functionality benchmarks used in our experiments, along with their test methods and performance metrics. Our model ranks choices in multiple-choice tasks and generates answers for open-ended generation tasks.

\begin{table*}[t]
   \centering
   \small
   \begin{tabular}{lllcc}
   \toprule
       \textbf{Domain} & \textbf{Benchmark} & \textbf{Metric} & \textbf{n-shot} & \textbf{Reference} \\ 
       \midrule
        \multirow{5}{*}{\textbf {Commonsense Reasoning}} & PIQA & Accuracy & 0  & \cite{PIQA} \\ 
        & Hellaswag & Accuracy & 0  & \cite{hellaswag} \\ 
        & Winogrande & Accuracy & 0  & \cite{sakaguchi2019winogrande} \\ 
        & ARC\_easy & Accuracy & 0  & \cite{ARC} \\ 
        & ARC\_challenge & Accuracy & 0  & \cite{ARC} \\ 
       \midrule
        \multirow{5}{*}{\textbf {Reading Comprehension}} & OpenBookQ & Accuracy & 0  & \cite{OpenBookQA2018} \\ 
        & LAMBADA & Accuracy & 0  & \cite{LAMBADA} \\ 
        & BoolQ & Accuracy & 0  & \cite{clark2019boolq} \\ 
        & SQuADv2 & HasAns\_EM & 2  & \cite{SQuAD} \\ 
        & SQuADv2 & HasAns\_F1 & 2  & \cite{SQuAD} \\ 
       \midrule
       \multirow{2}{*}{\textbf {World Knowledge}} & NaturalQuestions & Exact Match & 5  & \cite{nq_open} \\ 
        & TriviaQA & Exact Match & 5  & \cite{TriviaQA2017} \\ 
       \midrule
       \multirow{2}{*}{\textbf{Code}} & HumanEval & Pass@1 & 0  & \cite{humanEval} \\ 
        & MBPP & Pass@1 & 1  & \cite{mbpp} \\ 
       \midrule
       \textbf{Math} & GSM8K & Exact Match & 8  & \cite{gsm8k} \\ 
       \midrule
       \multirow{2}{*}{\textbf{General Ability}} & MMLU & Accuracy & 5  & \cite{mmlu} \\ 
       & BBH & Accuracy & 3  & \cite{bbh} \\ 
   \bottomrule
   \end{tabular}
   \caption{Details of the Sixteen Functionality Benchmarks}
   \label{tab:benchmark datasets2}
\end{table*}

\subsection{Model Customization}
\label{append:Customization Training Set}
\textbf{Datasets.}
To fine-tune the models for domain-specific tasks, we utilized several datasets tailored to different sectors, including Code~\citep{CodeInstruct}, Math~\citep{mathInstruct},  Medical~\citep{medicalInstruct}, Finance~\citep{FinanceInstruct}, Law~\citep{guha2024legalbench}, and Alignment~\citep{simpo}. Table~\ref{tab:Customization Training Datasets} lists the customization training datasets used in the experiments. For the code domain, we combine the datasets from \href{https://huggingface.co/datasets/m-a-p/CodeFeedback-Filtered-Instruction}{CodeFeedback} and \href{https://huggingface.co/datasets/lucasmccabe-lmi/CodeAlpaca-20k}{CodeAlpaca}. For law and finance, we merge all training datasets from \href{https://huggingface.co/datasets/Equall/legalbench_instruct?row=0}{Legalbench} and \href{https://huggingface.co/FinGPT}{FinGPT} respectively. These datasets are then prepared for model training using the Alpaca prompts outlined in Table~\ref{tab:combined-prompts}. Additionally, we randomly select 3,000 samples to serve as the validation dataset.

\begin{table*}[t]
\centering
\begin{tabular}{lclcc}
\toprule
\textbf{Domain} & \textbf{Dataset Name}     & \textbf{Size} & \textbf{Reference}  \\ \midrule
\multirow{2}{*}{\textbf{Code}}    & \href{https://huggingface.co/datasets/m-a-p/CodeFeedback-Filtered-Instruction}{CodeFeedback}    &156k     &~\cite{Code-Feedback}        \\  
                 & \href{https://huggingface.co/datasets/lucasmccabe-lmi/CodeAlpaca-20k}{CodeAlpaca}      & 20k &\cite{codealpaca}   \\ \midrule
\textbf{Math} & \href{https://huggingface.co/datasets/TIGER-Lab/MathInstruct}{MathInstruction} & 262K &\cite{mathInstruct}\\ \midrule
\textbf{Medical} & \href{https://huggingface.co/datasets/openlifescienceai/medmcqa}{MedMCQA} &183k & \cite{medicalInstruct} \\ \midrule
\textbf{Law}    & \href{https://huggingface.co/datasets/Equall/legalbench_instruct?row=0}{Legalbench}    &90k  & \cite{legalbench}                   & \\ \midrule
\textbf{Finance} & \href{https://huggingface.co/FinGPT}{FinGPT}      &204k & \cite{FinGpt}  \\ \midrule
\textbf{Alignment}  &  \href{https://huggingface.co/datasets/HuggingFaceH4/ultrafeedback_binarized}{Ultrafeedback}   & 62k& \cite{ultrafeedback}  \\ \bottomrule
\end{tabular}
\caption{Customization Training Datasets Composition}
\label{tab:Customization Training Datasets}
\end{table*}

\textbf{Customization Training Hyperparameters.}
In model customization, we use different hyperparameters depending on the model size. For LLaMA2-70B, we apply QLoRA with the settings outlined in Table~\ref{tab:appendix:custom-Hyperparameters}, while for 7B models, we use LoRA. For smaller models like Phi2 and Phi-1.5, we fine-tune all model parameters.
For LLaMA2-70B, we fine-tune it as a quantized 4-bit model over 1 epoch, starting with a learning rate of $1.5 \times 10^{-6}$. For the 7B models, we train for 3 epochs, with a seed value of 42. The training setup includes a weight decay of 0.1, a batch size of 128, a warmup ratio of 0.03, and input sequences of 512 tokens, following standard experimental practices \citep{LoRA}.
For Phi2 and Phi-1.5, we use the training hyperparameters from the LLaMA2 report. We employ the AdamW optimizer with a cosine learning rate scheduler, starting with a learning rate of $2 \times 10^{-5}$, a weight decay of 0.1, a batch size of 128, and use bfloat16 precision for 512-token input sequences.
Specifically, for alignment, we follow SimPO~\cite{simpo} and set the preference parameters $\beta = 2$ and $\gamma = 1$. The learning rate is $1\times10^{-6}$ for LLaMA2-70B and $5\times 10^{-7}$ for the 7B and smaller models. All experiments are conducted using the \href{https://github.com/hiyouga/LLaMA-Factory}{LLaMA-Factory} on Nvidia 4090 24G, 6000 Ada 48G, and A100 80G GPUs, with PyTorch 2.2.0 and CUDA 11.8 on Ubuntu 20.04.6 LTS.

\begin{table*}[t]
    \centering
    \setlength{\tabcolsep}{3pt}
    \begin{tabular}{lccccccccc}
    \toprule
        \textbf{Model} & \textbf{Method} & \textbf{Rank $r$} & \textbf{Lora $\alpha$} & \textbf{Dropout} & \textbf{Learning Rate} & \textbf{Epochs} & \textbf{Warmup R.}\\ 
        \midrule
        \textbf{Llama2-70B} & QLoRA  & 96 & 16 & 0.05 & 1.50E-04 & 1 & 0.03\\ 
        \textbf{Llama2-7B} & LoRA  & 32 & 64 & 0.05 & 2.00E-05 & 3 & 0.03\\ 
        \textbf{Mistral-7B} & LoRA & 32 & 64 & 0.05 & 1.00E-06 & 3 & 0.03\\ 
        \bottomrule
    \end{tabular}
    \caption{The Hyperparameters for Customization Training.}
    \label{tab:appendix:custom-Hyperparameters}
\end{table*}



\subsection{Security and Customization Transitions}
\label{append::Experiment transition layer}
For the LLaMA2-7B model, the smallest secure-source layer set identified by SOLID consists of a single decoder layer, whereas for Phi-2, it includes two decoder layers. Consequently, for LLaMA2-7B, we opted to secure-source each even-indexed layer, while for Phi-2, we chose to secure-source non-overlapping pairs of layers (e.g., layers 0-1, 2-3). For each selected layer set, we first secure-source them, then subjected the semi-open model to FT-all attacks, and subsequently calculated the $\Delta\text{ADR}$ of the layer set to assess its security.

When verifying the customization transition, due to computational constraints, we validated only every other layer set for both models  (e.g., secure-source layers 0, 0-4, 0-8 …). Specifically, we applied LoRA-based customization on LLaMA2-7B in the math domain, while for Phi-2, we utilized the full finetuning approach. The experimental hyperparameters remain consistent with those outlined in the Appendix~\ref{append:Customization Training Set}.

\begin{figure}[h]
  \centering
  \includegraphics[width=0.9\linewidth]{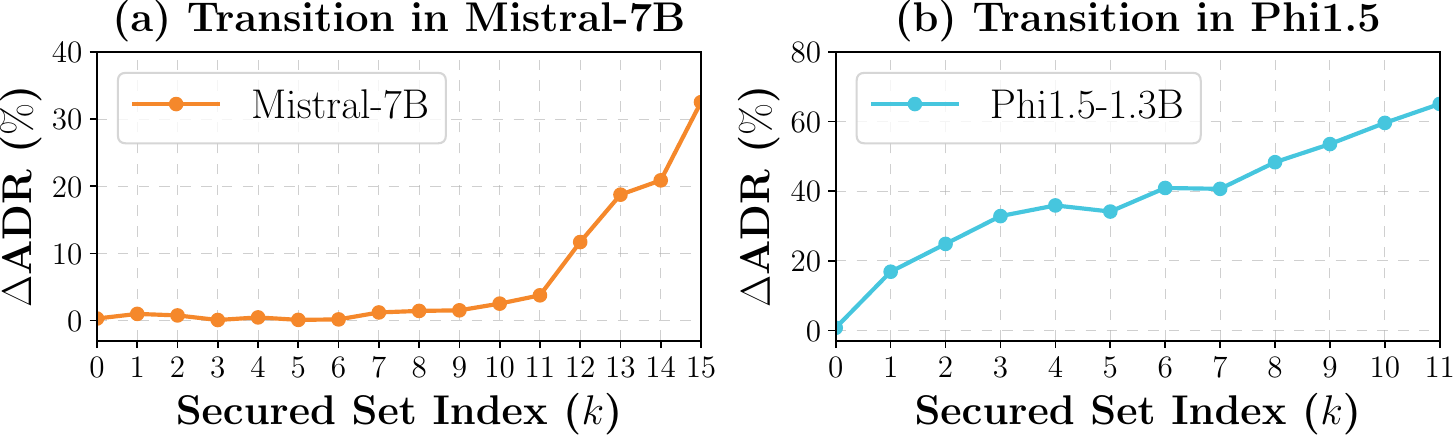}
  \caption{Security changes in Miatral-7B and Phi-1.5.}
  \label{fig:transition-appendix}
\end{figure}

We further computed the $\Delta\text{ADR}$ for each secure-source set within Mistral-7B-v0.1 and Phi-1.5. In these models, the smallest secure-source set identified by SOLID consists of one decoder layer and two decoder layers, respectively. Following the same experimental configuration as LLaMA2-7B and Phi-2, we secured each even-indexed layer for Mistral-7B, and non-overlapping pairs of layers for Phi-1.5. The complete results demonstrating the transition layers within the Mistral-7B and Phi-1.5 model that secure two non-overlapping consecutive layers are depicted in Figure~\ref{fig:transition-appendix}. Once again, we observed a distinct presence of transition layers. Specifically, in Mistral-7B, the transition layer appears at the 24th layer, while in Phi-1.5, it is located within the first layer set. Further results for can be found in Appendix~\ref{append:result:Transition}.

\subsection{Security Across Secure Sizes}
\label{append:Security Across Secure Sizes.}

To examine the influence of Secure layer size on model security, we conduct experiments on Secure-sourcing different amounts and proportions of parameters in the model's decoder layer. We give instructions on the detailed setting of secured models in Table~\ref{appd tab:Priv set}. The module names are all derived from the overall implementation functions of each model in the Transformers open-source repositories in Table~\ref{tab:model detials}. We utilize abbreviated module names to denote specific settings.

\begin{table*}[t]
\centering
    {\tiny
    \setlength\tabcolsep{2.4pt} 
\begin{tabular}{cccccc} 
\toprule
\multicolumn{1}{l}{\textbf{}}         & \multicolumn{1}{l}{\textbf{}} & \textbf{Llama-7B} & \textbf{Mistral-7B} & \textbf{Phi2-2.7B} & \textbf{Phi1.5-1.3B}\\ 
\midrule
                                      & 0.25\%                        &$\mathrm{W}_k$ &$\mathrm{W}_q, \mathrm{W}_k$ &$\mathrm{W}_k$ &$\mathrm{W}_k$ \\
                                      & 0.50\%                        & $\mathrm{W}_q, \mathrm{W}_k$&$\mathrm{W}_o,MLP_{\text{up}}$ &$\mathrm{W}_q, \mathrm{W}_k$  &$\mathrm{W}_q, \mathrm{W}_k$     \\
                                      & 1\%                           & $\mathrm{W}_q, \mathrm{W}_k,\mathrm{W}_v,\mathrm{W}_o$ &$\mathrm{W}_q, \mathrm{W}_k,\mathrm{W}_v,\mathrm{W}_o$ & $\mathrm{W}_q, \mathrm{W}_k,\mathrm{W}_v, \mathrm{W}_d$ &$\mathrm{W}_q, \mathrm{W}_k,, \mathrm{W}_v$\\
                                      & 3\%                           & 0 & 0 & 0 & 0  \\
                                      & 7\%                           & 0-1& 0-1 & 0-1 & 0-1 \\
                                      & 15\%                          & 0-4& 0-4 & 0-3 & 0-3  \\
                                      & 30\%                          & 0-9 & 0-9 & 0-9 & 0-6 \\
                                      & 50\%                          & 0-15 & 0-15 & 0-15,$\mathrm{W}_{em}$ & 0-11,$\mathrm{W}_{em}$ \\ 
\multirow{-9}{*}{\textbf{Proportion}} &100\% & Fully-secured & Fully-secured & Fully-secured & Fully-secured \\ \midrule
                                      & 20M                           &$\mathrm{W}_k$&$\mathrm{W}_q, \mathrm{W}_k$ & $\mathrm{W}_q, \mathrm{W}_k, \mathrm{W}_v$ & $\mathrm{W}_q, \mathrm{W}_k,\mathrm{W}_v, \mathrm{W}_d$  \\
                                      & 50M                           & $\mathrm{W}_q, \mathrm{W}_k, \mathrm{W}_v$  & $\mathrm{W}_q, \mathrm{W}_k, \mathrm{W}_v, \mathrm{W}_o$ & $MLP$ & 0\\
                                      & 100M                          & $\mathrm{W}_q, \mathrm{W}_k, \mathrm{W}_v, MLP $  &$\mathrm{W}_q, \mathrm{W}_k, \mathrm{W}_v, \mathrm{W}_o, MLP $ & 0, $\mathrm{W}_q, \mathrm{W}_k, \mathrm{W}_v$ &0-1                       \\
                                      & 160M                          & $\mathrm{W}_q, \mathrm{W}_k, \mathrm{W}_v, \mathrm{W}_o, MLP$   &$\mathrm{W}_q, \mathrm{W}_k, \mathrm{W}_v, \mathrm{W}_o , MLP$    &0-1&0-2             \\
                                      & 200M                          & 0  & 0 & 0-1, $\mathrm{W}_q, \mathrm{W}_k,\mathrm{W}_v, \mathrm{W}_d, MLP_{\text{f1}}$  & 0-3                                \\
                                      & 300M                          & 0, $\mathrm{W}_q, \mathrm{W}_v,\mathrm{W}_o, MLP_{\text{up}}$&0, $\mathrm{W}_q, \mathrm{W}_v,\mathrm{W}_o, MLP_{\text{up}}$  & 0-3 & 0-5                    \\
\multirow{-7}{*}{\textbf{Quantity}}   & 600M                          & 0-2  & 0-2 & 0-7 & 0-11                              \\
\bottomrule
\end{tabular}}
\caption{ Secured Sizes Setting. ``*'' indicates an entire decoder layer.}
\label{appd tab:Priv set}
\end{table*}

\begin{figure}[ht]
  \centering
  \includegraphics[width=0.9\linewidth]{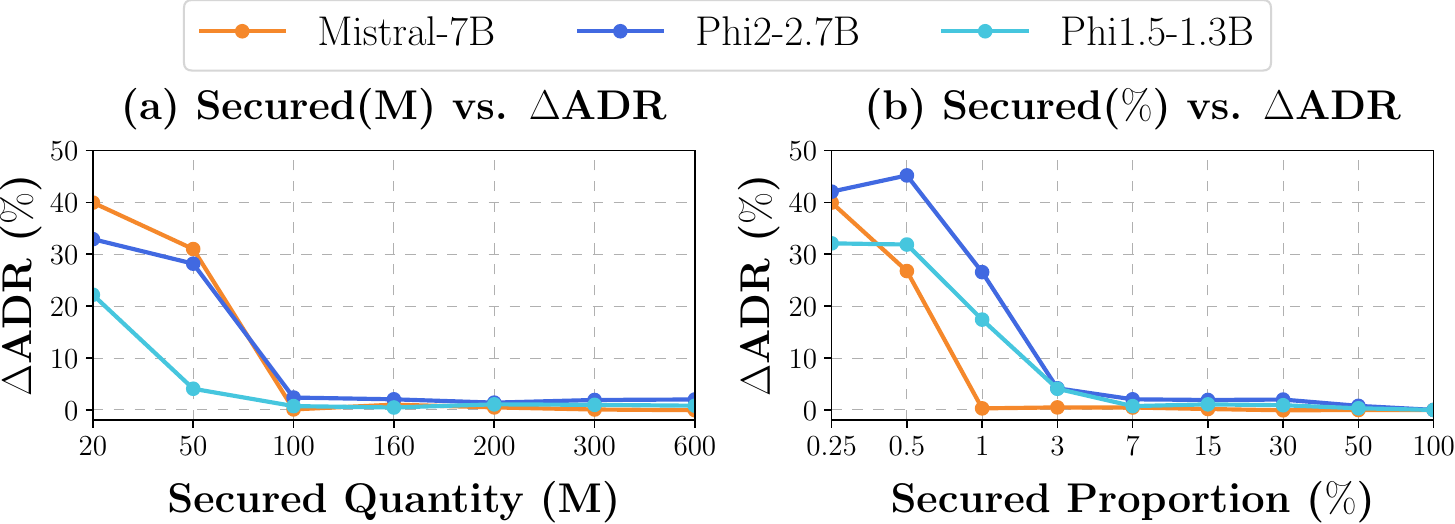}
  \caption{$\Delta$ ADR for different secure parameter quantities and proportions.}
  \label{fig:appendix:appendix-size-othermodels}
\end{figure}

\begin{figure}[ht]
  \centering
  \subfloat[Secured Ratio vs. $\Delta$ADR]{%
    \includegraphics[width=0.37\linewidth]{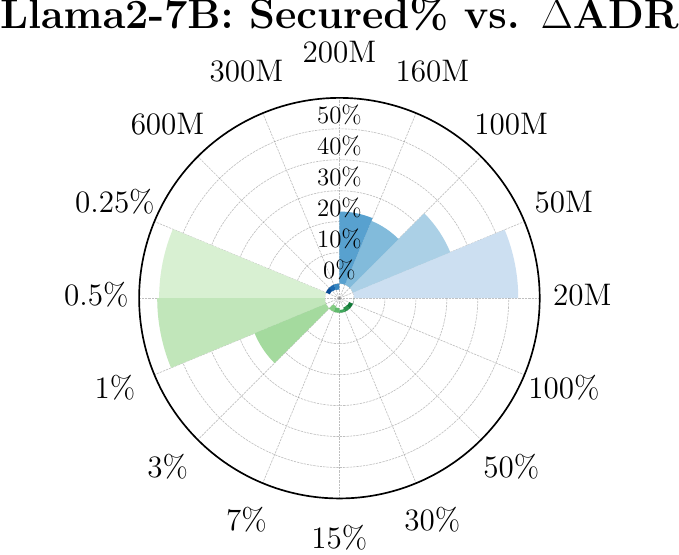}%
  } \hspace{0.5cm}
  \subfloat[Secured Ratio vs. $\Delta$DR]{%
    \includegraphics[width=0.37\linewidth]{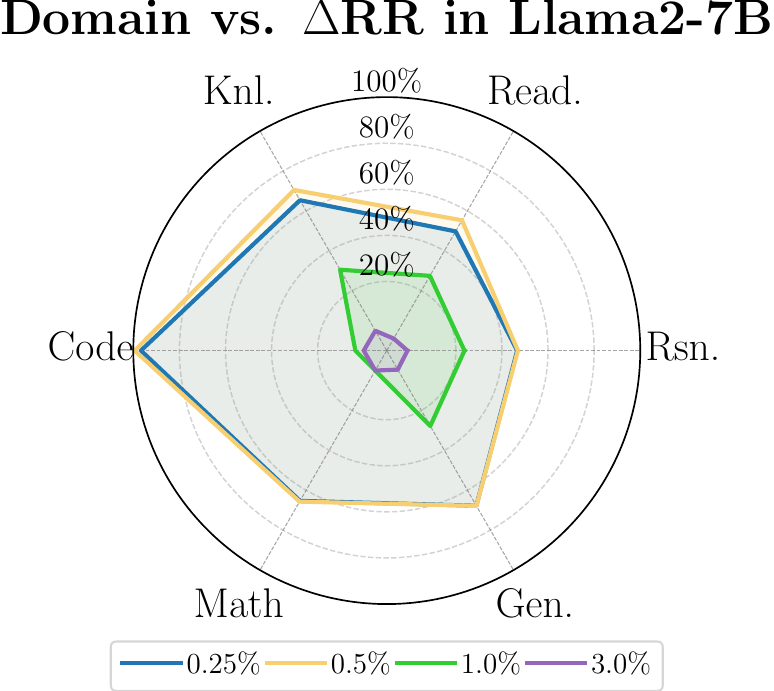}%
  }
  \caption{$\Delta$ADR and $\Delta$DR changes in Llama2-7B with varying secured parameter ratios.}
  \label{fig:appendix:appendix-specific-domain}
\end{figure}

We further computed $\Delta\text{ADR}$ by close-sourcing varying quantities and proportions of parameters under FT-all attacks on three additional models. As shown in Figure~\ref{fig:appendix:appendix-size-othermodels} and Figure~\ref{fig:appendix:appendix-specific-domain}(a), we observed the same pattern as with Llama2-7B, where security emerges once a sufficient number of parameters are secured. For example, on Mistral-7B, security occurs after secure-sourcing 100 million parameters, which is less than a single decoder layer. Secure-sourcing fewer parameters leads to a notable drop in security, with $\Delta\text{ADR}$ rising to around 40\%. Beyond this threshold, security stabilizes near 0\% $\Delta\text{ADR}$. This pattern holds across all models, highlighting a critical threshold for effective secure-source. Furthermore, different architectures require varying secure-sourcing quantities to achieve security, even with similar model sizes. For instance, Mistral-7B reaches security by secure-sourcing 100 million parameters, Llama2-7B requires 200 million, and Phi-1.5 needs a higher rate of 7\%, compared to 3\% for Llama2-7B.

{We explore how secured parameter ratio impacts the model security in Llama2-7B, as shown in Figure~\ref{fig:appendix:appendix-specific-domain}(b). For instance, technical skills such as Math show earlier transitions, with security emerging at 1\% parameters secured, whereas domains such as Commonsense Reasoning require hiding 3\%. In summary, secure-sourcing a small portion of parameters can provide sufficient security against model distillation, meanwhile, technical capabilities tend to be more challenging to distill than other domains.}

\subsection{Effectiveness of distillation difficulty}
\label{append::RS-ADR-coefficient}

\begin{figure*}[t]
  \centering
  \includegraphics[width=\linewidth]{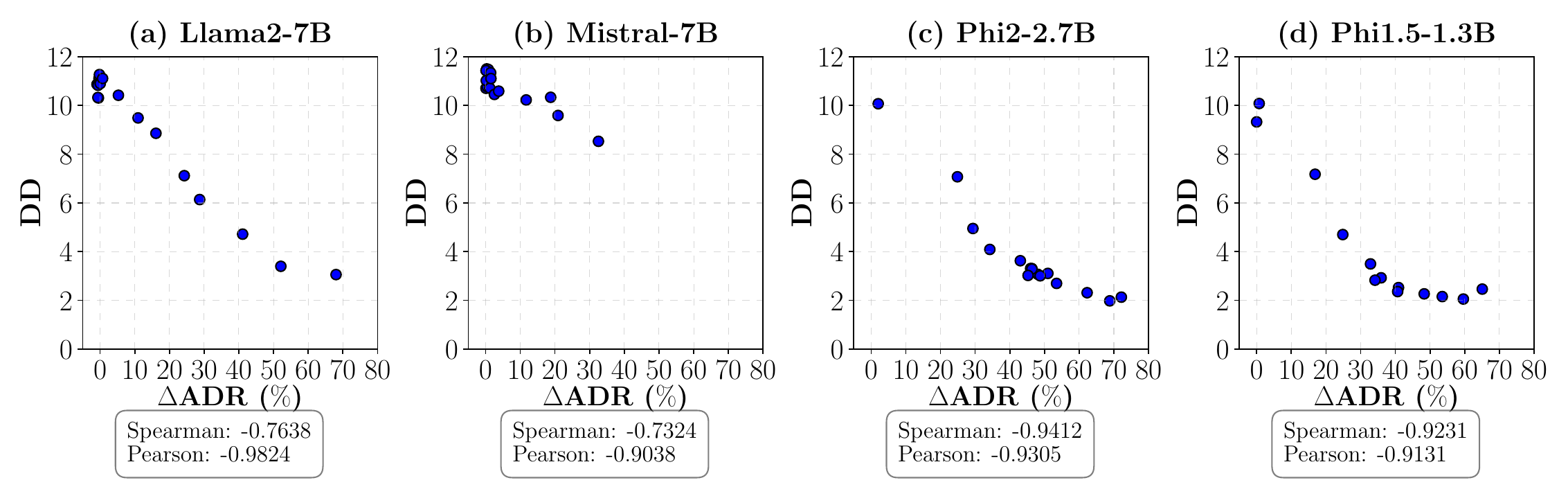}
  \caption{ Correlation Analysis of $\Delta$ADR and Distillation Difficulty Across Different Models.}
  \label{fig:appendix:correlation_scatter}
\end{figure*}

The complete Pearson and Spearman results are presented in Table~\ref{tab:correlation}, revealing a negative correlation between RS and the average distillation ratio. For example, in Llama2-7B, both Pearson and Spearman coefficients fall below -0.77. Similar trends are seen in models with varying architectures and sizes, confirming that RD is a reliable predictor of distilled model performance and demonstrating the effectiveness of SOLID. Additionally, Figure~\ref{fig:appendix:correlation_scatter} shows scatter plots depicting the relationship between $\Delta$ADR and Distillation Difficulty($\uparrow$)s across four models, along with the corresponding Pearson and Spearman correlation coefficients. The Distillation Difficulty($\uparrow$)s were obtained from Section~\ref{subsection:sensitivity}. As illustrated in Figure~\ref{fig:appendix:correlation_scatter}, we observe a clear trend: an increase in $\Delta$ADR corresponds to a decrease in model scores across all models analyzed. This inverse relationship is consistently supported by strong negative values for both Pearson and Spearman correlation coefficients, with the most significant negative correlation seen in Phi2-2.7B, indicating a substantial drop in model scores as $\Delta$ADR increases.

\begin{table*}[t]
    
    \centering

    \begin{tabular}{@{}lcccccc@{}}
    \toprule
    \textbf{\small Model} & \textbf{\small Rsn.} & \textbf{\small Read.} & \textbf{\small Knl.} & \textbf{\small Code \& Math}& \textbf{\small Gen.} & \textbf{\small Avg.}\\ 
    \midrule
    \small Llama2-7B  & \small -0.83 $|$ -0.97  & \small -0.77 $|$ -0.96 & \small -0.83 $|$ -0.95 & \small -0.85 $|$ -0.90  & \small -0.82 $|$ -0.93 & \small -0.80 $|$ -0.98 \\
    \small Mistral-7B & \small  -0.83 $|$ -0.89 & \small -0.82  $|$ -0.91 & \small -0.82 $|$ -0.94 & \small -0.78 $|$ -0.95  & \small -0.76 $|$ -0.87 & \small -0.87 $|$ -0.92\\
    \small Phi-2  & \small -0.93 $|$ -0.96 & \small -0.84 $|$ -0.96 & \small -0.84 $|$ -0.87 & \small -0.84 $|$ -0.80  & \small -0.84 $|$ -0.84 & \small -0.87 $|$ -0.95 \\
    \small Phi-1.5  & \small -0.86 $|$ -0.97 & \small -0.78 $|$ -0.94 & \small -0.83 $|$ -0.94 & \small -0.90 $|$ -0.80  & \small -0.84 $|$ -0.89 & \small -0.80 $|$ -0.94\\
    \bottomrule
    \end{tabular}
    \caption{ Correlation coefficients (Spearman | Pearson) between distillation ratio and distillation difficult.}
    \label{tab:correlation}
\end{table*}

\subsection{Adversarial Attack}
\label{append::Adversarial Attack}
In this section, we provide a detailed comparison of SOLID and SAP-DP in their effectiveness against three types of black-box adversarial attacks on the Llama2-7B model. The attacks considered include Membership Inference Attacks (MIA), Attribute Inference Attacks (AIA), and Prompt Injection Attacks (PIA).

\textbf{Membership Inference Attack (MIA)}: This attack aims to determine whether a specific data point was included in the training dataset of the model. Attackers utilize model outputs to infer membership status, potentially exposing sensitive information about the training data~\citep{MIA,MIA-appendix}. We conducted our experiment following SPV\_MIA~\footnote{https://github.com/wjfu99/MIA-LLMs}, which provides a robust framework for assessing model vulnerabilities. We focus on the AUC scores for SPV-MIA against semi-open models across Ag News datasets~\citep{AGnews}.

\textbf{Attribute Inference Attack (AIA)}: In this scenario, the adversary attempts to infer specific attributes of training data based on the model's outputs. This can lead to privacy breaches, particularly when sensitive attributes are involved~\citep{AIA,AIA-appendix}. We conducted our experiments following the methodology outlined in~\cite{AIA}~\footnote{https://github.com/eth-sri/llmprivacy} and evaluated the top-3 accuracy on the PersonalReddit (PR) Dataset.

\textbf{Prompt Injection Attack (PIA)}: This attack manipulates input prompts to coerce the model into producing desired outputs that may compromise the integrity or security of the system~\citep{PIA-appendix,PIA-appendix2}. In our experiment, we follow AutoDAN~\footnote{https://github.com/SheltonLiu-N/AutoDAN}, which can automatically generate stealthy jailbreak prompts by the carefully designed hierarchical genetic algorithm.We evaluate the effectiveness of these prompts using the \textit{keyword-based attack success rate} (ASR), which measures the presence of predefined keywords in responses generated LLMs. For gold standard, LED~\footnote{https://github.com/ledllm/ledllm}, significantly enhances the security of LLMs against prompt injection attacks (PIA), reducing the ASR to 0.

\textbf{Limited Defense against Adversarial Attack.} We compare SOLID and SAP-DP in defending against three black-box adversarial attacks on Llama2-7B. Specifically, we apply the membership inference~\citep{MIA} (MIA), attribute inference~\citep{AIA} (AIA), and prompt injection~\citep{liu2023autodan} (PIA) attacks to the semi-open models produced by SAP-DP and SOLID. As shown in Table~\ref{tab:ASR}, we observe that SAP-DP outperforms SOLID across all three attacks, but still performs worse than the gold standard. This is because SOLID does not introduce additional output perturbation and thus provide limited defense against black-box adversarial attacks. Details can be found in Appendix~\ref{append::Adversarial Attack}.
\begin{table}[h]
    \centering
    \begin{tabular}{@{}lccc@{}}
    \toprule
    \textbf{\footnotesize Approach} & \textbf{\footnotesize MIA$\downarrow$} & \textbf{\footnotesize AIA$\downarrow$} & \textbf{\footnotesize PIA$\downarrow$} \\
    \midrule
    \footnotesize Gold Std.  & \footnotesize 58.0  & \footnotesize 43.9 & \footnotesize 0.00 \\
    \footnotesize SCARA  & \footnotesize 72.3  & \footnotesize 85.0 & \footnotesize 26.5 \\
    \footnotesize SAP-DP  & \footnotesize 72.2  & \footnotesize 83.9 & \footnotesize 24.9 \\
    \bottomrule
    \end{tabular}
    \caption{Performance of SCARA defending adversarial attacks. $\downarrow$ indicates the smaller the better.}
    \label{tab:ASR}
\end{table}

\section{Detailed Results}
\label{append:result}

\subsection{Comparison in two semi-open Llama2-70B}
\label{append:result:semi-open Llama2-70B}

In this experiment, we examine two semi-open Llama2-70B models, where either the first two decoder layers are secure-source (referred to as Bottom2-Secured) or the last two decoder layers are secure-source (referred to as Top2-Secured). The objective is to compare their performance in terms of customization and their security under the distillation attack. The results are summarized in Table~\ref{tab:appendix:cust-2semiopen} and Table~\ref{tab:appendix:distillation-2semiopen}.

\begin{table*}[h]
    \centering
    \begin{tabular}{ccccccc}
    \toprule
         & \textbf{Math} & \textbf{Code} & \textbf{Medical} & \textbf{Law} & \textbf{Finance} & \textbf{Alignment}  \\ 
         \midrule
        \textbf{Fully Secured} & 53.15  & 24.90  & 53.68  & 79.63  & 37.54  & 7.19   \\ 
        \textbf{Bottom2-Secured}  & 62.40  & 43.99  & 62.73  & 93.85  & 87.51  & 7.46   \\ 
        \textbf{Top2-Secured} & 62.53  & 42.36  & 62.72  & 93.91  & 87.90  & 7.46  \\ 
    \bottomrule
    \end{tabular}
    \caption{Customization Performance of Llama2-70B under Different Secured Layers}
    \label{tab:appendix:cust-2semiopen}
\end{table*}

\begin{table*}[t]
    \centering
    \begin{tabular}{ccccc}
    \toprule
        \textbf{} & \textbf{Benchmarks}  & \textbf{Fully Secured} & \textbf{Semi-Open-1} & \textbf{Semi-Open-2} \\ 
        \midrule
        \multirow{5}{*}{\textbf {Rsn.}} & PIQA & 50.82  & 50.49  & 79.05   \\ 
        \textbf{} & winogrande & 51.07  & 51.22  & 72.93   \\ 
        \textbf{} & arc\_easy & 25.17  & 25.63  & 76.30   \\ 
        \textbf{} & arc\_challenge & 23.55  & 20.48  & 50.17   \\ 
        \textbf{} & Hellaswag & 26.65  & 25.77  & 79.49   \\ 
        \midrule
        \multirow{5}{*}{\textbf {Read.} } & lambada & 0.00  & 0.01  & 57.25   \\ 
        \textbf{} & BoolQ & 43.30  & 37.92  & 84.95   \\ 
        \textbf{} & SQuADv2\_EM & 0.00  & 0.00  & 1.54   \\ 
        \textbf{} & SQuADv2\_f1 & 0.23  & 1.01  & 35.59   \\ 
        \textbf{} & OBQA & 25.60  & 24.40  & 44.00   \\ 
        \midrule
        \multirow{2}{*}{\textbf {Knl.} } & NQ & 0.00  & 0.00  & 15.18   \\ 
        \textbf{} & TriviaQA & 0.00  & 0.00  & 52.67   \\ 
        \midrule
        \multirow{2}{*}{\textbf{Code}} & mbpp & 0.00  & 0.00  & 16.00   \\ 
        \textbf{} & HumanEval & 0.00  & 0.00  & 13.41   \\ 
        \midrule
        \textbf{Math} & GSM8K & 0.03  & 0.01  & 27.75   \\ 
        \midrule
        \multirow{2}{*}{\textbf{Gen.}} & MMLU & 23.01  & 23.22  & 63.61   \\ 
        \textbf{} & BBH & 0.00  & 0.00  & 49.45   \\ 
        \midrule
        \multicolumn{2}{c}{\textbf{Average Distillation Ratio($\downarrow$)}}  & 22.55  & 21.73  & 74.94  \\ 
        \bottomrule
    \end{tabular}
    \caption{Customization Performance of Llama2-70B under Different Secured Layers}
    \label{tab:appendix:distillation-2semiopen}
\end{table*}

\subsection{Evaluation Results under FT-all attack}
\label{append:result:Table1}
In this section, we provide a comprehensive analysis of the evaluation results, comparing SOLID with two baseline methods: SAP-DP and a fully-secured approach. This comparison is conducted across 16 benchmarks under the FT-all attack scenario.
The detailed results for Llama2-70B are presented in Table~\ref{tab:appendix:70B FT-all}, while the results for Llama2-7B and Mistral-7B are shown in Table~\ref{tab:appendix:7B FT-all}. Additionally, the outcomes for Phi-2 and Phi-1.5 are provided in Tables~\ref{tab:appendix:phi FT-all}.

\begin{table*}[t]
    \centering
    \begin{tabular}{cccccc}
    \toprule
        \textbf{} & \textbf{} & \textbf{Pre-train} & \textbf{SOLID} & \textbf{SAP-DP} & \textbf{Fully-secured } \\  
        \midrule
        \multirow{5}{*}{\textbf {Rsn.}} & PIQA & 80.69  & 50.49  & 48.26  & 50.82   \\ 
        \textbf{} & Winogrande & 74.74  & 51.22  & 50.59  & 51.07   \\ 
        \textbf{} & ARC-easy & 80.35  & 25.63  & 26.35  & 25.17   \\ 
        \textbf{} & ARC-challenge & 53.24  & 20.48  & 20.31  & 23.55   \\ 
        \textbf{} & Hellaswag & 82.15  & 25.77  & 25.76  & 26.65   \\  
        \midrule
        \multirow{5}{*}{\textbf{Read.}} & LAMBADA & 75.07  & 0.01  & 0.00  & 0.00   \\ 
        \textbf{} & BoolQ & 86.70  & 37.92  & 37.83  & 43.30   \\ 
        \textbf{} & SQuADv2\_EM & 51.23  & 0.00  & 0.00  & 0.00   \\ 
        \textbf{} & SQuADv2\_f1 & 67.43  & 1.01  & 1.13  & 0.23   \\ 
        \textbf{} & OBQA & 44.80  & 24.40  & 24.40  & 25.60   \\  
        \midrule
        \multirow{2}{*}{\textbf{Knl.}} & NaturalQuestions & 32.38  & 0.00  & 0.00  & 0.00   \\ 
        \textbf{} & TriviaQA & 73.47  & 0.00  & 0.02  & 0.00   \\  
        \midrule
        \multirow{2}{*}{\textbf{Code}} & MBPP & 24.80  & 0.00  & 0.00  & 0.00   \\ 
        \textbf{} & HumanEval & 25.00  & 0.00  & 0.00  & 0.00   \\  
        \midrule
        \textbf{Math} & GSM8K & 53.15  & 0.01  & 0.00  & 0.03   \\  
        \midrule
        \multirow{2}{*}{\textbf{Gen.}} & MMLU & 63.09  & 23.22  & 24.19  & 23.01   \\ 
        \textbf{} & BBH & 61.40  & 0.00  & 0.00  & 0.00   \\
        \midrule
        \multicolumn{2}{c}{\textbf{Average Distillation Ratio($\downarrow$)}}& - & 21.73  & 21.64  & 22.55  \\
        \bottomrule
    \end{tabular}
    \caption{Evaluation results of Llama2-70B under FT-all attack}
    \label{tab:appendix:70B FT-all}
\end{table*}

\begin{table*}[t]
    \centering
    \small
    \setlength{\tabcolsep}{3pt}
    \begin{tabular}{cc|ccc|ccc}
    \toprule
            \multicolumn{2}{c}{} & \multicolumn{3}{c}{\textbf{Llama2-7B}} & \multicolumn{3}{c}{\textbf{Mistral-7B}} \\ 
            ~ & &\textbf{ SOLID} & \textbf{SAP-DP} & \textbf{Fully-secured} & \textbf{SOLID} & \textbf{SAP-DP} & \textbf{Fully-secured}  \\ 
            \midrule
            \multirow{5}{*}{\textbf {Rsn.}} & PIQA & 49.56  & 49.56  & 49.47  & 51.63  & 50.22  & 49.35   \\ 
            \textbf{} & Winogrande & 50.99  & 49.66  & 50.83  & 49.78  & 51.07  & 50.59   \\ 
            \textbf{} & ARC-easy & 27.04  & 26.43  & 25.98  & 26.12  & 28.03  & 25.83   \\ 
            \textbf{} & ARC-challenge & 21.07  & 20.56  & 22.47  & 19.94  & 21.42  & 22.35   \\ 
            \textbf{} & Hellaswag & 25.56  & 25.69  & 26.39  & 26.10  & 25.97  & 25.39   \\ 
        \midrule
            \multirow{5}{*}{\textbf{Read.}} & LAMBADA & 0.01  & 0.00  & 0.01  & 0.12  & 0.00  & 0.01   \\ 
            \textbf{} & BoolQ & 44.30  & 41.70  & 48.34  & 39.05  & 37.83  & 45.80   \\ 
            \textbf{} & SQuADv2\_EM & 0.00  & 0.00  & 0.00  & 0.00  & 0.00  & 0.00   \\ 
            \textbf{} & SQuADv2\_f1 & 0.49  & 0.63  & 0.59  & 1.21  & 0.26  & 0.66   \\ 
            \textbf{} & OBQA & 25.13  & 23.00  & 25.93  & 25.60  & 25.20  & 25.00   \\ 
        \midrule
            \multirow{2}{*}{\textbf{Knl.}} & NaturalQuestions & 0.01  & 0.01  & 0.04  & 0.00  & 0.00  & 0.02   \\ 
            \textbf{} & TriviaQA & 0.00  & 0.00  & 0.02  & 0.00  & 0.00  & 0.01   \\ 
        \midrule
            \multirow{2}{*}{\textbf{Code}} & MBPP & 0.00  & 0.00  & 0.00  & 0.00  & 0.00  & 0.00   \\ 
            \textbf{} & HumanEval & 0.00  & 0.00  & 0.00  & 0.00  & 0.00  & 0.00   \\ 
        \midrule
            \textbf{Math} & GSM8K & 0.00  & 0.00  & 0.00  & 0.00  & 0.00  & 0.00   \\ 
        \midrule
            \multirow{2}{*}{\textbf{Gen.}} & MMLU & 24.26  & 22.92  & 24.45  & 25.24  & 23.05  & 23.26   \\ 
            \textbf{} & BBH & 0.00  & 0.00  & 0.00  & 0.00  & 0.00  & 0.00   \\ 
        \midrule
            \multicolumn{2}{c}{\textbf{Average Distillation Ratio($\downarrow$)}} & 25.03  & 24.16  & 25.62  & 22.41  & 22.28  & 22.68  \\
        \bottomrule
    \end{tabular}
    \caption{Evaluation results of 7B models under FT-all attack}
    \label{tab:appendix:7B FT-all}
\end{table*}

\begin{table*}[t]
    \centering
    \small
    \setlength{\tabcolsep}{3pt}
    \begin{tabular}{cc|ccc|ccc}
    \toprule
        \multicolumn{2}{c}{}& \multicolumn{3}{c}{\textbf{Phi-2}} & \multicolumn{3}{c}{\textbf{Phi-1.5}} \\ 
        ~ & &\textbf{ SOLID} & \textbf{SAP-DP} & \textbf{Fully-secured} & \textbf{SOLID} & \textbf{SAP-DP} & \textbf{Fully-secured}  \\ 
        \midrule
        \multirow{5}{*}{\textbf {Rsn.}} & PIQA & 54.17  & 52.01  & 52.07  & 53.43  & 52.61  & 50.44   \\ 
        ~ & Winogrande & 51.56  & 48.93  & 48.91  & 51.09  & 49.25  & 49.12   \\ 
        ~ & ARC\_easy & 34.57  & 28.20  & 27.03  & 30.81  & 28.79  & 27.50   \\ 
        ~ & ARC\_challenge & 19.45  & 19.37  & 18.66  & 20.56  & 19.80  & 21.22   \\ 
        ~ & Hellaswag & 27.61  & 25.32  & 25.26  & 26.27  & 25.66  & 25.05   \\ 
        \multirow{5}{*}{\textbf {Read.} } & LAMBADA & 0.75  & 0.02  & 0.00  & 0.59  & 0.00  & 0.00   \\ 
        \midrule
        ~ & BoolQ & 45.29  & 40.21  & 44.60  & 46.98  & 41.80  & 46.28   \\ 
        ~ & SQuADv2\_EM & 0.02  & 0.00  & 0.00  & 0.00  & 0.00  & 0.00   \\ 
        ~ & SQuADv2\_f1 & 2.61  & 0.28  & 0.64  & 0.78  & 0.65  & 1.60   \\ 
        ~ & OBQA & 24.80  & 26.60  & 25.80  & 26.60  & 28.60  & 26.87   \\ 
        \midrule
        \multirow{2}{*}{\textbf {Knl.}}& NaturalQuestions & 0.00  & 0.00  & 0.02  & 0.04  & 0.00  & 0.00   \\ 
        ~ & TriviaQA & 0.01  & 0.00  & 0.01  & 0.01  & 0.00  & 0.00   \\ 
        \midrule
        \multirow{2}{*}{\textbf{Code}} &MBPP & 0.00  & 0.00  & 0.00  & 0.00  & 0.00  & 0.00   \\ 
        ~ & HumanEval & 0.00  & 0.00  & 0.00  & 0.00  & 0.00  & 0.00   \\ 
        \midrule
        \textbf{Math} & GSM8K & 0.00  & 0.00  & 0.00  & 0.00  & 0.00  & 0.00   \\ 
        \midrule
        \multirow{2}{*}{\textbf{Gen.}} & MMLU & 24.16  & 22.87  & 22.95  & 24.07  & 22.95  & 22.95   \\ 
        ~ & BBH & 0.01  & 0.00  & 0.00  & 0.00  & 0.00  & 0.00   \\ 
        \bottomrule

    \end{tabular}
    \caption{Evaluation results of small models under FT-all attack}
    \label{tab:appendix:phi FT-all}
\end{table*}

\subsection{Customization Performance of Models}
\label{append:result:custom performance}
In this section, we present detailed evaluation results of the model customization performance across six downstream tasks used in our experiments. 
The detailed results for Llama2-70B are presented in Table~\ref{fig:custom llama2-70B}, while the results for Llama2-7B and Mistral-7B are shown in Table~\ref{fig:custom llama2-7B} and Table~\ref{fig:custom Mistral-7B}. Additionally, the outcomes for Phi-2 and Phi-1.5 are provided in Tables~\ref{fig:custom Phi2} and Table~\ref{fig:custom phi1.5}.

\begin{table*}[t]
    \centering
    \begin{tabular}{ccccccc}
    \toprule
        ~ & \textbf{Math} & \textbf{Code} & \textbf{Medical} & \textbf{Law} & \textbf{Finance} & \textbf{Alignment }  \\ \midrule
        \textbf{Fully-Secure} & 53.15  & 24.90  & 53.68  & 79.63  & 55.63  & 7.19   \\ 
        \textbf{SAP-DP} & 61.10  & 36.87  & 54.55  & 83.40  & 65.78  & 7.41   \\ 
        \textbf{SOLID} & 62.40  & 43.99  & 62.73  & 93.85  & 87.51  & 7.46   \\ 
        \textbf{Fully-Open} & 64.06  & 44.58  & 63.40  & 94.17  & 88.22  & 7.42  \\
        \bottomrule
    \end{tabular}
    \caption{Detailed results of Llama2-70B secure by SOLID on six downstream tasks.}
    \label{fig:custom llama2-70B}
\end{table*}

\begin{table*}[t]
    \centering
    \begin{tabular}{ccccccc}
    \toprule
        ~ & \textbf{Math} & \textbf{Code} & \textbf{Medical} & \textbf{Law} & \textbf{Finance} & \textbf{Alignment }  \\ \midrule
        \textbf{Fully-Secure} & 20.24  & 13.75 & 36.91  & 51.80  & 38.71 & 6.51  \\ 
        \textbf{SAP-DP} & 20.24  & 13.75 & 36.91  & 51.80  & 38.71 & 6.52  \\ 
        \textbf{SOLID} & 28.96 & 21.37 & 46.52  & 90.84  & 81.95 & 6.63  \\ 
        \textbf{Fully-Open} & 29.34  & 21.265 & 47.60  & 90.49  & 84.09 & 6.63 \\ 
        \bottomrule
    \end{tabular}
    \caption{Detailed results of Llama2-7B secure by SOLID on six downstream tasks.}
    \label{fig:custom llama2-7B}
\end{table*}

\begin{table*}[t]
    \centering
    \begin{tabular}{ccccccc}
    \toprule
        ~ & \textbf{Math} & \textbf{Code} & \textbf{Medical} & \textbf{Law} & \textbf{Finance} & \textbf{Alignment }  \\ \midrule
        \textbf{Fully-Secure} & 38.21  & 33.83  & 61.50  & 50.47  & 37.39  & 3.20   \\ 
        \textbf{SAP-DP} & 41.47  & 34.44  & 63.08  & 50.37  & 38.10  & 2.47   \\ 
        \textbf{SOLID} & 46.10  & 43.16  & 66.78  & 84.94  & 86.19  & 3.87   \\ 
        \textbf{Fully-Open} & 45.26  & 46.08  & 66.47  & 88.13  & 84.91  & 3.78  \\
        \bottomrule
    \end{tabular}
    \caption{Detailed results of Mistral-7B secure by SOLID on six downstream tasks.}
    \label{fig:custom Mistral-7B}
\end{table*}

\begin{table*}[t]
    \centering
    \begin{tabular}{ccccccc}
    \toprule
        ~ & \textbf{Math} & \textbf{Code} & \textbf{Medical} & \textbf{Law} & \textbf{Finance} & \textbf{Alignment }  \\ \midrule
        \textbf{Fully-Secure} & 57.77  & 47.59  & 43.13  & 56.46  & 54.07  & 5.22   \\ 
        \textbf{SAP-DP} & 58.52  & 46.65  & 43.40  & 56.81  & 54.37  & 5.11   \\ 
        \textbf{SOLID} & 59.59  & 47.79  & 45.85  & 57.11  & 56.26  & 5.26   \\ 
        \textbf{Fully-Open} & 59.60  & 48.40  & 45.93  & 57.19  & 56.68  & 5.27  \\
        \bottomrule
    \end{tabular}
    \caption{Detailed results of Phi-2 secure by SOLID on six downstream tasks.}
    \label{fig:custom Phi2}
\end{table*}

\begin{table*}[t]
    \centering
    \begin{tabular}{ccccccc}
    \toprule
        \textbf{} & \textbf{Math} & \textbf{Code} & \textbf{Medical} & \textbf{Law} & \textbf{Finance} & \textbf{Alignment }  \\ \midrule
        \textbf{Fully-Secure} & 30.33  & 35.09  & 30.78  & 52.18  & 34.60  & 3.24 \\ 
        \textbf{SAP-DP} & 30.25  & 35.45  & 32.66  & 51.99  & 34.27  & 3.68 \\ 
        \textbf{SOLID} & 33.66  & 37.10  & 33.14  & 52.26  & 39.60  & 3.87 \\ 
        \textbf{Fully-Open} & 34.49  & 37.45  & 33.23  & 52.34  & 39.90  & 3.68\\ \bottomrule
    \end{tabular}
    \caption{Detailed results of Phi-1.5 secure by SOLID on six downstream tasks.}
    \label{fig:custom phi1.5}
\end{table*}

\subsection{Comparison in deployment baselines on llama2-70B}
\label{appen::SOLID-Baselines-DATA}
We compare the distillation security of SOLID with SAP-DP and Fully-secured as baselines under FT-secure and SEM attack strategies. The evaluation results on sixteen benchmarks are shown in Table~\ref{tab:appendix:FT-secure and SEM attack}.

\begin{table*}[t]
    \centering
    \begin{tabular}{cc|cc|cc}
    \toprule
        \multicolumn{2}{c}{} & \multicolumn{2}{c}{\textbf{FT-secure}}   & \multicolumn{2}{c}{\textbf{\textbf{SEM}}} \\  
        \textbf{} & ~ & SOLID & SAP-DP & SOLID & SAP-DP  \\  
        \midrule
        \multirow{5}{*}{\textbf {Rsn.}} & PIQA & 49.78  & 49.40  & 48.62  & 49.00   \\ 
        \textbf{} & Winogrande & 51.30  & 49.01  & 50.99  & 51.13   \\ 
        \textbf{} & ARC-easy & 26.43  & 25.59  & 25.33  & 24.55   \\ 
        \textbf{} & ARC-challenge & 21.41  & 21.42  & 22.01  & 20.93   \\ 
        \textbf{} & Hellaswag & 26.07  & 26.10  & 25.90  & 25.22   \\  
        \midrule
        \multirow{5}{*}{\textbf{Read.}} & LAMBADA & 0.00  & 0.00  & 0.00  & 0.00   \\ 
        \textbf{} & BoolQ & 45.09  & 37.83  & 44.95  & 39.80   \\ 
        \textbf{} & SQuADv2\_EM & 0.00  & 0.00  & 0.00  & 0.00   \\ 
        \textbf{} & SQuADv2\_f1 & 0.98  & 1.01  & 0.59  & 1.00   \\ 
        \textbf{} & OBQA & 24.40  & 23.80  & 25.03  & 22.96   \\  
        \midrule
        \multirow{2}{*}{\textbf{Knl.}} & NaturalQuestions & 0.00  & 0.00  & 0.00  & 0.00   \\ 
        \textbf{} & TriviaQA & 0.00  & 0.00  & 0.00  & 0.00   \\  
        \midrule
        \multirow{2}{*}{\textbf{Code}} & MBPP & 0.00  & 0.00  & 0.00  & 0.00   \\ 
        \textbf{} & HumanEval & 0.00  & 0.00  & 0.00  & 0.00   \\ 
        \textbf{Math} & GSM8K & 0.00  & 0.00  & 0.00  & 0.00   \\  
        \midrule
        \multirow{2}{*}{\textbf{Gen.}} & MMLU & 23.18  & 23.66  & 22.98  & 22.83   \\ 
        \textbf{} & BBH & 0.00  & 0.00  & 0.00  & 0.00   \\  
        \midrule
        \multicolumn{2}{c}{\textbf{Average Distillation Ratio($\downarrow$)}} & 22.60  & 21.80  & 22.40  & 22.30  \\ 
        \bottomrule
    \end{tabular}
    \caption{Evaluation results of Llama2-70B under FT-secure and SEM attack}
    \label{tab:appendix:FT-secure and SEM attack}
\end{table*}

\subsection{Comparison in Distillation Attack Strategies}
\label{appen::reovery-strategies-DATA}

In this section, we present detailed evaluation results of the model distillation performance of SOLID under FT-secure and SEM attack strategies across six functionalities used in our experiments. 
The detailed results under the FT-secure distillation strategy are presented in Table~\ref{tab:appendix:result:SOLID under FT-Secure attacks}. The results under SEM attack strategies are shown in Table~\ref{tab:appendix:result:SOLID under SEM attacks}.

\begin{table*}[t]
    \centering

    \begin{tabular}{cccccc}
    \toprule
        & \textbf{} & \textbf{Llama2-7B} & \textbf{Mistral-7B} & \textbf{Phi-2} & \textbf{Phi-1.5 } \\ 
        \midrule
        \multirow{5}{*}{\textbf {Rsn.}} & PIQA & 49.95  & 49.55  & 54.57  & 52.45   \\ 
        \textbf{} & Winogrande & 49.88  & 49.68  & 52.33  & 52.41   \\ 
        \textbf{} & ARC-easy & 27.65  & 25.88  & 33.33  & 31.06   \\ 
        \textbf{} & ARC-challenge & 20.81  & 22.69  & 19.03  & 18.77   \\ 
        \textbf{} & Hellaswag & 26.04  & 25.01  & 27.62  & 26.88   \\ 
        \midrule
        \multirow{5}{*}{\textbf{Read.}} & LAMBADA & 0.00  & 0.00  & 0.77  & 0.71   \\ 
        \textbf{} & BoolQ & 38.13  & 46.01  & 44.34  & 57.49   \\ 
        \textbf{} & SQuADv2\_EM & 0.00  & 0.00  & 0.00  & 0.00   \\ 
        \textbf{} & SQuADv2\_f1 & 0.22  & 0.36  & 3.07  & 2.27   \\ 
        \textbf{} & OBQA & 25.70  & 25.12  & 24.40  & 25.20   \\ 
        \midrule
        \multirow{2}{*}{\textbf{Knl.}} & NaturalQuestions & 0.00  & 0.00  & 0.00  & 0.00   \\ 
        \textbf{} & TriviaQA & 0.00  & 0.00  & 0.01  & 0.00   \\ 
        \multirow{2}{*}{\textbf{Code}} & MBPP & 0.00  & 0.00  & 0.00  & 0.00   \\ 
        \textbf{} & HumanEval & 0.00  & 0.00  & 0.00  & 0.00   \\ 
        \midrule
        \textbf{Math} & GSM8K & 0.00  & 0.00  & 0.00  & 0.00   \\ 
        \midrule
        \multirow{2}{*}{\textbf{Gen.}} & MMLU & 24.23  & 23.56  & 23.03  & 24.10   \\ 
        \textbf{} & BBH & 0.00  & 0.00  & 0.00  & 0.00   \\ 
        \midrule
        \multicolumn{2}{c}{\textbf{Average Distillation Ratio($\downarrow$)}} & 24.80  & 22.50  & 23.56  & 26.97  \\ 
        \bottomrule
    \end{tabular}    
    \caption{Distillation Performance of SOLID under FT-Secure attacks.}
    \label{tab:appendix:result:SOLID under FT-Secure attacks}
\end{table*}

\begin{table*}[t]
    \centering
    \begin{tabular}{cccccc}
        \toprule
        & \textbf{} & \textbf{Llama2-7B} & \textbf{Mistral-7B} & \textbf{Phi-2} & \textbf{Phi-1.5 } \\ 
        \midrule 
        \multirow{5}{*}{\textbf {Rsn.}} & PIQA & 51.52  & 48.53  & 49.46  & 50.82   \\ 
         & Winogrande & 50.28  & 51.02  & 48.70  & 50.59   \\ 
         & ARC-easy & 24.83  & 25.83  & 25.93  & 24.62   \\ 
         & ARC-challenge & 24.99  & 22.35  & 20.65  & 21.08   \\ 
         & Hellaswag & 25.58  & 25.39  & 25.84  & 25.39   \\ 
        \midrule 
        \multirow{5}{*}{\textbf{Read.}} & LAMBADA & 0.00  & 0.01  & 0.00  & 0.01   \\ 
         & BoolQ & 53.30  & 45.80  & 38.41  & 61.07   \\ 
         & SQuADv2\_EM & 0.00  & 0.00  & 0.00  & 0.00   \\ 
         & SQuADv2\_f1 & 0.77  & 0.66  & 0.00  & 1.35   \\ 
         & OBQA & 25.00  & 25.00  & 27.80  & 30.40   \\ 
        \midrule 
        \multirow{2}{*}{\textbf{Knl.}} & NaturalQuestions & 0.00  & 0.02  & 0.00  & 0.00   \\ 
         & TriviaQA & 0.00  & 0.01  & 0.01  & 0.00   \\ 
        \midrule 
        \multirow{2}{*}{\textbf{Code}} & MBPP & 0.00  & 0.00  & 0.00  & 0.00   \\ 
         & HumanEval & 0.00  & 0.00  & 0.00  & 0.00   \\ 
        \midrule 
       \textbf{ Math} & GSM8K & 0.00  & 0.00  & 0.00  & 0.00   \\ 
        \midrule 
        \multirow{2}{*}{\textbf{Gen.}} & MMLU & 25.39  & 23.26  & 22.95  & 23.11   \\ 
         & BBH & 0.00  & 0.00  & 0.00  & 0.00   \\ 
        \midrule 
        \multicolumn{2}{c}{\textbf{Average Distillation Ratio($\downarrow$)}}& 25.00  & 22.00  & 22.10  & 24.70  \\ 
        \bottomrule
    \end{tabular}
    \caption{Distillation Performance of SOLID under SEM attacks.}
    \label{tab:appendix:result:SOLID under SEM attacks}
\end{table*}

\subsection{Comparison in Distillation datasets scales}
\label{appendix:datasets scales}
To investigate the impact of attack dataset scales on the efficiency of SOLID, we conduct model distillation attack on the Llama2-7B model using four different attack datasets of varying sizes: 100k, 200k, 300k, and 500k. The evaluation performance under different attack set scales are in Table~\ref{tab:appendix:result:Attack Set Scales}

\begin{table*}[t]
    \centering
    \begin{tabular}{ccccccc}
    \toprule
        \textbf{} & \textbf{} & \textbf{51K} & \textbf{100K} & \textbf{200K} & \textbf{300K} & \textbf{500K } \\ 
        \midrule
        \multirow{5}{*}{\textbf {Rsn.}} & PIQA & 49.56  & 49.89  & 49.18  & 49.18  & 49.59   \\ 
        \textbf{} & Winogrande & 50.99  & 47.99  & 49.49  & 50.20  & 50.20   \\ 
        \textbf{} & ARC-easy & 27.04  & 27.06  & 27.06  & 27.02  & 27.01   \\ 
        \textbf{} & ARC-challenge & 21.07  & 21.33  & 20.90  & 21.16  & 21.48   \\ 
        \textbf{} & Hellaswag & 25.56  & 26.49  & 26.46  & 26.50  & 26.19   \\ 
        \midrule
        \multirow{5}{*}{\textbf{Read.}} & LAMBADA & 0.01  & 0.01  & 0.00  & 0.00  & 0.01   \\ 
        \textbf{} & BoolQ & 44.30  & 44.41  & 44.10  & 44.07  & 44.96   \\ 
        \textbf{} & SQuADv2\_EM & 0.00  & 0.00  & 0.00  & 0.02  & 0.00   \\ 
        \textbf{} & SQuADv2\_f1 & 1.05  & 0.32  & 0.51  & 0.52  & 0.71   \\ 
        \textbf{} & OBQA & 25.13  & 25.00  & 23.80  & 25.20  & 25.60   \\ 
        \midrule
        \multirow{2}{*}{\textbf{Knl.}} & NaturalQuestions & 0.01  & 0.08  & 0.08  & 0.06  & 0.06   \\ 
        \textbf{} & TriviaQA & 0.00  & 0.02  & 0.01  & 0.03  & 0.01   \\ 
        \midrule
        \multirow{2}{*}{\textbf{Code}} & MBPP & 0.00  & 0.00  & 0.00  & 0.00  & 0.00   \\ 
        \textbf{} & HumanEval & 0.00  & 0.00  & 0.00  & 0.00  & 0.00   \\ 
        \midrule
        \textbf{Math} & GSM8K & 0.00  & 0.00  & 0.00  & 0.00  & 0.00   \\ 
        \midrule
        \multirow{2}{*}{\textbf{Gen.}} & MMLU & 24.26  & 25.34  & 25.43  & 26.14  & 26.41   \\ 
        \textbf{} & BBH & 0.00  & 0.00  & 0.00  & 0.00  & 0.00   \\ 
        \midrule
        \multicolumn{2}{c}{\textbf{Average Distillation Ratio($\downarrow$)}} & 25.07  & 25.03  & 24.89  & 25.26  & 25.48  \\ 
        \bottomrule
    \end{tabular}
    \caption{Evaluation Results of SOLID on Llama2-7B under Various Attack Set Scales.}
    \label{tab:appendix:result:Attack Set Scales}
\end{table*}

\subsection{Transition Layer Results.}
\label{append:result:Transition}

\textbf{Security Performance. }We close same-sized layer sets with different start points, and attack them using FT-all. Specifically, the sets consist of one layer for Llama2-7B (Table~\ref{llama-1.1}, Table~\ref{llama-1.2}), and two layers for Phi-2 (Table~\ref{Phi2-1-part1}, Table~\ref{Phi2-1-part2}).
We further computed the $\Delta\text{ADR}$ for each secure-source set within Mistral-7B-v0.1 and Phi-1.5 in Appendix~\ref{append::Experiment transition layer}. The results for the Mistral-7B-v0.1 model are presented in Table~\ref{Mistral-1} and Table~\ref{Mistral-1.1}. Additionally, the performance outcomes for the Phi-1.5 model can be found in Table~\ref{Phi1.5-1}. 

In all the above tables, ``Pretrain'' represents the model's original performance without any layers secured. These columns indicate the index of layers in the model that have been secured. ``*'' indicates fully-secured. All evaluation scores are averages from three different seed tests, corresponding to the values 20, 42, and 1234, following the details of the Sixteen Functionality Benchmarks in Appendix~\ref{appendix:evaluation datasets}.

\begin{table*}[t]
    \centering
    {
    \small
    \setlength{\tabcolsep}{3pt}
    \begin{tabular}{@{}ccccccccccc@{}}
    \toprule
       &&\textbf{Pretrain} & \textbf{0} & \textbf{1} & \textbf{2} & \textbf{3} & \textbf{4} & \textbf{5} & \textbf{6} & \textbf{7} \\ \midrule
       \multirow{5}{*}{\textbf {Rsn.}} & PIQA & 76.66  & 49.56  & 51.43  & 49.53  & 50.45  & 49.84  & 50.27  & 50.96  & 51.09   \\ 
        ~ & Hellaswag & 75.45  & 25.56  & 25.75  & 25.88  & 26.16  & 25.91  & 27.20  & 29.39  & 28.89   \\ 
        ~ & Winogrande & 66.38  & 50.99  & 50.86  & 50.15  & 49.75  & 49.96  & 50.91  & 51.64  & 51.36   \\ 
        ~ & ARC\_easy & 74.41  & 27.04  & 27.23  & 26.10  & 26.30  & 25.51  & 26.44  & 28.24  & 27.96   \\ 
        ~ & ARC\_challenge & 44.11  & 21.07  & 20.31  & 20.19  & 21.30  & 22.04  & 21.56  & 20.62  & 22.92   \\ \midrule
       \multirow{5}{*}{\textbf {Read.} } & OpenBookQA & 68.49  & 0.01  & 0.11  & 0.02  & 0.02  & 0.01  & 0.00  & 0.05  & 0.04   \\ 
        ~ & LAMBADA & 80.67  & 44.30  & 41.22  & 38.36  & 41.43  & 38.08  & 38.14  & 38.40  & 41.55   \\ 
        ~ & BoolQ & 59.48  & 0.00  & 0.04  & 0.00  & 0.00  & 0.00  & 0.00  & 0.01  & 0.03   \\ 
        ~ & SQuADv2\_em & 71.88  & 1.05  & 1.31  & 0.63  & 1.07  & 0.45  & 0.44  & 1.13  & 1.10   \\ 
        ~ & SQuADv2\_f1 & 43.80  & 25.13  & 24.60  & 23.60  & 24.93  & 25.67  & 24.47  & 25.07  & 26.00   \\ \midrule
        \multirow{2}{*}{\textbf {Knl.}} & NaturalQuestions  & 22.47  & 0.01  & 0.00  & 0.01  & 0.03  & 0.02  & 0.01  & 0.13  & 0.08   \\ 
        ~ & TriviaQA & 57.23  & 0.00  & 0.01  & 0.00  & 0.02  & 0.01  & 0.01  & 0.07  & 0.10   \\ \midrule
        \multirow{2}{*}{\textbf{Code}} & HumanEval & 10.90  & 0.00  & 0.00  & 0.00  & 0.00  & 0.00  & 0.00  & 0.00  & 0.00   \\ 
        ~ & MBPP & 16.60  & 0.00  & 0.00  & 0.00  & 0.00  & 0.00  & 0.00  & 0.00  & 0.00   \\ \midrule
         \textbf{Math} & GSM8K & 20.24  & 0.00  & 0.00  & 0.00  & 0.00  & 0.00  & 0.00  & 0.00  & 0.00   \\ \midrule
        \multirow{2}{*}{\textbf{Gen.}} & MMLU & 45.83  & 24.26  & 25.37  & 23.98  & 24.26  & 24.75  & 24.01  & 25.23  & 27.45   \\ 
        ~ & BBH & 39.86  & 0.00  & 0.00  & 0.00  & 0.00  & 0.00  & 0.00  & 0.50  & 0.38   \\ \midrule
        \multicolumn{2}{c}{\textbf{Avg. Performance Score($\downarrow$)}} & 51.44  & 15.82  & 15.78  & 15.20  & 15.63  & 15.43  & 15.50  & 15.97  & 16.41   \\ 
        \multicolumn{2}{c}{\textbf{Average Distillation Ratio($\downarrow$)}}  & -  & 30.76  & 30.67  & 29.55  & 30.39  & 29.99  & 30.13  & 31.04  & 31.90   \\ 
        \multicolumn{2}{c}{\textbf{Distillation Difficulty($\uparrow$)}}  & - & 11.11  & 11.27  & 10.87  & 10.31  & 10.83  & 10.33  & 10.90  & 11.11   \\
    \bottomrule
    \end{tabular}}
    \caption{Evaluation Results of Llama2-7B under Different Secure Layers (Part1)}
    \label{llama-1.1}
\end{table*}

\begin{table*}[t]
    \centering
    {
    \small
    \setlength{\tabcolsep}{3pt}
    \begin{tabular}{@{}cccccccccccc@{}}
    \toprule
    && \textbf{16} & \textbf{18} & \textbf{20} & \textbf{22} & \textbf{24} & \textbf{26} & \textbf{28} & \textbf{30} & \textbf{*} \\
    \midrule
    \multirow{5}{*}{\textbf{Rsn.}} & PIQA & 51.47 & 52.99 & 58.22 & 65.83 & 69.60 & 73.45 & 75.46 & 75.99 & 49.47 \\
    & Hellaswag & 31.38 & 36.55 & 45.61 & 56.60 & 62.70 & 67.88 & 71.37 & 72.94 & 26.39 \\
    & Winogrande & 53.09 & 55.98 & 58.96 & 64.12 & 64.80 & 65.25 & 65.46 & 66.53 & 50.83 \\
    & ARC\_easy & 30.58 & 35.35 & 43.85 & 55.92 & 62.56 & 68.36 & 70.85 & 72.60 & 25.98 \\
    & ARC\_challenge & 24.26 & 26.85 & 30.97 & 35.38 & 38.17 & 41.41 & 43.00 & 44.17 & 22.47 \\
    \midrule
    \multirow{5}{*}{\textbf{Read.}} & OpenBookQA & 0.28 & 1.58 & 6.79 & 30.88 & 44.58 & 56.23 & 62.33 & 63.11 & 0.01 \\
    & LAMBADA & 57.55 & 70.53 & 71.36 & 78.85 & 79.69 & 80.29 & 79.39 & 80.40 & 48.34 \\
    & BoolQ & 0.08 & 0.90 & 2.34 & 7.07 & 6.04 & 6.87 & 3.54 & 9.46 & 0.00 \\
    & SQuADv2\_em & 2.21 & 13.48 & 21.47 & 35.72 & 36.96 & 39.32 & 37.08 & 42.08 & 0.59 \\
    & SQuADv2\_f1 & 27.33 & 28.20 & 30.47 & 32.13 & 34.93 & 39.27 & 39.93 & 41.53 & 25.93 \\
    \midrule
    \multirow{2}{*}{\textbf{Knl.}} & NaturalQuestions & 0.13 & 0.41 & 1.60 & 2.94 & 4.29 & 2.69 & 7.28 & 11.87 & 0.04 \\
    & TriviaQA & 0.25 & 1.79 & 4.93 & 11.02 & 15.73 & 17.95 & 33.19 & 42.26 & 0.02 \\
    \midrule
    \multirow{2}{*}{\textbf{Code}} & HumanEval & 0.00 & 0.00 & 0.00 & 0.00 & 0.00 & 3.25 & 8.34 & 10.98 & 0.00 \\
    & MBPP & 0.00 & 0.00 & 0.00 & 0.07 & 0.47 & 2.27 & 8.80 & 13.27 & 0.00 \\
    \midrule
    \textbf{Math} & GSM8K & 0.00 & 0.00 & 0.00 & 0.13 & 0.81 & 8.42 & 6.90 & 15.77 & 0.00 \\
    \midrule
    \multirow{2}{*}{\textbf{Gen.}} & MMLU & 43.17 & 48.20 & 49.38 & 49.58 & 49.72 & 50.03 & 50.75 & 50.61 & 24.45 \\
    & BBH & 0.76 & 11.44 & 19.79 & 28.87 & 31.16 & 35.98 & 38.24 & 40.54 & 0.00 \\
    \midrule
    \multicolumn{2}{c}{\textbf{Avg. Performance Score($\downarrow$)}} & 18.97 & 22.60 & 26.22 & 32.65 & 35.42 & 38.76 & 41.29 & 44.36 & 16.15 \\
    \multicolumn{2}{c}{\textbf{Average Distillation Ratio($\downarrow$)}} & 36.89 & 43.94 & 50.98 & 63.48 & 68.87 & 75.35 & 80.27 & 86.24 & 31.39 \\
    \multicolumn{2}{c}{\textbf{Distillation Difficulty($\uparrow$)}} & 10.42 & 9.49 & 8.86 & 7.12 & 6.14 & 4.72 & 3.40 & 3.06 & 11.19 \\
    \bottomrule
    \end{tabular}}
    \caption{Evaluation Results of Llama2-7B under Different Secured Layers (Part2). ``*'' indicates the fully secured model.}
    \label{llama-1.2}
\end{table*}

\begin{table*}[t]
    \centering
    {
    \small
    \setlength{\tabcolsep}{3pt}
    \begin{tabular}{@{}ccccccccccc@{}}
    \toprule
        \textbf{} & \textbf{} & \textbf{Pretrain}& \textbf{0}  & \textbf{1} & \textbf{2} & \textbf{3} & \textbf{4} & \textbf{5} & \textbf{6} & \textbf{7} \\ \midrule
        \multirow{5}{*}{\textbf {Rsn.}} & PIQA & 81.99  & 51.63  & 53.20  & 53.63  & 53.47  & 51.56  & 52.61  & 50.71  & 55.15   \\ 
        ~ & Hellaswag & 81.04  & 26.10  & 26.36  & 26.36  & 26.66  & 27.10  & 25.51  & 26.18  & 28.16   \\ 
        ~ & Winogrande & 74.03  & 49.78  & 49.78  & 51.01  & 50.38  & 49.91  & 50.14  & 49.70  & 51.17   \\ 
        ~ & ARC\_easy & 80.77  & 33.03  & 31.96  & 30.71  & 29.66  & 30.25  & 30.35  & 26.44  & 32.38   \\ 
        ~ & ARC\_challenge & 50.26  & 19.94  & 21.27  & 20.45  & 19.60  & 20.05  & 21.36  & 21.25  & 20.73   \\ \midrule
        \multirow{5}{*}{\textbf {Read.} } & OpenBookQA & 44.40  & 25.60  & 25.20  & 25.20  & 25.47  & 25.87  & 26.33  & 25.07  & 27.20   \\ 
        ~ & LAMBADA & 73.29  & 0.12  & 0.44  & 1.91  & 2.08  & 0.80  & 0.30  & 0.17  & 1.95   \\ 
        ~ & BoolQ & 83.67  & 39.05  & 53.12  & 45.95  & 38.61  & 47.35  & 38.06  & 46.44  & 47.66   \\ 
        ~ & SQuADv2\_em & 64.04  & 0.00  & 0.00  & 0.01  & 0.01  & 0.01  & 0.00  & 0.00  & 0.01   \\ 
        ~ & SQuADv2\_f1 & 71.37  & 1.21  & 0.84  & 1.05  & 1.03  & 1.27  & 0.43  & 0.07  & 0.86   \\ \midrule
       \multirow{2}{*}{\textbf {Knl.} } &NaturalQuestions  & 28.98  & 0.00  & 0.01  & 0.00  & 0.04  & 0.01  & 0.00  & 0.02  & 0.07   \\ 
        ~ & TriviaQA & 70.79  & 0.00  & 0.00  & 0.02  & 0.01  & 0.01  & 0.01  & 0.00  & 0.16   \\ \midrule
        \multirow{2}{*}{\textbf{Code}} & HumanEval & 29.88  & 0.00  & 0.00  & 0.00  & 0.00  & 0.00  & 0.00  & 0.00  & 0.00   \\ 
        ~ & MBPP & 38.40  & 0.00  & 0.00  & 0.00  & 0.00  & 0.00  & 0.00  & 0.00  & 0.00   \\ \midrule
        \textbf{Math} & GSM8K & 38.21  & 0.00  & 0.00  & 0.00  & 0.00  & 0.00  & 0.00  & 0.00  & 0.00   \\ \midrule
        \multirow{2}{*}{\textbf{Gen.}} & MMLU & 62.50  & 25.24  & 24.68  & 25.11  & 23.43  & 23.65  & 24.26  & 24.26  & 24.99   \\ 
        ~ & BBH & 56.40  & 0.00  & 0.00  & 0.00  & 0.00  & 0.00  & 0.00  & 0.00  & 0.01   \\ \midrule
        \multicolumn{2}{c}{\textbf{Avg. Performance Score($\downarrow$)}} & 60.59  & 15.98  & 16.87  & 16.55  & 15.91  & 16.34  & 15.84  & 15.90  & 17.09   \\ 
        \multicolumn{2}{c}{\textbf{Average Distillation Ratio($\downarrow$)}} &  - & 26.38  & 27.85  & 27.32  & 26.25  & 26.97  & 26.15  & 26.24  & 28.20   \\ 
         \multicolumn{2}{c}{\textbf{Average Distillation Ratio($\downarrow$)}} & - & 11.50  & 11.31  & 11.48  & 10.71  & 10.77  & 11.44  & 11.02  & 10.71   \\ 
    \bottomrule
   \end{tabular}}
    \caption{Evaluation Results of Mistral-7B under Different Secured Layers (Part1)}
    \label{Mistral-1}
\end{table*}

\begin{table*}[t]
    \centering
    {
    \small
    \setlength{\tabcolsep}{3pt}
    \begin{tabular}{@{}cccccccccccc@{}}
    \toprule
     & & \textbf{16} & \textbf{18} & \textbf{20} & \textbf{22} & \textbf{24} & \textbf{26} & \textbf{28} & \textbf{30} & \textbf{*} \\
    \midrule
    \multirow{5}{*}{\textbf{Rsn.}} & PIQA & 54.50 & 52.32 & 52.72 & 57.13 & 62.82 & 64.67 & 67.23 & 75.61 & 49.35 \\
    & Hellaswag & 29.31 & 29.02 & 29.99 & 33.46 & 46.21 & 52.12 & 52.46 & 67.73 & 25.39 \\
    & Winogrande & 51.20 & 54.17 & 51.07 & 55.75 & 58.59 & 62.41 & 63.09 & 66.33 & 50.59 \\
    & ARC\_easy & 32.84 & 29.35 & 30.80 & 38.04 & 47.24 & 51.99 & 54.74 & 69.95 & 25.83 \\
    & ARC\_challenge & 21.19 & 23.04 & 23.78 & 26.34 & 30.86 & 33.22 & 35.04 & 40.53 & 22.35 \\
    \midrule
    \multirow{5}{*}{\textbf{Read.}} & OpenBookQA & 26.00 & 27.87 & 26.87 & 29.67 & 28.73 & 32.67 & 33.40 & 36.40 & 25.00 \\
    & LAMBADA & 2.61 & 0.18 & 1.28 & 4.17 & 21.89 & 29.93 & 24.49 & 48.32 & 0.01 \\
    & BoolQ & 53.98 & 53.60 & 58.79 & 55.76 & 64.10 & 74.72 & 68.48 & 81.30 & 45.80 \\
    & SQuADv2\_em & 0.01 & 0.00 & 0.47 & 0.13 & 2.39 & 3.59 & 1.87 & 1.82 & 0.00 \\
    & SQuADv2\_f1 & 0.96 & 0.18 & 1.27 & 2.60 & 14.88 & 22.61 & 21.12 & 34.16 & 0.66 \\
    \midrule
    \multirow{2}{*}{\textbf{Knl.}} & NaturalQuestions & 0.01 & 0.10 & 0.19 & 0.58 & 1.84 & 3.15 & 3.53 & 8.87 & 0.02 \\
    & TriviaQA & 0.03 & 0.01 & 0.61 & 0.62 & 5.14 & 7.51 & 10.32 & 25.44 & 0.01 \\
    \midrule
    \multirow{2}{*}{\textbf{Code}} & HumanEval & 0.00 & 0.00 & 0.00 & 0.61 & 2.24 & 4.88 & 2.44 & 9.75 & 0.00 \\
    & MBPP & 0.00 & 0.00 & 0.00 & 2.00 & 4.33 & 8.33 & 0.93 & 13.07 & 0.00 \\
    \midrule
    \textbf{Math} & GSM8K & 0.00 & 0.00 & 0.00 & 0.00 & 0.00 & 0.00 & 0.00 & 0.25 & 0.00 \\
    \midrule
    \multirow{2}{*}{\textbf{Gen.}} & MMLU & 24.30 & 25.84 & 29.54 & 24.55 & 34.77 & 40.77 & 40.84 & 50.44 & 23.26 \\
    & BBH & 0.00 & 0.00 & 0.02 & 0.30 & 7.55 & 18.76 & 21.05 & 30.07 & 0.00 \\
    \midrule
    \multicolumn{2}{c}{\textbf{Avg. Performance Score($\downarrow$)}} & 17.47 & 17.39 & 18.08 & 19.51 & 25.51 & 30.08 & 29.47 & 38.83 & 15.78 \\
    \multicolumn{2}{c}{\textbf{Average Distillation Ratio($\downarrow$)}} & 28.83 & 28.71 & 29.84 & 32.20 & 42.09 & 49.64 & 48.64 & 64.08 & 26.05 \\
    \multicolumn{2}{c}{\textbf{Average Distillation Ratio($\downarrow$)}} & 11.34 & 11.11 & 10.45 & 10.59 & 10.23 & 10.34 & 9.59 & 8.53 & 11.20 \\
    \bottomrule
    \end{tabular}}
    \caption{Evaluation Results of Mistral-7B under Different Secured Layers (Part2)}
    \label{Mistral-1.1}
\end{table*}

\begin{table*}[t]
    \centering
    {
    \small
    \setlength{\tabcolsep}{3pt}
    \begin{tabular}{@{}ccccccccccc@{}}
    \toprule
        & & \textbf{Pretrain} & \textbf{0} & \textbf{2} & \textbf{4} & \textbf{6} & \textbf{8} & \textbf{10} & \textbf{12} & \textbf{14} \\ 
        \midrule
        \multirow{5}{*}{\textbf {Rsn.}} & PIQA & 79.27 & 54.17 & 72.85 & 73.76 & 75.03 & 76.75 & 78.00 & 78.91 & 77.84 \\
        & Hellaswag & 73.73 & 27.61 & 56.49 & 57.73 & 60.47 & 62.84 & 66.39 & 66.91 & 66.95 \\
        & Winogrande & 75.45 & 51.56 & 59.17 & 59.98 & 59.88 & 64.32 & 68.11 & 68.95 & 70.38 \\
        & ARC\_easy & 79.92 & 34.57 & 72.94 & 73.40 & 73.97 & 76.51 & 78.33 & 78.66 & 78.63 \\
        & ARC\_challenge & 52.90 & 19.45 & 41.75 & 39.82 & 44.11 & 45.65 & 47.92 & 49.74 & 48.78 \\\midrule
        \multirow{5}{*}{\textbf {Read.}} & OpenBookQA & 51.20 & 25.80 & 35.73 & 37.47 & 40.13 & 42.00 & 44.00 & 45.67 & 44.80 \\
        & LAMBADA & 56.28 & 3.25 & 28.55 & 30.42 & 34.64 & 40.05 & 45.41 & 45.52 & 46.66 \\
        & BoolQ & 83.36 & 47.29 & 65.20 & 62.64 & 66.39 & 71.39 & 73.42 & 72.95 & 75.83 \\
        & SQuADv2\_em & 61.30 & 0.02 & 10.49 & 17.63 & 21.94 & 33.94 & 19.54 & 19.15 & 29.14 \\
        & SQuADv2\_f1 & 71.38 & 2.61 & 37.22 & 40.35 & 45.53 & 59.16 & 48.21 & 50.09 & 54.87 \\\midrule
        \multirow{2}{*}{\textbf {Knl.}} & NaturalQuestions & 9.58 & 0.00 & 3.60 & 4.97 & 6.13 & 7.55 & 7.95 & 8.10 & 9.25 \\
        & TriviaQA & 39.29 & 0.01 & 13.57 & 16.29 & 24.74 & 28.60 & 31.58 & 33.71 & 32.79 \\\midrule
        \multirow{2}{*}{\textbf{Code}} & HumanEval & 48.78 & 0.00 & 1.42 & 6.50 & 10.98 & 16.66 & 22.76 & 19.51 & 23.17 \\
        & MBPP & 46.80 & 0.00 & 5.07 & 6.87 & 9.47 & 19.60 & 25.67 & 23.47 & 25.73 \\\midrule
        \textbf{Math} & GSM8K & 57.77 & 0.00 & 7.25 & 8.64 & 4.42 & 9.63 & 14.18 & 11.35 & 17.31 \\\midrule
        \multirow{2}{*}{\textbf{Gen.}} & MMLU & 56.73 & 26.16 & 34.29 & 37.01 & 39.90 & 43.11 & 45.63 & 48.17 & 49.82 \\
        & BBH & 59.53 & 0.01 & 15.27 & 18.37 & 16.38 & 14.58 & 4.93 & 4.35 & 11.37 \\
        \midrule
        \multicolumn{2}{c}{\textbf{Avg. Performance Score($\downarrow$)}} & 59.02 & 17.21 & 32.99 & 34.81 & 37.30 & 41.90 & 42.47 & 42.66 & 44.90 \\
        \multicolumn{2}{c}{\textbf{Average Distillation Ratio($\downarrow$)}} & - & 29.15 & 55.90 & 58.99 & 63.21 & 71.00 & 71.97 & 72.28 & 76.09 \\
        \multicolumn{2}{c}{\textbf{Distillation Difficulty($\uparrow$)}} & - & 10.07 & 7.07 & 4.95 & 4.09 & 3.63 & 3.31 & 3.31 & 3.11 \\
        \bottomrule
    \end{tabular}}
    
    \caption{Evaluation Results of Phi-2 under Different Secured Layers (Part 1)}
    \label{Phi2-1-part1}
\end{table*}

\begin{table*}[t]
    \centering
    {
    \small
    \setlength{\tabcolsep}{3pt}
    \begin{tabular}{@{}cccccccccccc@{}}
    \toprule
     &  & \textbf{16} & \textbf{18} & \textbf{20} & \textbf{22} & \textbf{24} & \textbf{26} & \textbf{28} & \textbf{30} & \textbf{*} \\
    \midrule
    \multirow{2}{*}{\textbf{Rsn.}} & PIQA & 77.44 & 77.80 & 77.69 & 76.77 & 76.89 & 77.55 & 78.16 & 78.58 & 52.07 \\
    & Hellaswag  & 67.20 & 66.90 & 67.13 & 68.00 & 68.86 & 70.01 & 71.44 & 71.18 & 25.26 \\
    & Winogrande & 70.82 & 71.40 & 73.11 & 74.46 & 75.79 & 75.72 & 75.93 & 74.77 & 48.91 \\
    & ARC\_easy & 78.30 & 77.27 & 77.33 & 76.82 & 78.09 & 77.76 & 79.53 & 79.56 & 27.03 \\
    & ARC\_challenge & 49.71 & 48.29 & 48.52 & 48.04 & 49.80 & 50.68 & 53.16 & 52.67 & 18.66 \\
    \midrule
    \multirow{2}{*}{\textbf{Read.}} & OpenBookQA & 46.53 & 46.47 & 45.87 & 45.27 & 46.33 & 45.53 & 46.53 & 48.27 & 20.80 \\
    & LAMBADA & 45.67 & 46.88 & 47.95 & 50.17 & 50.54 & 52.77 & 53.01 & 53.23 & 0.00 \\
    & BoolQ & 80.56 & 80.72 & 82.22 & 83.31 & 83.98 & 83.54 & 82.54 & 83.41 & 39.60 \\
    & SQuADv2\_em & 7.88 & 1.30 & 1.69 & 1.31 & 0.15 & 0.23 & 3.54 & 10.03 & 0.56 \\
    & SQuADv2\_f1 & 40.84 & 34.51 & 34.25 & 35.94 & 35.64 & 36.68 & 39.57 & 44.87 & 0.90 \\
    \midrule
    \multirow{2}{*}{\textbf{Knl.}} & NaturalQuestions & 8.90 & 6.09 & 6.40 & 6.79 & 6.86 & 6.85 & 7.20 & 8.37 & 0.02 \\
    & TriviaQA & 31.48 & 27.03 & 25.08 & 24.54 & 22.89 & 22.99 & 24.24 & 26.93 & 0.01 \\
    \midrule
    \multirow{2}{*}{\textbf{Code}} & HumanEval & 22.56 & 21.34 & 25.41 & 32.52 & 38.01 & 46.14 & 46.54 & 43.90 & 0.00 \\
    & MBPP & 26.73 & 25.33 & 24.80 & 31.73 & 36.67 & 41.80 & 43.13 & 43.20 & 0.00 \\
    \midrule
    \textbf{Math} & GSM8K & 16.68 & 16.02 & 14.66 & 12.31 & 17.24 & 30.12 & 45.41 & 49.79 & 0.00 \\
    \midrule
    \multirow{2}{*}{\textbf{Gen.}} & MMLU & 52.69 & 53.45 & 55.68 & 56.61 & 56.93 & 56.59 & 56.86 & 56.47 & 22.95 \\
    & BBH & 3.42 & 17.36 & 8.33 & 18.24 & 30.09 & 48.12 & 52.28 & 56.36 & 0.00 \\
    \midrule
    \multicolumn{2}{c}{\textbf{Avg. Performance Score($\downarrow$)}} & 42.79 & 42.25 & 42.12 & 43.70 & 45.57 & 48.42 & 50.53 & 51.86 & 15.10 \\
    \multicolumn{2}{c}{\textbf{Average Distillation Ratio($\downarrow$)}} & 72.51 & 71.58 & 71.38 & 74.04 & 77.22 & 82.04 & 85.63 & 87.87 & 25.59 \\
    \multicolumn{2}{c}{\textbf{Distillation Difficulty($\uparrow$)}} & 3.07 & 3.29 & 3.03 & 3.01 & 2.70 & 2.32 & 1.98 & 2.13 & 11.32 \\
    \bottomrule
    \end{tabular}}
    \caption{Evaluation Results of Phi-2 under Different Secured Layers (Part2). ``*'' indicates the fully secured model.}
    \label{Phi2-1-part2}
\end{table*}

\begin{table*}[t]
    \centering
    {\small
    \setlength{\tabcolsep}{1.2pt}
    \begin{tabular}{@{}cccccccccccccccc@{}}
    \toprule
    &&\textbf{Pretrain}&\textbf{0-1}&\textbf{2-3}&\textbf{4-5}&\textbf{6-7}&\textbf{8-9}&\textbf{10-11}&\textbf{12-13}&\textbf{14-15}&\textbf{16-17}&\textbf{18-19}&\textbf{20-21}&\textbf{22-23}&\textbf{*}  \\\midrule
        \multirow{5}{*}{\textbf{Rsn.}}&PIQA &75.68  &53.43  &69.52  &71.53  &73.50  &74.76  &75.08  &74.94  &74.64  &73.90  &74.63  &74.54  &74.81&50.44 \\
        &Hellaswag &62.56  &26.27  &46.66  &50.71  &52.98  &54.51  &55.11  &56.01  &56.78  &57.90  &58.76  &59.35  &58.58 &25.05\\
        &Winogrande &72.69  &51.09  &54.91  &59.22  &61.75  &64.85  &67.95  &68.88  &68.98  &71.25  &71.19  &72.87  &70.66 &49.12\\
        &ARC\_easy &76.14  &30.81  &61.70  &65.70  &70.10  &71.38  &70.01  &71.72  &71.93  &72.34  &73.39  &74.16  &73.74 &27.50\\
        &ARC\_challenge &44.62  &20.56  &32.85  &34.10  &38.08  &40.05  &40.30  &39.48  &40.87  &41.52  &42.84  &42.58  &45.42 &21.22\\\midrule
        \multirow{5}{*}{\textbf{Read.}}&OpenBookQA &48.00  &26.60  &33.93  &35.73  &40.40  &41.13  &40.67  &41.73  &41.67  &40.27  &41.33  &43.27  &45.47 &26.87 \\
        &LAMBADA &44.10  &0.59  &17.96  &26.45  &29.37  &33.83  &33.85  &36.46  &37.06  &37.96  &39.98  &41.10  &40.49 &0.00\\
        &BoolQ &75.05  &46.98  &59.12  &52.42  &57.41  &65.68  &68.52  &63.47  &65.12  &66.52  &73.91  &75.17  &77.0 &46.28\\
        &SQuADv2\_em &48.01  &0.00  &5.82  &10.94  &18.34  &13.96  &14.70  &23.22  &16.98  &26.05  &22.04  &20.16  &26.86 &0.00\\
        &SQuADv2\_f1 &60.84  &0.78  &24.49  &26.04  &34.86  &32.17  &32.36  &43.14  &38.23  &48.03  &45.75  &45.56  &49.62  &1.60\\\midrule
        \multirow{2}{*}{\textbf{Knl.}}&NaturalQuestions  &5.46  &0.04  &1.68  &2.73  &3.41  &3.06  &3.21  &4.25  &4.03  &4.06  &4.54  &4.17  &4.45 &0.01 \\
        &TriviaQA &16.94  &0.01  &5.70  &7.77  &10.85  &11.03  &9.11  &12.11  &11.84  &11.86  &12.02  &12.11  &13.19 &0.01\\\midrule
        \multirow{2}{*}{\textbf{Code}}&HumanEval &35.98  &0.00  &3.05  &10.57  &12.20  &16.26  &13.82  &17.48  &18.70  &23.17  &29.68  &31.91  &31.71 &0.00\\
        &MBPP &35.40  &0.00  &2.80  &7.80  &10.93  &17.40  &16.53  &16.13  &16.67  &22.27  &27.33  &28.27  &28.53 &0.00\\\midrule
        \textbf{Math}&GSM8K &30.33  &0.00  &0.05  &0.73  &0.15  &0.23  &0.75  &0.50  &2.17  &4.98  &9.73  &17.77  &23.45 &0.00 \\\midrule
        \multirow{2}{*}{\textbf{Gen.}}&MMLU &42.44  &24.07  &26.56  &28.77  &32.51  &32.87  &36.09  &39.42  &39.72  &43.23  &42.51  &42.82  &43.66 &23.95 \\
        &BBH &28.80  &0.00  &2.07  &3.97  &8.38  &7.37  &2.81  &7.79  &4.12  &10.63  &6.94  &10.34  &11.45 &0.00 \\\midrule
        \multicolumn{2}{c}{\textbf{Avg. Performance Score($\downarrow$)}}&47.24  &16.54  &26.40  &29.13  &32.66  &34.15  &34.17  &36.28  &35.85  &38.59  &39.80  &40.95  &42.30 &15.94 \\
         \multicolumn{2}{c}{\textbf{Average Distillation Ratio($\downarrow$)}}&-  &35.02  &55.90  &61.66  &69.14  &72.29  &72.34  &76.80  &75.90  &81.68  &84.25  &86.69  &89.56 &33.75 \\
        \multicolumn{2}{c}{\textbf{Average Distillation Ratio($\downarrow$)}}&-&10.08  &7.18  &4.70  &3.50  &2.93  &2.83  &2.53  &2.36  &2.27  &2.16  &2.06  &2.46 &9.33\\\bottomrule
    \end{tabular}}
    \caption{Evaluation Results of Phi-1.5 under Different Secured Layers}
    \label{Phi1.5-1}
\end{table*}

\textbf{Customizability Performance. }We close varying numbers of layers from the start and fine-tune the open set, and then we observe the customizability transition in models. Table~\ref{fig: cust transition result} shows the detailed evaluation results of Llama2-7B and Phi-2 on GSM8k benchmark.

\begin{table*}[t]
    \centering
    \begin{tabular}{cc|cc}
    \toprule
        \multicolumn{2}{c}{\textbf{Llama2-7B}} & \multicolumn{2}{c}{\textbf{Phi-2}} \\
        \textbf{Secure Layers} & \textbf{GSM8K}($\uparrow$) & \textbf{Secure Layers} & \textbf{GSM8K}($\uparrow$) \\ 
        \midrule
        \textbf{Fully-open} & 29.34  & \textbf{Fully-open} & 59.60   \\ 
        \textbf{0} & 28.96  & \textbf{0-1} & 59.59   \\
        \textbf{0-4} & 21.76  & \textbf{0-5} & 58.60   \\ 
        \textbf{0-8} & 21.46  & \textbf{0-9} & 58.45   \\ 
        \textbf{0-12} & 20.85  & \textbf{0-13} & 55.19   \\ 
        \textbf{0-16} & 20.11  & \textbf{0-17} & 56.25   \\ 
        \textbf{0-20} & 21.46  & \textbf{0-21} & 54.59   \\ 
        \textbf{0-24} & 21.44  & \textbf{0-25} & 55.34   \\ 
        \textbf{0-28} & 18.73  & \textbf{0-29} & 54.59   \\ 
        \textbf{Fully-Secure} & 20.32  & \textbf{Fully-Secure} & 57.77  \\  
        \bottomrule
    \end{tabular}
    \caption{Customization Performance under Different Secure Sets}
    \label{fig: cust transition result}
\end{table*}

\subsection{Evaluation Results under Different Secure size}
\label{append::Experiment Secure Size}

In this section, we present a comprehensive evaluation of the model's performance across sixteen benchmarks utilized in our experiments. The evaluation results for LLaMA2-7B, categorized by varying quantities and proportions of secure-source parameters, are displayed in Table~\ref{tab:appendix:result:llama2-Close-source Proportion} and Table~\ref{tab:appendix:result:llama2-Close-source Quantity}, respectively.
For the Mistral-7B model, the results are summarized in Table~\ref{tab:appendix:result:mistral-2} and Table~\ref{tab:appendix:result:mistral-3}. Furthermore, the evaluation outcomes for the Phi-2 model can be found in Tables~\ref{tab:appendix:result:Phi2-2} and Table~\ref{tab:appendix:result:Phi2-3}. The performance results for Phi-1.5 are also included in Tables~\ref{tab:appendix:result:Phi1.5-2} and Table~\ref{tab:appendix:result:Phi1.5-3} for comparison.
For further details regarding the secure-source settings employed in our experiments, please refer to Appendix~\ref{append::Experiment Secure Size}.

\begin{table*}[t]
    \centering
    {\small
    \setlength{\tabcolsep}{3pt}
    \begin{tabular}{@{}cccccccccccc@{}}
    \toprule
        \textbf{} & \textbf{} & \textbf{0.25\%} & \textbf{0.5\%} & \textbf{1\%} & \textbf{3\%} & \textbf{7\%} & \textbf{15\%} & \textbf{30\%} & \textbf{50\%} & \textbf{100\%} \\ \midrule
        \multirow{5}{*}{\textbf {Rsn.}} & PIQA & 77.78  & 77.69  & 67.73  & 49.42  & 49.55  & 50.05  & 49.98  & 49.31  & 49.47   \\ 
        \textbf{} & Hellaswag & 71.40  & 71.54  & 52.39  & 25.74  & 26.03  & 26.25  & 25.67  & 25.48  & 26.39   \\ 
        \textbf{} & Winogrande & 64.64  & 65.64  & 54.12  & 50.38  & 50.43  & 49.65  & 49.59  & 49.62  & 50.83   \\ 
        \textbf{} & ARC\_easy & 74.69  & 75.04  & 53.82  & 26.03  & 26.76  & 26.46  & 26.64  & 26.66  & 25.98   \\ 
        \textbf{} & ARC\_challenge & 43.66  & 43.29  & 26.99  & 20.16  & 21.39  & 19.74  & 21.44  & 21.73  & 22.47   \\ \midrule
        \multirow{5}{*}{\textbf {Read.} } & OpenBookQA & 63.15  & 63.62  & 33.20  & 0.01  & 0.00  & 0.02  & 0.01  & 0.01  & 0.01   \\ 
        \textbf{} & LAMBADA & 80.66  & 80.78  & 62.10  & 38.22  & 39.33  & 43.45  & 39.39  & 41.83  & 48.34   \\ 
        \textbf{} & BoolQ & 11.39  & 12.14  & 5.47  & 0.00  & 0.00  & 0.00  & 0.00  & 0.00  & 0.00   \\ 
        \textbf{} & SQuADv2\_em & 40.24  & 40.74  & 32.65  & 0.78  & 0.20  & 0.24  & 2.09  & 2.13  & 0.59   \\ 
        \textbf{} & SQuADv2\_f1 & 40.73  & 40.67  & 30.47  & 22.93  & 23.40  & 25.53  & 24.07  & 23.07  & 25.93   \\ \midrule
        \multirow{2}{*}{\textbf {Knl.} } & NaturalQuestions  & 7.83  & 7.89  & 5.61  & 0.00  & 0.01  & 0.02  & 0.01  & 0.00  & 0.04   \\ 
        \textbf{} & TriviaQA & 44.29  & 45.95  & 18.78  & 0.00  & 0.01  & 0.00  & 0.00  & 0.00  & 0.02   \\ \midrule
        \multirow{2}{*}{\textbf{Code}} & HumanEval & 11.39  & 12.00  & 0.00  & 0.00  & 0.00  & 0.00  & 0.00  & 0.00  & 0.00   \\ 
        \textbf{} & MBPP & 15.20  & 15.33  & 1.00  & 0.00  & 0.00  & 0.00  & 0.00  & 0.00  & 0.00   \\ \midrule
        \textbf{Math} & GSM8K & 13.22  & 13.29  & 0.00  & 0.00  & 0.00  & 0.00  & 0.00  & 0.00  & 0.00   \\ \midrule
        \multirow{2}{*}{\textbf{Gen.}} & MMLU & 45.04  & 45.03  & 30.90  & 24.06  & 24.04  & 25.01  & 23.19  & 23.11  & 24.45   \\ 
        \textbf{} & BBH & 37.45  & 37.51  & 17.36  & 0.00  & 0.00  & 0.00  & 0.00  & 0.00  & 0.00   \\ \midrule
        \multicolumn{2}{c}{\textbf{Avg. Performance Score($\downarrow$)}} & 43.69  & 44.01  & 28.98  & 15.16  & 15.36  & 15.67  & 15.42  & 15.47  & 16.15   \\ 
         \multicolumn{2}{c}{\textbf{Average Distillation Ratio($\downarrow$)}} & 84.94  & 85.56  & 56.33  & 29.48  & 29.86  & 30.47  & 29.97  & 30.07  & 31.39   \\ 
        \multicolumn{2}{c}{\textbf{Distillation Difficulty($\uparrow$)}} & 1.96  & 1.93  & 8.66  & 10.87  & 11.75  & 11.48  & 11.65  & 11.57  & 11.19   \\ 
    \bottomrule
    \end{tabular}}
    \caption{Evaluation Results of Llama2-7B under Different Secure-source Proportion}
    \label{tab:appendix:result:llama2-Close-source Proportion}
\end{table*}

\begin{table*}[t]
    \centering
    {\small
    \begin{tabular}{@{}cccccccccccccccc@{}}
    \toprule
        \textbf{} & \textbf{} & \textbf{20M} & \textbf{50M} & \textbf{100M} & \textbf{160M} & \textbf{200M} & \textbf{300M} & \textbf{600M } \\ \midrule
        \multirow{5}{*}{\textbf {Rsn.}} & PIQA & 77.78  & 73.49  & 67.55  & 67.12  & 49.42  & 50.36  & 49.97   \\ 
        \textbf{} & Hellaswag & 71.40  & 63.47  & 51.67  & 51.27  & 25.74  & 25.70  & 25.78   \\ 
        \textbf{} & Winogrande & 64.64  & 57.54  & 53.07  & 52.04  & 50.38  & 49.28  & 50.49   \\ 
        \textbf{} & ARC\_easy & 74.69  & 66.50  & 51.97  & 52.11  & 26.03  & 26.43  & 26.29   \\ 
        \textbf{} & ARC\_challenge & 43.66  & 36.04  & 26.51  & 25.99  & 20.16  & 20.79  & 21.70   \\ \midrule
        \multirow{5}{*}{\textbf {Read.} } & OpenBookQA & 63.15  & 45.34  & 30.22  & 28.75  & 0.01  & 0.05  & 0.01   \\ 
        \textbf{} & LAMBADA & 80.66  & 69.47  & 62.28  & 62.59  & 38.22  & 39.03  & 40.80   \\ 
        \textbf{} & BoolQ & 11.39  & 2.21  & 4.18  & 7.24  & 0.00  & 0.00  & 0.01   \\ 
        \textbf{} & SQuADv2\_em & 40.24  & 33.98  & 28.98  & 31.05  & 0.78  & 0.74  & 0.37   \\ 
        \textbf{} & SQuADv2\_f1 & 40.73  & 33.93  & 29.13  & 30.00  & 22.93  & 23.80  & 23.53   \\ \midrule
        \multirow{2}{*}{\textbf {Knl.} } & NaturalQuestions  & 7.83  & 2.98  & 5.33  & 5.73  & 0.00  & 0.00  & 0.02   \\ 
        \textbf{} & TriviaQA & 44.29  & 15.28  & 13.71  & 17.25  & 0.00  & 0.00  & 0.01   \\ \midrule
        \multirow{2}{*}{\textbf{Code}} & HumanEval & 11.39  & 0.41  & 0.00  & 0.00  & 0.00  & 0.00  & 0.00   \\ 
        \textbf{} & MBPP & 15.20  & 6.87  & 1.00  & 0.80  & 0.00  & 0.00  & 0.00   \\ \midrule
        \textbf{Math} & GSM8K & 9.00  & 0.10  & 0.00  & 0.00  & 0.00  & 0.00  & 0.00   \\ \midrule
        \multirow{2}{*}{\textbf{Gen.}} & MMLU & 45.04  & 36.15  & 28.95  & 29.04  & 24.06  & 23.70  & 23.45   \\ 
        \textbf{} & BBH & 37.45  & 28.53  & 14.99  & 16.99  & 0.00  & 0.00  & 0.00   \\ \midrule
        \multicolumn{2}{c}{\textbf{Avg. Performance Score($\downarrow$)}} & 43.44  & 33.66  & 27.62  & 28.12  & 15.16  & 15.29  & 15.44   \\ 
         \multicolumn{2}{c}{\textbf{Average Distillation Ratio($\downarrow$)}} & 84.46  & 65.44  & 53.69  & 54.66  & 29.48  & 29.72  & 30.01   \\ 
        \multicolumn{2}{c}{\textbf{Distillation Difficulty($\uparrow$)}} & 1.96  & 5.48  & 8.95  & 9.25  & 10.87  & 10.93  & 10.81   \\ 
    \bottomrule
   \end{tabular}}
    \caption{Evaluation Results of Llama2-7B under Different Secure-source Quantity}
    \label{tab:appendix:result:llama2-Close-source Quantity}
\end{table*}

\begin{table*}[t]
    \centering
    {\small
    \setlength{\tabcolsep}{3pt}
    \begin{tabular}{@{}ccccccccccc@{}}
    \toprule
    &  & \textbf{0.25\%} & \textbf{1\%} & \textbf{0.5\%} & \textbf{3\%} & \textbf{7\%} & \textbf{15\%} & \textbf{30\%} & \textbf{50\%} & \textbf{100\%} \\ \midrule
    \multirow{5}{*}{\textbf{Rsn.}} & PIQA & 77.79 & 74.36 & 52.16 & 53.34 & 52.07 & 52.19 & 50.04 & 50.60 & 49.35 \\ 
    & Hellaswag & 71.31 & 65.50 & 26.50 & 26.16 & 25.92 & 25.91 & 25.87 & 25.61 & 25.39 \\ 
    & Winogrande & 67.09 & 60.32 & 49.22 & 51.65 & 50.01 & 51.36 & 51.36 & 49.65 & 50.59 \\ 
    & ARC\_easy & 74.52 & 69.51 & 29.95 & 30.82 & 29.73 & 30.44 & 28.20 & 27.45 & 25.83 \\ 
    & ARC\_challenge & 42.32 & 38.40 & 20.76 & 20.71 & 21.10 & 20.25 & 22.61 & 22.47 & 22.35 \\ \midrule
    \multirow{5}{*}{\textbf{Read.}} & OpenBookQA & 42.13 & 34.60 & 25.13 & 25.33 & 26.47 & 26.07 & 25.20 & 25.87 & 25.00 \\ 
    & LAMBADA & 55.99 & 44.36 & 0.73 & 1.66 & 0.96 & 0.31 & 0.03 & 0.02 & 0.01 \\ 
    & BoolQ & 78.35 & 74.06 & 43.18 & 42.01 & 42.09 & 40.02 & 38.53 & 39.91 & 45.80 \\ 
    & SQuADv2\_em & 13.91 & 6.97 & 0.00 & 0.01 & 0.00 & 0.00 & 0.00 & 0.00 & 0.00 \\ 
    & SQuADv2\_f1 & 41.13 & 33.88 & 1.60 & 0.93 & 1.27 & 0.71 & 0.99 & 0.86 & 0.66 \\ \midrule
    \multirow{2}{*}{\textbf{Knl.}} & NaturalQuestions & 8.46 & 5.82 & 0.03 & 0.00 & 0.02 & 0.03 & 0.00 & 0.00 & 0.02 \\ 
    & TriviaQA & 34.04 & 17.03 & 0.01 & 0.01 & 0.02 & 0.01 & 0.00 & 0.00 & 0.01 \\ \midrule
    \multirow{2}{*}{\textbf{Code}} & HumanEval & 11.99 & 6.51 & 0.00 & 0.00 & 0.00 & 0.00 & 0.00 & 0.00 & 0.00 \\ 
    & MBPP & 16.93 & 12.80 & 0.00 & 0.00 & 0.00 & 0.00 & 0.00 & 0.00 & 0.00 \\ \midrule
    \textbf{Math} & GSM8K & 6.32 & 0.45 & 0.00 & 0.00 & 0.00 & 0.00 & 0.00 & 0.00 & 0.00 \\ \midrule
    \multirow{2}{*}{\textbf{Gen.}} & MMLU & 44.17 & 37.98 & 23.98 & 24.34 & 25.10 & 23.91 & 23.68 & 24.12 & 23.26 \\ 
    & BBH & 35.44 & 27.27 & 0.02 & 0.00 & 0.00 & 0.00 & 0.00 & 0.00 & 0.00 \\ \midrule
    \multicolumn{2}{c}{\textbf{Avg. Performance Score($\downarrow$)}} & 42.46 & 35.87 & 16.08 & 16.29 & 16.16 & 15.95 & 15.68 & 15.68 & 15.78 \\ 
    \multicolumn{2}{c}{\textbf{Average Distillation Ratio($\downarrow$)}} & 70.08 & 59.20 & 26.53 & 26.89 & 26.67 & 26.33 & 25.87 & 25.88 & 26.05 \\ 
    \multicolumn{2}{c}{\textbf{Distillation Difficulty($\uparrow$)}} & 2.22 & 5.48 & 10.92 & 11.29 & 11.35 & 11.19 & 11.17 & 11.20 & 11.20 \\
    \bottomrule
    \end{tabular}}
    \caption{Evaluation Results of Mistral-7B under Different Secured Proportion}
    \label{tab:appendix:result:mistral-2}
\end{table*}

\begin{table*}[t]
    \centering
    {\small
    \begin{tabular}{@{}cccccccccccccccc@{}}
    \midrule
        \textbf{} & \textbf{} & \textbf{20M} & \textbf{50M} & \textbf{100M} & \textbf{160M} & \textbf{200M} & \textbf{300M} & \textbf{600M } \\ \midrule
        \multirow{5}{*}{\textbf {Rsn.}} & PIQA & 77.79  & 73.74  & 51.36  & 52.86  & 53.34  & 50.98  & 51.62   \\ 
        \textbf{} & Hellaswag & 71.31  & 65.51  & 26.49  & 27.98  & 26.16  & 26.27  & 26.04   \\ 
        \textbf{} & Winogrande & 67.09  & 64.51  & 50.06  & 49.51  & 51.65  & 50.17  & 50.85   \\ 
        \textbf{} & ARC\_easy & 74.52  & 68.29  & 27.84  & 30.95  & 30.82  & 27.36  & 28.30   \\ 
        \textbf{} & ARC\_challenge & 42.32  & 37.97  & 20.85  & 21.67  & 20.71  & 21.28  & 20.17   \\ \midrule
        \multirow{5}{*}{\textbf {Read.} } & OpenBookQA & 42.13  & 37.27  & 25.60  & 25.87  & 25.33  & 26.60  & 27.00   \\ 
        \textbf{} & LAMBADA & 55.99  & 47.63  & 1.16  & 4.74  & 1.66  & 0.43  & 0.53   \\ 
        \textbf{} & BoolQ & 78.35  & 75.00  & 40.17  & 47.05  & 42.01  & 42.05  & 39.03   \\ 
        \textbf{} & SQuADv2\_em & 13.91  & 8.65  & 0.01  & 0.04  & 0.01  & 0.01  & 0.00   \\ 
        \textbf{} & SQuADv2\_f1 & 41.13  & 35.50  & 1.01  & 0.49  & 0.93  & 0.28  & 0.39   \\ \midrule
        \multirow{2}{*}{\textbf {Knl.} } & NaturalQuestions  & 8.46  & 7.82  & 0.02  & 0.05  & 0.00  & 0.01  & 0.02   \\ 
        \textbf{} & TriviaQA & 34.04  & 22.89  & 0.02  & 0.19  & 0.01  & 0.01  & 0.01   \\ \midrule
        \multirow{2}{*}{\textbf{Code}} & HumanEval & 11.99  & 7.93  & 0.00  & 0.00  & 0.00  & 0.00  & 0.00   \\ 
        \textbf{} & MBPP & 16.93  & 11.87  & 0.00  & 0.00  & 0.00  & 0.00  & 0.00   \\ \midrule
        \textbf{Math} & GSM8K & 6.32  & 2.48  & 0.00  & 0.00  & 0.00  & 0.00  & 0.00   \\ \midrule
        \multirow{2}{*}{\textbf{Gen.}} & MMLU & 44.17  & 41.28  & 24.22  & 24.44  & 24.34  & 23.78  & 23.33   \\ 
        \textbf{} & BBH & 35.44  & 33.43  & 0.00  & 0.40  & 0.00  & 0.00  & 0.00   \\ \midrule
        \multicolumn{2}{c}{\textbf{Avg. Performance Score($\downarrow$)}} & 42.46  & 37.75  & 15.81  & 16.84  & 16.29  & 15.84  & 15.72   \\ 
         \multicolumn{2}{c}{\textbf{Average Distillation Ratio($\downarrow$)}} & 70.08  & 62.31  & 26.10  & 27.79  & 26.89  & 26.14  & 25.95   \\ 
        \multicolumn{2}{c}{\textbf{Distillation Difficulty($\uparrow$)}} & 2.22  & 3.44  & 11.14  & 10.85  & 11.10  & 11.23  & 11.22   \\ 
    \bottomrule
   \end{tabular}}
    \caption{Evaluation Results of Mistral-7B under Different Secured Quantity}
    \label{tab:appendix:result:mistral-3}
\end{table*}

\begin{table*}[t]
    \centering
    {\small
    \setlength{\tabcolsep}{3pt}
    \begin{tabular}{@{}ccccccccccc@{}}
    \toprule
        \textbf{} & \textbf{} & \textbf{0.25\%} & \textbf{0.5\%} & \textbf{1\%} & \textbf{3\%} & \textbf{7\%} & \textbf{15\%} & \textbf{30\%} & \textbf{50\%} & \textbf{100\%} \\ \midrule
        \multirow{5}{*}{\textbf {Rsn.}} & PIQA & 70.40  & 70.71  & 74.64  & 54.43  & 54.17  & 54.75  & 54.37  & 52.39  & 52.07   \\ 
        \textbf{} & Hellaswag & 53.13  & 52.99  & 62.84  & 27.88  & 27.61  & 27.77  & 28.01  & 26.30  & 25.26   \\ 
        \textbf{} & Winogrande & 66.17  & 66.43  & 69.93  & 51.49  & 51.56  & 51.46  & 51.44  & 49.12  & 48.91   \\ 
        \textbf{} & ARC\_easy & 64.62  & 65.33  & 72.55  & 33.39  & 34.57  & 32.00  & 32.18  & 29.97  & 27.03   \\ 
        \textbf{} & ARC\_challenge & 43.26  & 43.86  & 40.67  & 20.82  & 19.45  & 20.00  & 20.56  & 19.88  & 18.66   \\ \midrule
        \multirow{5}{*}{\textbf {Read.} } & OpenBookQA & 41.80  & 42.67  & 38.87  & 26.87  & 25.80  & 26.33  & 26.53  & 26.07  & 20.80   \\ 
        \textbf{} & LAMBADA & 32.51  & 32.25  & 40.24  & 10.58  & 3.25  & 3.87  & 6.06  & 0.66  & 0.00   \\ 
        \textbf{} & BoolQ & 65.77  & 65.27  & 76.84  & 48.13  & 47.29  & 45.62  & 46.15  & 40.50  & 39.60   \\ 
        \textbf{} & SQuADv2\_em & 0.36  & 9.09  & 3.31  & 0.02  & 0.02  & 0.01  & 0.01  & 0.00  & 0.56   \\ 
        \textbf{} & SQuADv2\_f1 & 24.81  & 30.83  & 30.47  & 0.45  & 2.61  & 0.57  & 2.52  & 1.67  & 0.90   \\ \midrule
        \multirow{2}{*}{\textbf {Knl.} } & NaturalQuestions  & 5.70  & 5.06  & 1.14  & 0.03  & 0.00  & 0.01  & 0.07  & 0.03  & 0.02   \\ 
        \textbf{} & TriviaQA & 20.27  & 21.50  & 8.78  & 2.02  & 0.01  & 0.02  & 0.01  & 0.01  & 0.01   \\ \midrule
        \multirow{2}{*}{\textbf{Code}} & HumanEval & 22.16  & 26.83  & 17.68  & 0.00  & 0.00  & 0.00  & 0.00  & 0.00  & 0.00   \\ 
        \textbf{} & MBPP & 25.07  & 26.40  & 9.73  & 0.00  & 0.00  & 0.00  & 0.00  & 0.00  & 0.00   \\ \midrule
        \textbf{Math} & GSM8K & 29.26  & 31.36  & 2.00  & 0.00  & 0.00  & 0.00  & 0.00  & 0.00  & 0.00   \\ \midrule
        \multirow{2}{*}{\textbf{Gen.}} & MMLU & 41.76  & 42.17  & 43.86  & 30.31  & 26.16  & 25.79  & 24.85  & 24.03  & 22.95   \\ 
        \textbf{} & BBH & 18.98  & 21.55  & 9.59  & 3.06  & 0.01  & 0.79  & 0.24  & 0.00  & 0.00   \\ \midrule
        \multicolumn{2}{c}{\textbf{Avg. Performance Score($\downarrow$)}} & 36.83  & 38.49  & 35.48  & 18.20  & 17.21  & 17.00  & 17.24  & 15.92  & 15.10   \\ 
         \multicolumn{2}{c}{\textbf{Average Distillation Ratio($\downarrow$)}} & 62.40  & 65.22  & 60.12  & 30.95  & 29.15  & 28.81  & 29.21  & 26.97  & 25.59   \\ 
        \multicolumn{2}{c}{\textbf{Distillation Difficulty($\uparrow$)}} & 6.70  & 6.65  & 2.00  & 9.14  & 10.07  & 10.13  & 10.14  & 9.82  & 11.32   \\ \bottomrule
    \end{tabular}}
    \caption{Evaluation Results of Phi-2 under Different Secured Proportion}
    \label{tab:appendix:result:Phi2-2}
\end{table*}

\begin{table*}[t]
    \centering
    {\small
    \begin{tabular}{@{}cccccccccccccccc@{}}
    \midrule
        \textbf{} & \textbf{} & \textbf{20M} & \textbf{50M} & \textbf{100M} & \textbf{160M} & \textbf{200M} & \textbf{300M} & \textbf{600M } \\ \midrule
        \multirow{5}{*}{\textbf {Rsn.}} & PIQA & 73.70  & 70.00  & 53.90  & 54.17  & 53.01  & 54.75  & 54.28   \\ 
        \textbf{} & Hellaswag & 59.75  & 55.64  & 28.26  & 27.61  & 26.90  & 27.77  & 28.61   \\ 
        \textbf{} & Winogrande & 66.61  & 67.17  & 51.96  & 51.56  & 52.28  & 51.46  & 50.88   \\ 
        \textbf{} & ARC\_easy & 70.96  & 67.02  & 35.17  & 34.57  & 31.84  & 32.00  & 31.62   \\ 
        \textbf{} & ARC\_challenge & 48.30  & 42.52  & 21.84  & 19.45  & 20.39  & 20.00  & 20.56   \\ \midrule
        \multirow{5}{*}{\textbf {Read.} } & OpenBookQA & 45.33  & 41.27  & 26.13  & 25.80  & 25.60  & 26.33  & 26.53   \\ 
        \textbf{} & LAMBADA & 35.64  & 25.34  & 1.93  & 3.25  & 2.17  & 3.87  & 5.78   \\ 
        \textbf{} & BoolQ & 75.37  & 66.25  & 51.66  & 47.29  & 40.81  & 45.62  & 47.69   \\ 
        \textbf{} & SQuADv2\_em & 10.62  & 0.10  & 0.14  & 0.02  & 0.02  & 0.01  & 0.00   \\ 
        \textbf{} & SQuADv2\_f1 & 38.28  & 22.83  & 1.33  & 2.61  & 1.36  & 0.57  & 1.13   \\ \midrule
        \multirow{2}{*}{\textbf {Knl.} } & NaturalQuestions  & 5.44  & 4.51  & 0.06  & 0.00  & 0.02  & 0.01  & 0.05   \\ 
        \textbf{} & TriviaQA & 12.34  & 12.77  & 0.05  & 0.01  & 0.01  & 0.02  & 0.01   \\ \midrule
        \multirow{2}{*}{\textbf{Code}} & HumanEval & 20.94  & 10.98  & 0.00  & 0.00  & 0.00  & 0.00  & 0.00   \\ 
        \textbf{} & MBPP & 12.60  & 13.40  & 0.00  & 0.00  & 0.00  & 0.00  & 0.00   \\ \midrule
        \textbf{Math} & GSM8K & 7.52  & 7.78  & 0.00  & 0.00  & 0.00  & 0.00  & 0.00   \\ \midrule
        \multirow{2}{*}{\textbf{Gen.}} & MMLU & 43.07  & 39.45  & 26.26  & 26.16  & 25.85  & 25.79  & 25.38   \\ 
        \textbf{} & BBH & 12.35  & 18.02  & 0.00  & 0.01  & 0.00  & 0.79  & 0.12   \\ \midrule
        \multicolumn{2}{c}{\textbf{Avg. Performance Score($\downarrow$)}} & 37.57  & 33.24  & 17.57  & 17.21  & 16.49  & 17.00  & 17.22   \\ 
         \multicolumn{2}{c}{\textbf{Average Distillation Ratio($\downarrow$)}} & 63.67  & 56.32  & 29.77  & 29.15  & 27.93  & 28.81  & 29.17   \\ 
        \multicolumn{2}{c}{\textbf{Distillation Difficulty($\uparrow$)}} & 2.07  & 7.96  & 9.25  & 9.96  & 10.08  & 10.13  & 10.22   \\ \bottomrule
    \end{tabular}}
    \caption{Evaluation Results of Phi-2 under Different Secured Quantity}
    \label{tab:appendix:result:Phi2-3}
\end{table*}

\begin{table*}[t]
    \centering
    {\small
    \setlength{\tabcolsep}{3pt}
    \begin{tabular}{@{}ccccccccccc@{}}
    \toprule
        & & \textbf{0.25\%} & \textbf{0.5\%} & \textbf{1\%} & \textbf{3\%} & \textbf{7\%} & \textbf{15\%} & \textbf{30\%} & \textbf{50\%} & \textbf{100\%} \\ \midrule
        \multirow{5}{*}{\textbf {Rsn.}} & PIQA & 68.21  & 68.37  & 69.68  & 65.85  & 53.43  & 52.94  & 52.36  & 51.25  & 50.44   \\ 
        & Hellaswag & 49.05  & 49.18  & 49.30  & 30.72  & 26.27  & 26.74  & 27.02  & 26.10  & 25.05   \\ 
        & Winogrande & 63.83  & 64.91  & 61.20  & 58.04  & 51.09  & 51.38  & 50.25  & 50.22  & 49.12   \\ 
        & ARC\_easy & 62.94  & 62.89  & 62.25  & 35.15  & 30.81  & 29.27  & 29.64  & 27.99  & 27.50   \\ 
        & ARC\_challenge & 36.98  & 37.49  & 32.91  & 25.97  & 20.56  & 20.36  & 20.08  & 20.88  & 21.22   \\  \midrule
        \multirow{5}{*}{\textbf {Read.} } & OpenBookQA & 39.07  & 40.20  & 35.00  & 33.87  & 26.60  & 27.67  & 27.73  & 26.47  & 26.87   \\ 
        & LAMBADA & 24.71  & 24.99  & 25.36  & 0.11  & 0.59  & 0.78  & 1.15  & 0.06  & 0.00   \\ 
        & BoolQ & 59.43  & 59.35  & 63.49  & 41.01  & 46.98  & 51.59  & 46.46  & 44.02  & 46.28   \\ 
        & SQuADv2\_em & 15.65  & 16.00  & 3.13  & 0.50  & 0.00  & 0.01  & 0.03  & 0.00  & 0.00   \\ 
        & SQuADv2\_f1 & 32.62  & 32.62  & 14.88  & 0.56  & 0.78  & 1.24  & 2.29  & 1.58  & 1.60   \\ \midrule
        \multirow{2}{*}{\textbf {Knl.} } & NaturalQuestions  & 2.72  & 2.64  & 0.32  & 0.03  & 0.04  & 0.03  & 0.05  & 0.03  & 0.01   \\ 
        & TriviaQA & 8.17  & 7.96  & 5.69  & 0.01  & 0.01  & 0.01  & 0.01  & 0.01  & 0.01   \\ \midrule
        \multirow{2}{*}{\textbf{Code}} & HumanEval & 14.43  & 13.41  & 2.03  & 0.00  & 0.00  & 0.00  & 0.00  & 0.00  & 0.00   \\ 
        & MBPP & 17.20  & 18.67  & 6.47  & 0.00  & 0.00  & 0.00  & 0.00  & 0.00  & 0.00   \\ \midrule
        \textbf{Math} & GSM8K & 4.88  & 4.90  & 0.25  & 0.00  & 0.00  & 0.00  & 0.00  & 0.00  & 0.00   \\ \midrule
        \multirow{2}{*}{\textbf{Gen.}} & MMLU & 30.12  & 29.88  & 28.98  & 27.78  & 24.07  & 24.22  & 24.66  & 24.28  & 22.95   \\ 
        & BBH & 4.34  & 3.19  & 0.98  & 0.50  & 0.00  & 0.00  & 0.00  & 0.00  & 0.00   \\ \midrule
        \multicolumn{2}{c}{\textbf{Avg. Performance Score($\downarrow$)}} & 31.43  & 31.57  & 27.17  & 19.41  & 16.54  & 16.84  & 16.57  & 16.05  & 15.94   \\ 
         \multicolumn{2}{c}{\textbf{Average Distillation Ratio($\downarrow$)}} & 66.54  & 66.83  & 57.52  & 41.11  & 35.02  & 35.64  & 35.08  & 33.98  & 33.75   \\ 
        \multicolumn{2}{c}{\textbf{Distillation Difficulty($\uparrow$)}} & 6.18  & 6.15  & 2.76  & 9.28  & 10.08  & 11.19  & 10.54  & 10.23  & 11.26   \\ 
        \bottomrule
    \end{tabular}}
    \caption{Evaluation Results of Phi-1.5 under Different Secured Proportion}
    \label{tab:appendix:result:Phi1.5-2}
\end{table*}

\begin{table*}[t]
    \centering
    {\small
    \begin{tabular}{@{}cccccccccccccccc@{}}
        \toprule
        & & \textbf{20M} & \textbf{50M} & \textbf{100M} & \textbf{160M} & \textbf{200M} & \textbf{300M} & \textbf{600M } \\ \midrule
        \multirow{5}{*}{\textbf {Rsn.}} & PIQA & 69.80  & 65.85  & 53.43  & 52.52  & 52.94  & 53.06  & 53.81   \\ 
        &Hellaswag & 49.51  & 25.72  & 30.27  & 26.31  & 26.74  & 27.05  & 26.51\\
        &Winogrande  & 62.56  & 58.04  & 51.09  & 50.83  & 51.38  & 50.57  & 49.99   \\ 
        & ARC\_easy& 62.41  & 30.15  & 30.81  & 29.14  & 29.27  & 29.62  & 29.67   \\ 
        & ARC\_challenge & 32.51  & 25.97  & 20.56  & 19.97  & 20.36  & 20.48  & 20.79   \\  \midrule
        \multirow{5}{*}{\textbf {Read.} } & OpenBookQA & 35.53  & 33.87  & 26.60  & 26.93  & 27.67  & 28.20  & 26.87   \\ 
        & LAMBADA & 28.14  & 0.11  & 0.59  & 0.45  & 0.78  & 1.30  & 0.61   \\ 
        & BoolQ & 64.77  & 41.01  & 46.98  & 47.33  & 51.59  & 46.09  & 45.59   \\ 
        & SQuADv2\_em & 4.67  & 0.50  & 0.00  & 0.00  & 0.01  & 0.01  & 0.00   \\ 
        & SQuADv2\_f1 & 22.47  & 0.56  & 0.78  & 1.02  & 1.24  & 2.31  & 2.01   \\ \midrule
        \multirow{2}{*}{\textbf {Knl.} } & NaturalQuestions  & 1.64  & 0.03  & 0.04  & 0.05  & 0.03  & 0.06  & 0.03   \\ 
        & TriviaQA & 5.93  & 0.01  & 0.01  & 0.01  & 0.01  & 0.02  & 0.01   \\ \midrule
        \multirow{2}{*}{\textbf{Code}} & HumanEval & 7.73  & 0.00  & 0.00  & 0.00  & 0.00  & 0.00  & 0.00   \\ 
        & MBPP & 7.87  & 0.00  & 0.00  & 0.00  & 0.00  & 0.00  & 0.00   \\ \midrule
        \textbf{Math} & GSM8K & 0.28  & 0.00  & 0.00  & 0.00  & 0.00  & 0.00  & 0.00   \\ \midrule
        \multirow{2}{*}{\textbf{Gen.}} & MMLU & 31.11  & 27.78  & 24.07  & 23.41  & 24.22  & 24.54  & 24.68   \\ 
        & BBH & 3.38  & 0.50  & 0.00  & 0.00  & 0.00  & 0.00  & 0.00   \\ \midrule
        \multicolumn{2}{c}{\textbf{Avg. Performance Score($\downarrow$)}} & 28.84  & 19.89  & 16.54  & 16.35  & 16.84  & 16.67  & 16.50   \\ 
         \multicolumn{2}{c}{\textbf{Average Distillation Ratio($\downarrow$)}} & 61.06  & 41.11  & 35.02  & 34.61  & 35.64  & 35.28  & 34.94   \\ 
        \multicolumn{2}{c}{\textbf{Distillation Difficulty($\uparrow$)}} & 2.81  & 9.28  & 10.26  & 11.65  & 11.19  & 10.87  & 10.49   \\ \bottomrule
    \end{tabular}}
    \caption{Evaluation Results of Phi-1.5 under Different Secured Quantity}
    \label{tab:appendix:result:Phi1.5-3}
\end{table*}

\subsection{Limitation on OPT-350M}
\label{appendix:result:opt}
To investigate the limitations of SOLID, we calculate the Distillation ratio of each secure-source set within the smaller model, OPT-350M~\citep{opt-350} with only 350M parameters. We set the secure-source set size to 2 and subsequently calculate $\Delta \text{ADR}$s for each secure-source set. The detailed results are shown in Figure~\ref{tab:appendix:result:opt}.

\begin{table*}[t]
    \centering
    {\small
    \setlength{\tabcolsep}{1.2pt}
    \begin{tabular}{@{}cccccccccccccccc@{}}
    \toprule
    & & \textbf{Pretrain} & \textbf{0-2} & \textbf{3-5} & \textbf{6-8} & \textbf{9-11} & \textbf{12-14} & \textbf{15-17} & \textbf{18-20} & \textbf{21-23} & \textbf{24-26} & \textbf{27-29} & \textbf{30-32} & \textbf{33-35} & \textbf{*} \\
    \midrule
    \multirow{5}{*}{\textbf{Rsn.}} & PIQA & 64.69 & 61.40 & 62.50 & 61.11 & 56.46 & 58.47 & 58.94 & 61.86 & 62.59 & 63.13 & 61.93 & 62.67 & 63.11 & 49.53 \\
    &Hellaswag & 36.68 & 34.03 & 34.27 & 33.69 & 31.79 & 32.24 & 32.78 & 33.27 & 33.68 & 33.25 & 33.94 & 33.63 & 33.07 & 25.77 \\
    & Winogrande & 52.09 & 51.62 & 52.96 & 51.57 & 50.83 & 52.83 & 51.06 & 51.52 & 51.85 & 52.06 & 52.04 & 51.99 & 50.94 & 49.85 \\
    & ARC\_easy & 44.02 & 40.46 & 40.66 & 40.07 & 35.41 & 37.50 & 37.81 & 39.91 & 40.70 & 41.12 & 41.19 & 40.84 & 39.92 & 26.53 \\
    & ARC\_challenge & 20.82 & 22.27 & 22.61 & 21.25 & 21.39 & 21.25 & 22.01 & 21.36 & 20.25 & 20.48 & 19.88 & 20.99 & 20.28 & 19.82 \\
    \midrule
    \multirow{5}{*}{\textbf{Read.}} & OpenBookQA & 28.00 & 27.60 & 27.47 & 27.40 & 27.40 & 27.27 & 26.20 & 26.47 & 27.67 & 26.80 & 28.67 & 27.67 & 27.13 & 27.47 \\
    & LAMBADA & 40.47 & 30.62 & 32.97 & 28.62 & 21.65 & 23.87 & 28.23 & 29.07 & 29.83 & 29.81 & 31.72 & 31.43 & 18.08 & 0.00 \\
    & BoolQ & 57.74 & 50.87 & 48.51 & 50.58 & 51.60 & 52.83 & 53.42 & 54.37 & 53.30 & 51.42 & 59.79 & 53.14 & 60.42 & 37.83 \\
    & SQuADv2\_em & 11.34 & 6.87 & 7.88 & 4.74 & 4.19 & 0.27 & 0.87 & 2.22 & 3.79 & 3.05 & 4.11 & 4.69 & 2.35 & 0.00 \\
    & SQuADv2\_f1 & 19.35 & 16.27 & 17.00 & 12.00 & 11.72 & 9.04 & 6.92 & 10.11 & 10.90 & 10.08 & 8.88 & 11.47 & 7.30 & 0.01 \\
    \midrule
    \multirow{2}{*}{\textbf{Knl.}} & NaturalQuestions & 1.08 & 1.05 & 0.83 & 0.83 & 0.78 & 0.55 & 0.69 & 0.41 & 1.00 & 0.85 & 0.71 & 0.52 & 0.75 & 0.04 \\
    & TriviaQA & 4.48 & 2.24 & 2.66 & 2.01 & 2.16 & 1.41 & 1.06 & 2.39 & 2.38 & 2.29 & 1.57 & 1.90 & 1.76 & 0.02 \\
    \midrule
    \multirow{2}{*}{\textbf{Code}} & HumanEval & 0.00 & 0.00 & 0.00 & 0.00 & 0.00 & 0.00 & 0.00 & 0.00 & 0.00 & 0.00 & 0.00 & 0.00 & 0.00 & 0.00 \\
    & MBPP & 0.00 & 0.00 & 0.00 & 0.00 & 0.00 & 0.00 & 0.00 & 0.00 & 0.00 & 0.00 & 0.00 & 0.00 & 0.00 & 0.00 \\
    \midrule
    \textbf{Math} & GSM8K & 1.59 & 0.15 & 0.25 & 0.00 & 0.08 & 0.00 & 0.05 & 0.00 & 0.03 & 0.18 & 0.00 & 0.00 & 0.00 & 0.00 \\
    \midrule
    \multirow{2}{*}{\textbf{Gen.}} & MMLU & 26.05 & 25.52 & 26.02 & 25.20 & 25.60 & 25.05 & 25.73 & 23.97 & 25.57 & 26.04 & 25.13 & 25.17 & 25.73 & 22.95 \\
    & BBH & 16.97 & 6.87 & 12.58 & 5.51 & 5.11 & 2.55 & 5.98 & 2.74 & 11.15 & 13.98 & 14.25 & 13.02 & 12.57 & 0.00 \\
    \midrule
    \multicolumn{2}{c}{\textbf{Avg. Performance Score($\downarrow$)}} & 25.02  & 22.23  & 22.89  & 21.44  & 20.36  & 20.30  & 20.69  & 21.16  & 22.04  & 22.03  & 22.58  & 22.30  & 21.38&15.28  \\ 
     \multicolumn{2}{c}{\textbf{Average Distillation Ratio($\downarrow$)}} & - & 88.83  & 91.49  & 85.71  & 81.38  & 81.13  & 82.69  & 84.55  & 88.08  & 88.05  & 90.23  & 89.13  & 85.43  &61.08\\ 
    \multicolumn{2}{c}{\textbf{Distillation Difficulty($\uparrow$)}} &-& 5.92  & 9.32  & 9.04  & 8.60  & 8.83  & 8.73  & 7.17  & 5.82  & 5.18  & 4.65  & 4.92  & 4.15 &10.89 \\ 
        \bottomrule
    \end{tabular}}
    \caption{Evaluation Results of OPT-350M under Different Secured Layers. ``*'' indicates the fully secured model.}
    \label{tab:appendix:result:opt}
\end{table*}

\end{document}